\documentclass{article}

  \usepackage[main, final]{neurips_2026}

\usepackage[utf8]{inputenc} 
\usepackage[T1]{fontenc}    
\usepackage[dvipsnames]{xcolor}
\usepackage[
  colorlinks=true,
  linkcolor=NavyBlue,
  citecolor=TealBlue,
  urlcolor=RoyalBlue
]{hyperref}
\usepackage{url}            
\usepackage{booktabs}       
\usepackage{amsfonts}       
\usepackage{nicefrac}       
\usepackage{microtype}      
\usepackage[table]{xcolor}
\usepackage{tabularx}
\usepackage{array}

\usepackage{amsmath}
\usepackage{amssymb}
\usepackage{mathtools}
\usepackage{amsthm}
\usepackage{inconsolata}    
\usepackage{float}
\usepackage{xspace}
\usepackage{url}
\usepackage[most]{tcolorbox} 
\usepackage{enumitem}
\usepackage{listings}
\tcbuselibrary{listings,skins,breakable}
\usepackage{adjustbox} 
\usepackage[ruled,linesnumbered,noend]{algorithm2e}
\usepackage{wrapfig}
\usepackage{caption}
\usepackage{graphicx}
\usepackage{subcaption}
\usepackage{algcompatible}
\usepackage{titletoc}
\usepackage{changepage}

\newcommand{\evotd}{\textsc{EvoTD}\xspace}
\definecolor{rowblue}{RGB}{230, 245, 255}
\definecolor{tblue}{RGB}{10,186,181}
\newcommand{\rowhighlight}{\rowcolor{gray!15}}

\title{Evolutionary Task Discovery:\\Advancing Reasoning Frontiers via Skill Composition and Complexity Scaling}

\author{%
  \textbf{Liqin Ye\textsuperscript{1}}\thanks{Correspondence to: Liqin Ye $\langle$\texttt{liqiny@gatech.edu}$\rangle$, Chao Zhang $\langle$\texttt{chaozhang@gatech.edu}$\rangle$} \quad \quad
  \textbf{Yanbin Yin\textsuperscript{1}} \quad \quad
  \textbf{Michael Galarnyk\textsuperscript{1}} \quad \quad
  \textbf{Yuzhao Heng\textsuperscript{1}} \\
  \textbf{Sudheer Chava\textsuperscript{1}} \quad \quad
  \textbf{Chao Zhang\textsuperscript{1}}\\[3pt]
  \textsuperscript{1}\,Georgia Institute of Technology
}

\begin{document}

\maketitle

\begin{abstract}
The reasoning frontier of Large Language Models (LLMs) has advanced significantly through modern post-training paradigms (e.g., Reinforcement Learning from Verifiable Rewards (RLVR)). However, the efficacy of these methods remains fundamentally constrained by the diversity and complexity of the training data. One practical solution is data synthesis; yet, prevalent methods relying on unstructured mutation or exploration suffer from \emph{homogeneity collapse}, failing to systematically expand the reasoning frontier. To overcome this, we propose \textbf{Evo}lutionary \textbf{T}ask \textbf{D}iscovery (\evotd), a framework that treats data synthesis as a directed search over a dual-axis manifold of \emph{Algorithmic Skills} and \emph{Complexity Attributes}. We introduce structured evolutionary operators to navigate this space: a \texttt{Crossover} operator that synthesizes novel skill compositions to enhance diversity, and a Parametric \texttt{Mutation} operator that scales structural constraints (e.g., input size, tree depth) to drive robust generalization. Crucially, we integrate a dynamic \emph{Zone of Proximal Development} filter, ensuring tasks lie within the learnable region of the model. Empirically, \textsc{EvoTD} delivers substantial reasoning gains that generalize consistently across model architectures, pretraining regimes, and scales, demonstrating that structured evolutionary curricula can effectively support reasoning improvement. We release our code on \hyperlink{https://github.com/liqinye/EvoTD}{https://github.com/liqinye/EvoTD}.
\end{abstract}
\label{section:abstract}
\section{Introduction}
\label{section:introduction}

The reasoning abilities of LLMs have advanced significantly, with models like DeepSeek-R1~\citep{guo2025deepseek} and Gemini 3 Pro~\citep{gemini2025gemini} achieving impressive performance on challenging benchmarks such as Humanity's Last Exam \citep{phan2025humanity} and AIME \citep{aime2025}. A key driver of this progress is Reinforcement Learning with Verifiable Rewards (RLVR), which leverages programmatically validated signals such as code correctness or solution accuracy, instead of relying on human-annotated examples \citep{lambert2024tulu, guo2025deepseek}. Yet the effectiveness of RLVR is tied to the diversity and quality of the training corpus. Reasoning training relies heavily on costly human-curated question–answer pairs~\citep{cobbe2021trainingverifierssolvemath} or noisy web-scraped corpora~\citep{moshkov2025aimo2winningsolutionbuilding}, both of which remain static and constrain scalability as well as the exploration of reasoning difficulty~\citep{zhang2025surveyreinforcementlearninglarge}.

One promising response to these data limitations is data synthesis, where strong LLMs manufacture training samples~\citep{jung2025prismaticsynthesisgradientbaseddata}. However, current methods suffer from critical structural limitations. Standard training synthesis approaches \citep{zhao2025absolute, xu2024wizardlm, luo2023wizardcoder, luo2023wizardmath} typically rely on unstructured mutation elicited via prompting (e.g., ``make this problem harder'', ``add more constraints''). This frequently results in generating near-duplicate samples \citep{shumailov2024curserecursiontraininggenerated, kim2025sedi}. Diversity-aware approaches regularize this by preserving samples with low textual similarity scores (e.g., BLEU distance, ROUGH-L) \citep{xia2025agent0, huang2025r, wang2023self}; yet, such textual similarity is not necessarily equivalent to the algorithm diversity. A single algorithm can be expressed using completely different linguistic formulations. Recent works have begun to organize synthesis around explicit skill taxonomies \citep{shah2025aiassistedgenerationdifficultmath, zeng2025evaltree, khan2025dataenvgymdatagenerationagents}, ensuring that the training distribution covers a diverse set of logical concepts. While these methods successfully categorize \emph{what} logic is required (the skill), they fundamentally neglect \emph{how} that logic is constrained (the complexity). This leads to a ``flat'' curriculum that covers the breadth of algorithmic concepts but fails to systematically scale the computational complexity to induce generalization. Consequently, there remains a critical gap for a generative framework capable of automatically manipulating both logical backbones and structural scales of reasoning tasks to drive robust, post-training improvement.

In light of this, we propose \textbf{Evo}lutionary \textbf{T}ask \textbf{D}iscovery (\evotd), a novel data synthesis framework that treats task generation as a directed search over a structured abstraction space. As the total complexity of a task is a product of both intrinsic complexity of the general logic and the extrinsic constraints of the problem instance \citep{smith2012measuring, doctor2023toward}, we conceptualize a reasoning task as a composition of two orthogonal dimensions: \emph{Algorithmic Skills} (the logic backbone, e.g., binary search, heaps) and \emph{Complexity Attributes} (the structural constraints, e.g., input size, graph size). Rather than relying on black-box prompting to ``make this diverse and harder'',  we introduce a principled evolutionary task discovery mechanism. Motivated by Evolutionary Algorithms \citep{holland1973genetic}, we formalize a \texttt{Attribute Mutation} operator (which scales complexity attributes) and a \texttt{Skill Crossover} operator (which explores novel skill combinations) to systematically traverse the task space. Our framework adopt a proposer-solver paradigm where the proposer synthesizes tasks and the solver learns from the tasks. To ensure the task falls within solver's \emph{Zone of Proximal Development} (ZPD) \citep{vygotsky1978mind}, we retain tasks that are neither trivial nor impossible. Crucially, because learnability is assessed against the current solver policy $\pi_{\theta}$, our filter implements a dynamic curriculum: as the solver improves, previously impossible tasks become learnable, and the training distribution automatically shifts toward harder instances. Complementing this learnability filter, we leverage LLM's metacognitive capability \citep{didolkar2024metacognitive} to verify the skill alignment of synthesized tasks. Through this rigorous cycle of evolutionary proposal and filtering, \evotd constructs a diverse, adaptive curriculum that continuously pushes models' reasoning frontier.

Empirical evaluations across five diverse benchmarks, spanning code generation and mathematical reasoning, demonstrate that \evotd significantly outperforms standard heuristic synthesis paradigms. Notably, on the Qwen3-4B reasoning model, our method achieves substantial gains on the AIME 24 (+8.2\%) and AIME 25 (+7.2\%) benchmarks. We further show that our method generalizes across three orthogonal axes of model variation: architecture (Qwen3, LLaMA), pretraining regime
(Base, Instruct), and parameter scale (3--8B), attaining the best performance on every evaluated backbone. Beyond code and math, gains transfer cleanly to general-domain reasoning benchmarks (MMLU-Pro, SuperGPQA), with the largest improvements
concentrated on STEM-adjacent disciplines where rigorous reasoning is the
binding constraint.


We make the following contributions:
\begin{itemize}[topsep=0pt, itemsep=0.2pt, leftmargin=1.55em]
    \item We formalize reasoning task synthesis as directed traversal over a dual-axis manifold: \emph{Algorithmic Skills} and  \emph{Complexity Attributes}. This abstraction moves beyond naive prompting, allowing control of logic and constraints to generate quality reasoning curricula.
    \item We introduce two evolutionary operators for task synthesis: \texttt{Attribute Mutation} for complexity scaling and \texttt{Skill Crossover} for composition discovery. This dual-operator design empowers \evotd to systematically explore the boundaries of the task space.
    \item We conduct extensive evaluations across five benchmarks including code generation and mathematical reasoning, demonstrating the utility of our framework. 
\end{itemize}

\section{\evotd}
\label{section:method}

\begin{figure*}[t]
  \centering
  \includegraphics[width=1.0\linewidth]{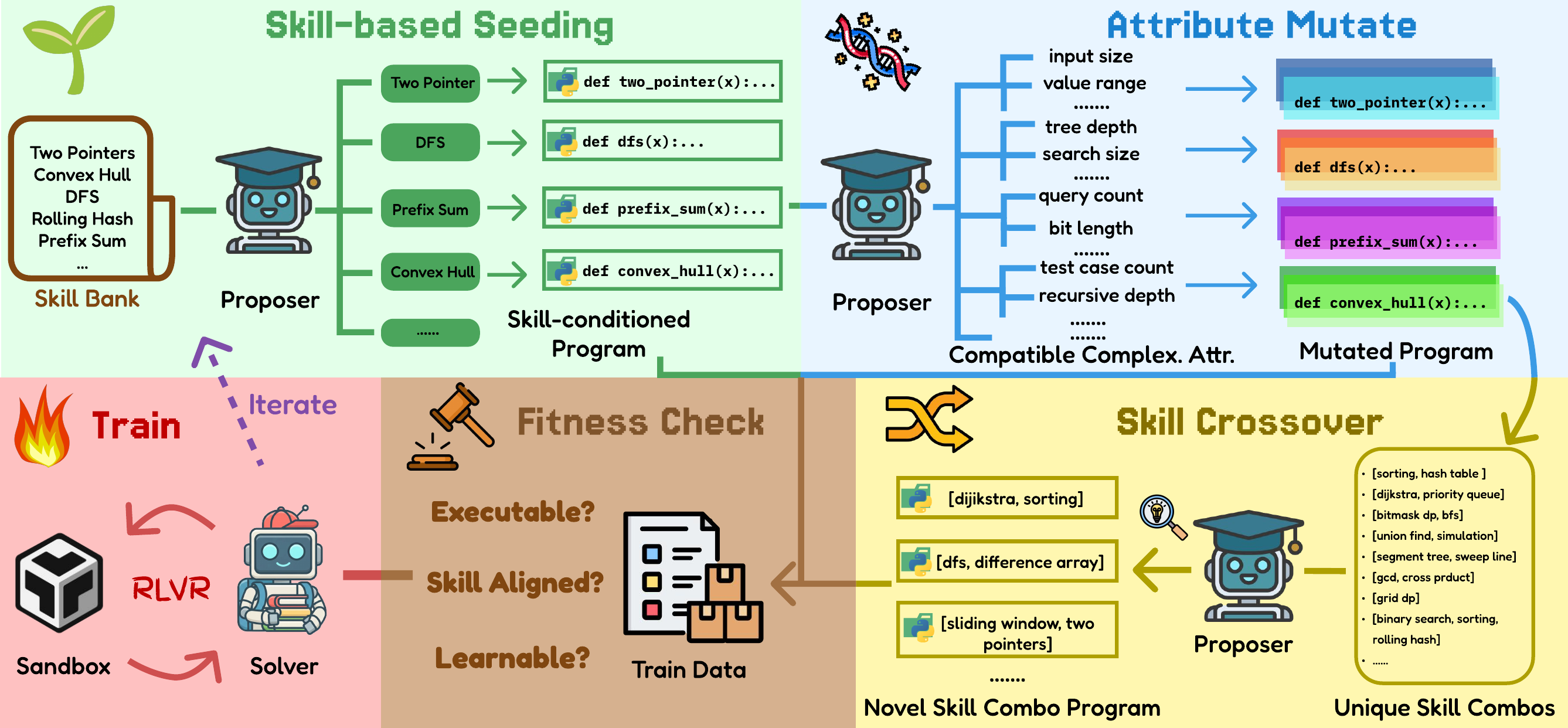}
  \caption{\evotd Framework. (1) Skill-based Seeding: generate skill conditioned programs. (2) Complexity Attribute Mutation: mutate existing programs based on applicable complexity attributes. (3) Skill Crossover: synthesize programs with new skill combinations. (4) Fitness Check: verify programs via executability, skill alignment, and learnability. (5) Train: train solvers with valid tasks.}
  \label{fig:main}
\end{figure*}

In this section, we introduce \evotd, an automated framework designed to systematically expand the diversity and complexity of reasoning tasks through the learning process. We begin by formalizing the space of executable reasoning tasks across deduction, abduction, and induction modes (\textsection\ref{subsection:formulation}). We then define our dual-axis abstraction, which explicitly decouples \emph{algorithmic skills} from \emph{structural complexity} to enable precise evolutionary control (\textsection\ref{subsection:abstraction}). Finally, we detail the evolutionary synthesis engine, describing how attribute mutation and skill crossover operators cooperatively traverse this space to discover novel training samples (\textsection\ref{subsection:synthesis}). Our training pipeline is built upon verifier-based RL paradigm; formal background is provided in Appendix \ref{section:preliminaries}. The complete procedural flow of our framework is outlined in Figure \ref{fig:main} and Algorithm \ref{alg:evotd}.

\subsection{Problem Formulation}
\label{subsection:formulation}
To study how training data can systematically \emph{push} an LLM's algorithmic reasoning frontier, we need to first define the reasoning task. We follow the same task formalization as \citet{zhao2025absolute}. Let $\mathcal{P}$ denote a program space, $\mathcal{I}$ an input space, and $\mathcal{O}$ an output space. An algorithmic reasoning task can be represented as a program-input-output triplet $(p,i,o)\in \mathcal{P}\times \mathcal{I}\times \mathcal{O}$, where $o=p(i)$ is obtained by executing $p$ on $i$. We define distinct \emph{modes of reasoning} by which elements of the triplet are observed versus predicted:
\begin{itemize}[topsep=0pt, itemsep=0.2pt, leftmargin=1.55em]
    \item \textbf{Deduction} $\mathcal{T}_{\text{ded}}$: Given a program $p$ and an input $i$, predict the output $o$. This tests the model's forward thinking to simulate logic step-by-step, $(p,i \rightarrow o)$.
    \item \textbf{Abduction} $\mathcal{T}_{\text{abd}}$: Given a program $p$ and an output $o$, infer a valid input $i$ such that $p(i)=o$. This requires reverse-engineering the logical constraints, $(p,o \rightarrow i)$.
    \item \textbf{Induction} $\mathcal{T}_{\text{ind}}$: Given a natural language problem statement $m$ and a set of input-output pairs $\{(i_n,o_n)\}_{n=1}^N$, synthesize a program $p$ that satisfies $p(i_n)=o_n$ for all $n$. This tests the model's ability to infer executable rules from specifications and observations, $(m,\{(i_n,o_n)\}_{n=1}^N \rightarrow p)$.
\end{itemize}

\paragraph{Executable Task Instantiation}
Each synthesized task is grounded in executable code, an expressive and verifiable representation for reasoning problems. To ensure all tasks are grounded, we follow the generative procedure from \citet{zhao2025absolute}. We employ a Python code executor as an oracle. For Deduction and Abduction, the task proposer synthesizes a program-input pair $(p,i)$ and executes it to obtain the deterministic output $o$. The solver is then presented with the respective partial tuple ($(i,p)$ for Deduction and $(p,o$) for Abduction) to predict the corresponding missing part. For Induction, we utilize the pool of verified programs generated during Deduction and Abduction phases. The proposer conditions on an existing program $p$ to generate a natural language problem statement $m$ and a set of input-output pairs $\{(i_n,o_n)\}_{n=1}^N$. The solver is provided with a problem prompt and a subset of input-output pairs, while the remainder pairs serve as hidden test cases to verify the correctness of the solver-synthesized program. 

\subsection{Task Abstraction}
\label{subsection:abstraction}
To enable evolutionary discovery, we first establish a structured representation of the task space. 
Direct mutation of raw problem specifications is often inefficient, as the space of valid coding tasks is sparse and highly discontinuous \citep{luo2025tree, orvalho2025large}. Random textual perturbations often fail to induce meaningful shifts in algorithmic reasoning, instead producing syntactically malformed code or preserving original computational complexity \citep{zhang2025semantic, novikov2025alphaevolve, abed2025increasing}. Consequently, standard evolutionary approaches which rely on unstructured mutation struggle to discover novel regions of the solution space without explicit semantic guidance \citep{van2025code, hemberg2024evolving}. Therefore, we propose to control task evolution through a low-dimensional abstraction that decouples the \emph{semantic logic} of a coding task from its \emph{computational structure}.

We conceptualize a coding task $t$ as a point in a dual-axis manifold $\mathcal{T} \approx \mathcal{S} \times \mathcal{C}$, where $\mathcal{S}$ represents \textbf{Algorithmic Skills} and $\mathcal{C}$ represents \textbf{Complexity Attributes}. We describe these two axes as follows:

\paragraph{The Semantic Axis: Algorithmic Skills ($\boldsymbol{\mathcal{S}}$)}  Reasoning is inherently compositional. Complex problems are rarely monolithic: they are aggregates of atomic logical primitives \citep{piantadosi2016logical,lippl2025algorithmic}. We define a ``Skill'' $s \in \mathcal{S}$ as a fundamental algorithmic pattern required to solve a problem (e.g. \emph{two pointers, dijkstra}). By abstracting tasks into skills, we transform the opaque goal of ``diversity'' into a measurable coverage problem. This allows our evolutionary operators to target specific ``blind spots'' in model's skill set, rather than hoping for diversity through randomness.

\paragraph{The Structural Axis: Complexity Attributes ($\boldsymbol{\mathcal{C}})$} While the algorithmic skills establish the \emph{qualitative} reasoning method, the complexity attributes dictate the \emph{quantitative} magnitude of the problem instance. A \emph{Depth-First Search} task can be trivial on a tree of depth 3 but challenging on a graph with cycles and $10^4$ nodes. We define ``Complexity Attributes'' $c \in \mathcal{C}$ as the set of parametric constraints governing the execution environment (e.g. \emph{input size, value range}). LLMs often learn brittle heuristics for simple coding instances that fail to generalize as problem scale increases \citep{power2022grokking}. By treating $\mathcal{C}$ as an independent axis, we can force the model to move beyond superficial pattern matching toward scale-invariant algorithmic solutions.

The candidate lists for $\mathcal{S}$ and $\mathcal{C}$ can be populated via expert curation, LLM synthesis, or automated extraction. While our framework is agnostic to the source of these axes, in this work, we adopt a data-driven initialization, leveraging LLM metacognitive ability \citep{didolkar2024metacognitive} to extract from a seed corpus \citep{shi2024can}. This approach offers an optimal balance between scalability and domain-specific fidelity. It circumvents the bottleneck of human annotation while anchoring the quality of skills directly in rigorous, real-world problem distributions. The extracted skill bank and complexity attributes are detailed in Figure \ref{fig:skill_list} and \ref{fig:attribute_name_list}.

\subsection{Evolutionary Task Synthesis}
\label{subsection:synthesis}
Given this orthogonal dual-axis abstraction, we propose an evolutionary synthesis engine to systematically traverse this manifold. Let $\Phi(t) = (s, \mathbf{c})$ represent the projection of a task into its skill and complexity attributes. We design two distinct operators: \texttt{Attribute Mutation} and \texttt{Skill Crossover}, aiming for population diversity and frontier expansion. We manipulate these two task axes to generate novel training samples, filtered by a rigorous multi-objective verification protocol.

\paragraph{Skill-based Seeding} 
We initialize the population $\mathcal{D}_{\text{skill}}$ by instantiating a baseline task for each unique skill $s \in \mathcal{S}$ in our skill bank, which we extract from a seed dataset \citep{shi2024can}. During the synthesis of Deduction and Abduction tasks, the generation of program-input pair $(p,i)$ is explicitly conditioned on the target skill. The proposer is prompted to generate a program $p$ that implements the core algorithmic logic of $s$ (See Figure \ref{fig:code_input_prompt} and \ref{fig:code_output_prompt}). This guarantees that our evolutionary search begins with maximal semantic coverage of the foundational skill space.

\paragraph{\texttt{Attribute Mutation}} 
The mutation operator drives the ``vertical'' evolution of complexity by intensifying structural constraints while strictly preserving the logical backbone $s$. Unlike static scaling, $\mathcal{M}_{\text{attr}}$ leverages the proposer LLM's metacognitive ability to perform an attribute compatibility audit. Formally, given a parent program $p \in \{\mathcal{T}_{\text{ded}},\mathcal{T}_{\text{abd}}\}$ synthesized from skill-based seeding stage, the proposer first selects what attributes are applicable: $\pi_{\text{prop}}(p) \to \mathbf{c}^*$, where $\mathbf{c}^* \in 2^{\mathcal{C}}$ is a subset of attributes that are valid to be scaled in $p$ (e.g., selecting \textit{tree depth} only for $p$ containing tree). The mutation is then defined as:
\begin{equation}\mathcal{M}_{\text{attr}}(t, \pi_{\text{prop}}) = t' \quad \text{s.t.} \quad \Phi(t') = (s, \mathbf{c} + \Delta\mathbf{c}')
\end{equation}
where $t$ is a task instance and $\Delta\mathbf{c}'$ represents the directed intensification of the selected attributes by the. For each parent, we generate a diverse cohort of $n$ variants ($n=10$ in this work), each embodying a distinct combinatorial intersection of the LLM-selected complexity attributes.

\paragraph{\texttt{Skill Crossover}} 
The crossover operator $\mathcal{X}_{\text{skill}}: 2^{\mathcal{S}} \rightarrow \mathcal{T}$ enables horizontal exploration through combinatorial logic synthesis. Let $\Lambda(\mathcal{D}) = \{S_1, S_2, \dots, S_n\}$ be the set of skill combinations currently realized in the population via proposer LLM's metacognitive labeling during filtering process (elaborated in the following paragraph). The operator performs a novelty combinatorial search to generate a task $t_{\text{new}}$ possessing undiscovered skill combinations:
\begin{equation}
\mathcal{X}_{\text{skill}}(\Lambda(\mathcal{D}),\pi_{\text{prop}})=t_{\text{new}} \quad \text{s.t.}\;\;\text{Skill}(t_{\text{new}}) = S_{\text{new}} \notin \Lambda(\mathcal{D})
\end{equation}
Because arbitrary permutations of skills can yield conceptually incongruent tasks, we prevent combinatorial degeneration by explicitly prompting the proposer to discover naturally synergistic skill combinations. To ensure this integration is not merely superficial, we further enforce that all selected skills must be essential and deeply interconnected within generated tasks, rather than appearing sequentially (see Figures \ref{fig:code_output_crossover_prompt} and \ref{fig:code_input_crossover_prompt}). By targeting the sparse regions of the meaningful skill-composition set, $\mathcal{X}_{\text{skill}}$ prevents logic stagnation and forces the model to synthesize increasingly novel, multi-skill reasoning tasks.

\paragraph{Multi-objective Fitness Check} 

To ensure the synthesized task is both computationally valid and pedagogically instructive, we enforce a rigorous verification protocol $\mathcal{V}(t)$ modeled as a composite indicator function of three hierarchical objectives as below:
\begin{equation}
\mathcal{V}(t) = v_{\text{exec}}(t) \cdot v_{\text{skill}}(t) \cdot v_{\text{learn}}(t)
\end{equation}
\begin{enumerate}[topsep=0pt, itemsep=0.2pt, leftmargin=1.55em]
\item \textbf{Executability ($\boldsymbol{v}_{\text{exec}}$):} For $\mathcal{T}_{\text{ded}}$ and $\mathcal{T}_{\text{abd}}$, we confirm that the program-input pair $(p, i)$ is free of syntax errors and terminates within limits by a Python executor. For $\mathcal{T}_{\text{ind}}$, we validate that input-output pairs $(i_n,o_n)\}_{n=1}^N$ are faithful by the ground-truth program $p$. 
\item \textbf{Skill Alignment ($\boldsymbol{v}_{\text{skill}}$):} we perform a task's skill audit to ensure the generated task strictly adheres to the requested algorithmic requirements. The task is retained iff the target skill is preserved in the detected skill sets $\hat{\mathcal{S}}$: $\boldsymbol{v}_{\text{align}}(t) = \mathbb{I}[s_{\text{target}} \in \hat{\mathcal{S}}]$.
\item \textbf{Learnability ($\boldsymbol{v}_{\text{learn}}$):} We assess the empirical difficulty using the current solver policy $\pi_{\theta}$'s pass rate over $k$ attempts. We strictly filter for non-trivial yet solvable instances (avg@$k \in (0, 1)$):
\begin{equation}
v_{\text{learn}}(t) = \mathbb{I}[0 < \mathbb{E}_{k \sim \pi_{\theta}}[\text{solved}(t_k)] < 1]
\end{equation}
\end{enumerate}


\section{Experiments}
\label{section:experiments}

\subsection{Experimental Setup}
\label{subsection:exp_setup}
\paragraph{Implementation Details} We evaluate \evotd on two model types as our solver: 1) Base models: Qwen3-4B-Base and Qwen3-8B-Base \citep{qwen3technicalreport} and 2) Reasoning models: Qwen3-4B with thinking enabled \citep{qwen3technicalreport}. Our training pipeline is built based on VeRL framework \citep{sheng2025hybridflow}. We adopt o4-mini \citep{singh2025openai} as the proposer in our main experiments. To assess \evotd's sensitivity to this choice, Appendix~\ref{app:proposer_ablation} reports an ablation with two open-weight proposers, gpt-oss-120b and deepseek-r1-0528. To efficiently leverage the synthesized data and modulate the difficulty, the proposer is prompted to inject natural language hints to the induction task which is too opaque (\textit{avg@k}$=0$). This helps bootstrap more samples falling within solver's ZPD. For additional implementation details, see Appendix \ref{app:exp_details}. 

\begin{table*}[t!]
\caption{Results on Thinking models across code and math benchmarks. \textbf{Bold} means peak performance. \underline{Underline} means second-best.}
\centering
\normalsize
\resizebox{\textwidth}{!}{%
\begin{tabular}{l |
                c c |
                c c c |
                c c c}
  \toprule[1.2pt]
    \textbf{Model} & LCB\textsuperscript{v6} & MBPP\textsuperscript{+}
    & AME24 & AME25 & Olympiad
    & \textbf{CAvg} & \textbf{MAvg} & \textbf{AVG} \\
    \midrule[1pt]
    \rowhighlight
    \multicolumn{9}{c}{\textit{Pass@1}} \\
    \midrule[1pt]
    
    Qwen3-4B (Thinking)
    & 42.9 & 51.3 & 34.0 & 21.8 & 45.6 & 47.1 & 33.8 & 39.1 \\
    \quad +Evol-Instruct
    & 47.4 & 53.7 & 35.4 & 22.4 & 46.9 & 50.6 & 34.9 & 41.2 \\
    \rowcolor{rowblue}
    \quad +\textsc{EvoTD} (Iter1)
    & 46.8 &  54.7 & 36.7 & 25.8 & 47.3 & 50.8 &  36.6 & 42.3 \\
    \rowcolor{rowblue}
    \quad +\textsc{EvoTD} (Iter2)
    & \underline{48.5} & \bfseries56.2 & \underline{37.9} & \underline{28.3} & \underline{49.1} & \underline{52.4} & \underline{38.4} & \underline{44.0} \\
    \rowcolor{rowblue}
    \quad +\textsc{EvoTD} (Iter3)
    & \bfseries 49.8 & \underline{55.8} & \bfseries 42.2 & \bfseries 29.0 & \bfseries 49.4 & \bfseries 52.8 & \bfseries 40.2 & \bfseries 45.2 \\

    \midrule[1.2pt]
    \rowhighlight
    \multicolumn{9}{c}{\textit{Pass@8}} \\
    \midrule[1.2pt]

    Qwen3-4B (Thinking)
    & 53.8 & 68.3 & 55.4 & 39.0 & 59 & 56.4 & 51.1 & 55.1 \\
    \quad +Evol-Instruct
    & 58.0 & 54.8 & 55.0 & 44.2 & 59.1 & 56.4 & 52.8 & 54.2 \\
    \rowcolor{rowblue}
    \quad +\textsc{EvoTD} (Iter1)
    &  58.9 &  70.9 &  60.2 & 47.3 &  59.7 &  64.9 &  55.7 &  59.4 \\
    \rowcolor{rowblue}
    \quad +\textsc{EvoTD} (Iter2)
    & \underline{60.8} & \bfseries 72.8 & \underline{61.4} & \underline{47.6} & \underline{61.5} & \bfseries 66.8 & \underline{56.8} & \underline{60.8} \\
    \rowcolor{rowblue}
    \quad +\textsc{EvoTD} (Iter3)
    & \bfseries 61.5 & \underline{71.2} & \bfseries 63.2 & \bfseries 49.9 & \bfseries 62.1 & \underline{66.4} & \bfseries 58.4 & \bfseries 61.6 \\

  \bottomrule[1.2pt]
\end{tabular}
}
\label{tab:think_performance}
\end{table*}
\begin{table*}[t!]
\vspace{-5pt}
\caption{Results on Base and Instruct models across code and math benchmarks.}
\centering
\small
\resizebox{\textwidth}{!}{%
\begin{tabular}{l|
                c c |
                c c c |
                c c c}
  \toprule[1.2pt]
    \textbf{Model} & LCB\textsuperscript{v6} & MBPP\textsuperscript{+}
    & AME24 & AME25 & Olympiad
    & \textbf{CAvg} & \textbf{MAvg} & \textbf{AVG} \\
    \midrule[1.2pt]
    
    Qwen3-4B-Base
    & 20.6 & 53.7 & 9.1 & 3.3 & 26.2 & 37.2 & 12.9 & 22.6 \\
    \quad +Evol-Instruct
    & 23.2 & 52.1 & 10.0 & 10.0 & 27.0 & 37.7 & 15.7 & 24.5 \\
    \quad +SPIRAL\footnotemark[1]
    & \underline{27.3} & \underline{61.4} & \underline{10.0} & \textbf{13.3} & \textbf{40.0} & \underline{44.4} & \underline{21.1} & \underline{30.4} \\
    \quad +Agent0
    & \textbf{27.6} & \textbf{63.5} & \underline{10.0} & 3.3 & 34.7 & \textbf{45.6} & 16.0 & 27.8 \\
    \rowcolor{rowblue}
    \quad +\textsc{EvoTD} (Ours)
    & 26.4 & 58.5 & \textbf{19.5} & \underline{10.0} & \underline{39.6} & 42.5 & \bfseries 23.0 & \bfseries 30.8 \\
    
    \midrule[1.2pt]
    
    Qwen3-8B-Base
    & 29.6 &  58.3 & 13.3 & 6.7 & 34.8 & 44.0 & 18.3 & 28.5 \\
    \quad +Evol-Instruct
    & 31.9 & 59.8 & 11.4 & 9.6 & 40.1 & 45.9 & 20.4 & 30.6 \\
    \quad +SPIRAL\footnotemark[1]
    & 28.8 & 58.7 & \underline{14.2} & \textbf{19.6} & \bfseries 42.5 & 43.8 & \bfseries 25.4 & 32.8 \\
    \quad +Agent0
    & \underline{32.0} & \bfseries 71.2 & 13.1 & 10.1 & 41.0 & \textbf{51.6} & 21.4 & \underline{33.5} \\
    \rowcolor{rowblue}
    \quad +\textsc{EvoTD} (Ours)
    & \bfseries 32.8 & \underline{65.3} & \bfseries 16.7 & \underline{16.7} & \bfseries 42.5 & \underline{49.1} & \underline{25.3} & \textbf{34.8} \\
    
    \midrule[1.2pt]

    LLaMA-3.2-3B-Instruct
    & 11.0 & 25.9 & 3.3 & 0.0 & 11.0 & 18.5 & 4.8 & 10.2 \\
    \quad +Evol-Instruct
    & \underline{11.1} & 25.4 & \bfseries 10.0 & 0.0 & \underline{12.3} & 18.3 & \underline{7.4} & \underline{11.8} \\
    \quad +Agent0
    & \bfseries 11.6 & \underline{31.5} & 3.3 & 0.0 & 11.0 & \underline{21.6} & 4.8 & 11.5 \\
    \rowcolor{rowblue}
    \quad +\textsc{EvoTD} (Ours)
    & 10.7 & \bfseries 42.1 & \bfseries 10.0 & 0.0 & \bfseries 14.8 & \bfseries 26.4 & \bfseries 8.3 & \bfseries 15.5 \\
    
    \midrule[1.2pt]
    
    LLaMA-3.1-8B-Instruct
    & 16.6 &  59.8 & 3.3 & 0.0 & 15.4 & 38.2 & 6.2 & 19.0 \\
    \quad +Evol-Instruct
    & 15.7 & 61.1 & 3.3 & \underline{3.3} & \bfseries 18.4 & 38.4 & 8.3 & 20.4 \\
    \quad +SPIRAL\footnotemark[1]
    & \underline{16.7} & 60.6 & 0.8 & \underline{3.3} & 17.5 & \underline{38.7} & 7.2 & 19.8 \\
    \quad +Agent0
    & 16.1 & \underline{61.1} & \bfseries 10.0 & 0.0 & 16.0 & 38.6 & \underline{8.7} & \underline{20.6} \\
    \rowcolor{rowblue}
    \quad +\textsc{EvoTD} (Ours)
    & \bfseries 17.6 & \bfseries 62.4 & \underline{6.7} & \bfseries 6.7 & \underline{17.6} & \bfseries 40.0 & \bfseries 10.3 & \bfseries 22.2 \\
    
  \bottomrule[1.2pt]
\end{tabular}
}
\label{tab:base_performance}
\end{table*}
\paragraph{Datasets} 
To evaluate the model's ability to generalize beyond its training distribution, we conduct evaluations across five benchmarks spanning code generation and mathematical reasoning: 1) Code Generation: MBPP+ \citep{austin2021program} for rigorous correctness checking;  LiveCodeBench v6 \citep{jain2024livecodebenchholisticcontaminationfree} to assess contamination-free problems. 2) Math Reasoning: AIME 24 and 25 \citep{aime2025} to probe the frontier of competitive problem-solving; OlympiadBench \citep{he2024olympiadbench} to assess logical deduction across diverse domains. 3) General Domain Reasoning: MMLU-Pro \citep{wang2024mmlu} for robust multi-disciplinary reasoning; SuperGPQA \citep{du2025supergpqa} for graduate-level expertise across disciplines. For additional details on datasets, see Appendix~\ref{app:dataset_details}.

\paragraph{Metrics} For Base and Instruct models, we employ greedy decoding to report Pass@1, with the exception of the AIME benchmarks, where we duplicate each problem 32 times to report \emph{mean@32}. For Thinking models, we align with the evaluation protocol of \citet{qwen3technicalreport}, using temperature $T=0.6$ and top-$p=0.95$. We generate $k=8$ responses per question across all benchmarks (increased to $k=32$ for AIME) and report both Pass@1 and Pass@8.

\footnotetext[1]{We use the checkpoint \href{https://huggingface.co/spiral-rl/Spiral-Qwen3-4B-Multi-Env}{Spiral-Qwen3-4B-Multi-Env} and \href{https://huggingface.co/spiral-rl/Spiral-Qwen3-8B-Multi-Env}{Spiral-Qwen3-8B-Multi-Env} available on HuggingFace.}

\paragraph{Baselines} We benchmark \evotd against a spectrum of data synthesis paradigms, ranging from heuristic mutation to self-improving. For experiments on base models, we compare against three distinct categories: (1) The Base Model without any finetuning; (2) Evol-Instruct \citep{luo2023wizardcoder, zhao2025absolute}, which represents the standard paradigm of using heuristic prompts to generate data; and (3) Self-Evolving Frameworks: SPIRAL \citep{liu2025spiral} which refines reasoning via self-play zero-sum games, and Agent0 \citep{xia2025agent0} which self-enhances with tool integration. For reasoning model experiments, we limit our comparison to the Base thinking model and Evol-Instruct.

\subsection{Main Results}
\label{subsection:results}
We evaluate \evotd across two regimes: reasoning-tuned ``Thinking'' backbones (Table~\ref{tab:think_performance}) and Base/Instruct backbones (Table~\ref{tab:base_performance}). Complementary results on general-domain benchmarks (MMLU-Pro, SuperGPQA) are reported in Appendix~\ref{app:bench_general_domain}. The first regime tests whether evolutionary task synthesis can extend the frontier of models that already possess strong reasoning capability. The second probes generalization across model families, training regimes, and scales, benchmarking against state-of-the-art self-evolving methods.

\paragraph{Efficacy on Reasoning Models.}
We highlight three distinct dimensions of improvement that validate our evolutionary hypothesis:
\textbf{(1) Holistic Cross-Domain Generalization.} Unlike standard synthesis methods that often incur an ``alignment tax'' \citep{lin2024mitigating} where optimizing for math degrades coding capability, \evotd produces simultaneous improvements on in-domain coding (+5.7\% CAvg) and out-of-domain mathematics (+6.4\% MAvg), This confirms that our \emph{Skill-based Seeding} and \texttt{Skill Crossover} surface abstract logical primitives that transfer across syntactic and symbolic reasoning modalities.
\textbf{(2) Resilience to Homogeneity Collapse.} A canonical failure mode of distillation-style training is mode collapse, where the model overfits to a single solution path, inflating Pass@1 while degrading Pass@$k$ ($k>1$). Evol-Instruct exhibits exactly this pathology: its Pass@8 average (54.2\%) falls below the untuned backbone (55.1\%) despite a higher Pass@1. \evotd instead achieves consistent performance gains in both Pass@1 (+4\%) and Pass@8 (+7.4\%), evidence that our diversity-driven curriculum proliferates reasoning trajectories rather than narrowing them.
\textbf{(3) Amplified Gains on Harder Benchmarks.} The margin between \evotd and the baseline widens with problem difficulty, reaching +8.2\% on AIME24 and +7.2\% on AIME25. This trend substantiates the role of our \emph{Complexity Attribute} axis: training on parametrically more complex mutations teaches the model to disentangle core algorithmic reasoning from superficial instance scale, improving robustness on reasoning-intensive tasks.

\paragraph{Generalization across Architectures, Regimes, and Scales.}
We further demonstrate that \evotd transfers robustly across three orthogonal axes of model variation: architecture (Qwen3 vs.\ LLaMA), pretraining regime (Base vs.\ Instruct), and parameter scale (3/4B vs.\ 8B). \evotd attains the highest overall Pass@1 on every one of the four resulting backbones, indicating that the evolutionary curriculum is not bound to a particular model family, training stage, or size. 
\begin{wrapfigure}{r}{0.53\textwidth}
    \vspace{-0.2\baselineskip}
    \begin{minipage}{0.52\textwidth}
        \centering






\captionsetup{type=table}
\captionof{table}{Evolutionary Operators Ablation}
\vspace{-0.5em}
\label{tab:ablation}
\centering
\footnotesize
\setlength{\tabcolsep}{8pt}
\begin{tabular}{l | c c c}
  \toprule[1.2pt]
    \textbf{Method} & \textbf{CAvg} & \textbf{MAvg} & \textbf{AVG} \\
    \midrule[0.8pt]
    \rowcolor{gray!15}
    \multicolumn{4}{c}{Qwen3-4B Thinking \textit{Pass@1}} \\
    \midrule[1.2pt]

    \rowcolor{rowblue}
    \evotd
    & \underline{52.8} & \textbf{40.2} & \textbf{45.2}  \\
    \quad w/o Attribute Mutation
    & 52.7 & 38.1 & 43.9  \\
    \quad w/o Skill Crossover
    & \textbf{53.1} & \underline{38.6} & \underline{44.4}  \\

    \midrule[0.8pt]
    \rowcolor{gray!15}
    \multicolumn{4}{c}{Qwen3-4B Thinking \textit{Pass@8}} \\
    \midrule[1.2pt]

    \rowcolor{rowblue}
    \evotd
    & \underline{66.4} & \textbf{58.4} & \textbf{61.6}  \\
    \quad w/o Attribute Mutation
    & 66.3 & 56.6 & 60.5  \\
    \quad w/o Skill Crossover
    & \textbf{67.2} & \underline{57.1} & \underline{61.1}  \\

  \bottomrule[1.2pt]
\end{tabular}

        \vspace{0.8em}

        \includegraphics[width=\linewidth]{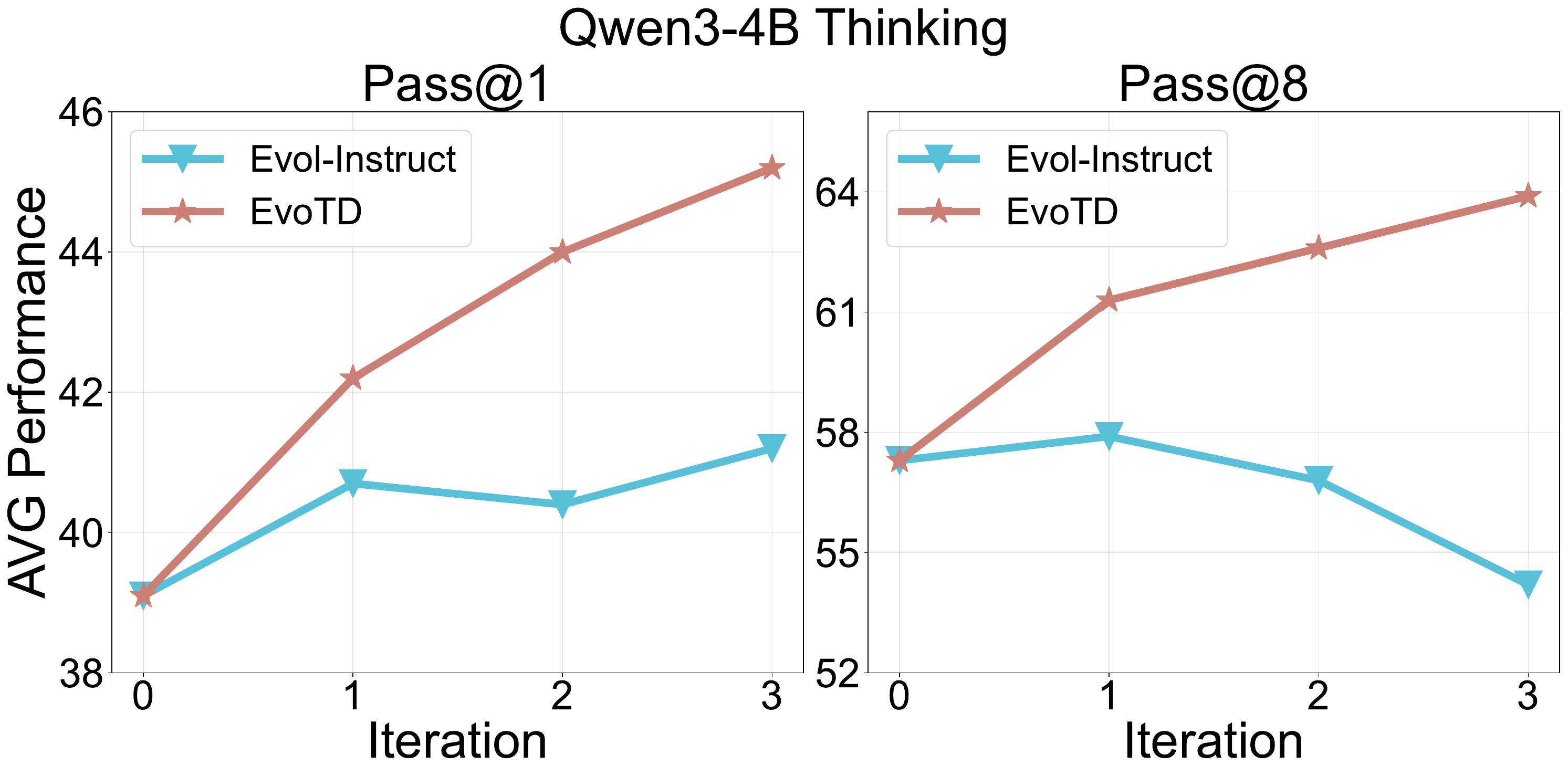}
        \captionsetup{type=figure}
        \captionof{figure}{Scaling performance comparison between Evol-Instruct and \evotd in Pass@1 and Pass@8.}
        \label{fig:iterative_scaling}
    \end{minipage}
    \vspace{-4\baselineskip}
\end{wrapfigure}
Notably, this holds even against two resource-intensive self-evolving baselines: SPIRAL which requires curated zero-sum self-play environments, and Agent0 which utilizes external tools. \evotd leads the next-best method on each backbone by +0.4\%, +1.3\%, +3.7\%, and +1.6\%, respectively. Together, these confirm that \evotd offers a pragmatic way for reasoning enhancement, proving that complexity modulation allows base models to scale efficiently without the prohibitive computational overhead of self-training.

\subsection{Analysis}
\label{subsection:analysis}
\vspace{-5pt}
To decompose the mechanisms driving \evotd's strong performance, we conduct a fine-grained main analysis along four critical dimensions: the necessity of each evolutionary operator via ablation, the iteration dynamics of performance scaling, the skill coverage and compositional novelty, and the difficulty modulation induced by attribute mutation. We also provide qualitative examples for \texttt{Attribute Mutation} and \texttt{Skill Crossover} in Appendix \ref{app:qual_examples}.

\begin{figure*}[ht!]
  \centering
  \begin{minipage}[t]{0.316\textwidth}
    \vspace{0pt}
    \centering
    \includegraphics[width=\linewidth]{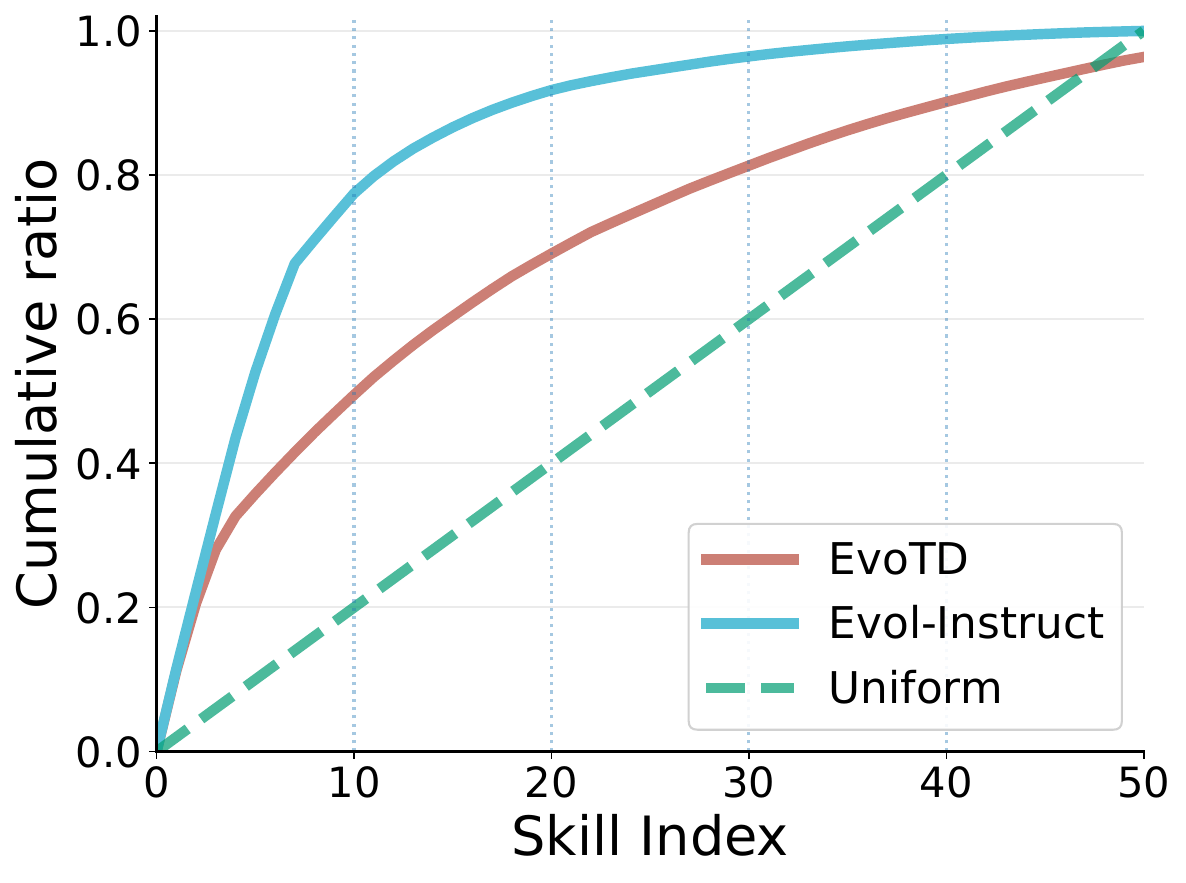}
    \vspace{-0.85em}
    \subcaption{Skill frequency CDF}
    \label{fig:skill_freq}
  \end{minipage}\hfill
  \begin{minipage}[t]{0.32\textwidth}
    \vspace{0pt}
    \centering
    \includegraphics[width=\linewidth]{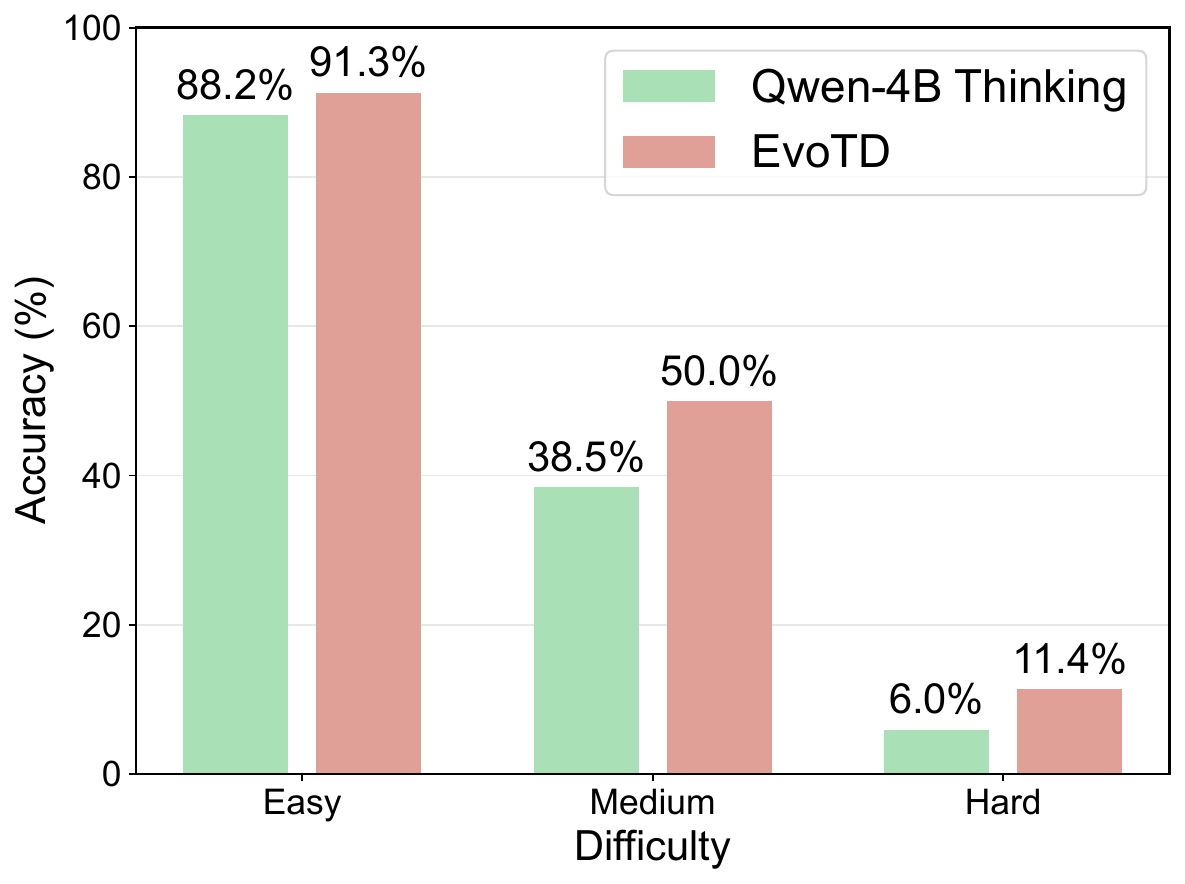}
    \subcaption{LCB\textsuperscript{v6} performance breakdown}
    \label{fig:lcb_breakdown}
  \end{minipage}\hfill
  \begin{minipage}[t]{0.33\textwidth}
    \vspace{1.1em}
    \centering

    \setlength{\tabcolsep}{4.3pt}
    \renewcommand{\arraystretch}{1.05}
    \small
    \begin{tabular}{lcc}
      \toprule
      Iter & \textsc{EvoTD} & Evol-Instruct \\
      \midrule
      1 & \textbf{41.8\%} & 33.2\% \\
      2 & \textbf{43.5\%} & 36.4\% \\
      3 & \textbf{43.1\%} & 33.5\% \\
      \midrule
      Overall & \textbf{32.7\%} & 24.2\% \\
      \bottomrule
    \end{tabular}
    \vspace{1.55em}
    \subcaption{Unique skill combination ratio}
    \label{tab:skill_combo_unique}
  \end{minipage}

  \caption{(a) shows the cumulative distribution of skill frequencies, where the uniform reference indicates perfectly balanced skill coverage. (b) shows the detailed performance breakdown on LiveCodeBench\textsuperscript{v6}. (c) shows unique skill combination ratio over iterations.}
  \label{fig:analysis_panels}
\end{figure*}

\paragraph{Ablation of Evolutionary Operators}
To disentangle the individual contributions of our dual-axis design, we ablate the \texttt{Attribute Mutation} and \texttt{Skill Crossover} operators. As shown in Table \ref{tab:ablation}, removing \texttt{Attribute Mutation} causes a sharp degradation in mathematical reasoning ($\Delta$Pass@1 of -2.1\%), confirming that complexity scaling is prerequisite for deep, scale-invariant logic. Similarly, eliminating \texttt{Skill Crossover} leads to a notable decline in aggregate performance, indicating that combinatorial expansion is essential for broad generalization. We include a more detailed experiment results and discussion in Appendix \ref{app:evo-op-ablation}.

\paragraph{Better Scaling of Iterative Evolution} 
As illustrated in Figure \ref{fig:iterative_scaling}, \evotd demonstrates robust, monotonic scaling in both Pass@1 and Pass@8 metrics throughout the iterative process. In contrast, the Evol-Instruct baseline exhibits early saturation, with performance plateauing after the first iteration. Notably, in the multi-sample regime (Pass@8), the baseline suffers performance degradation in later stages (dropping from 58\% to 54\%), a phenomenon indicator of mode collapse. Conversely, \evotd maintains a positive gradient, confirming that our structured, verified evolution is essential for sustaining long-term learning gains.

\paragraph{Skill Uniformity and Composition Coverage} To validate the efficacy of \emph{Skill-based Seeding} and \texttt{Skill Crossover}, we analyze the topological diversity of the synthesized task manifold. We first quantify semantic balance by examining the Cumulative Distribution Function (CDF) of skill frequencies (See Figure \ref{fig:skill_freq}). \evotd aligns significantly closer to a uniform ideal than baselines, confirming that our seeding strategy maximizes the semantic entropy of the training signal. Unlike Evol-Instruct, it prevents mode collapse into dominant patterns. Complementing this, we measure unique skill combination ratio, a proxy for combinatorial novelty. Table \ref{tab:skill_combo_unique} reveals that the crossover operator consistently drives a novelty search, synthesizing a significantly higher volume of previously unseen skill combinations. Together, these metrics confirm that \evotd does not merely scale data volume, but actively expands the combinatorial frontier of the curriculum, ensuring the model is exposed to structurally diverse reasoning primitives.

\paragraph{Complexity Modulation via Attribute Mutation} Finally, a core hypothesis of \evotd is that evolving complexity attributes ($c \in \mathcal{C}$) drives the model to master harder instances of the same logic. To verify this, we measure the empirical difficulty of mutated tasks compared to their seeds. The mutation operator consistently lowers the solver's success rate, reducing \textit{mean@10} from 56.7\% to 51.4\%. Crucially, this exposure to ``harder'' training distributions translates directly to superior generalization on rigorous benchmarks. As illustrated in Figure \ref{fig:lcb_breakdown}, \evotd achieves its largest relative gains (of 90\%) on the ``Hard'' subset of LiveCodeBench\textsuperscript{v6}.

\paragraph{Additional Analyses}
Beyond the four dimensions above, we provide complementary studies in Appendix~\ref{app:add_analysis}: our extracted skills' coverage and quality (\ref{app:skill-qual}), \evotd's sensitivity to the initial skill bank(\ref{app:evotd-sense}), the fidelity of synthesized crossover tasks(\ref{app:cross-align}), and the alignment between target skills and the actual solutions(\ref{app:skill-solution-align}). We also provide a cost analysis of our framework, including latency, API cost, and synthesis efficiency in Appendix~\ref{app:latency_cost}.




\section{Related Work}
\label{section:related-work}
\subsection{Evolutionary Algorithms in ML}

Recent work has revived evolutionary search to harness LLMs under verifiable feedback, transitioning from single-pass generation to iterative refinement \citep{novikov2025alphaevolve, surina2025algorithm}. Frameworks such as FunSearch \citep{romera2024mathematical} and AlphaEvolve \citep{novikov2025alphaevolve} demonstrate that LLM-guided evolution can discover novel mathematical constructions and optimize complex algorithmic artifacts. Beyond code, LLMs increasingly serve as evolutionary operators across diverse domains, from combinatorial and physical engineering optimization \citep{liu2024large, brahmachary2025large} to scientific discovery spaces like molecular search and biomedical hypothesis refinement \citep{wang2024efficient, gottweis2025towards}. Crucially, these prior approaches primarily employ evolution as an inference-time search mechanism to optimize specific candidate solutions or hypotheses. In contrast, our method shifts the evolutionary paradigm toward data synthesis: we evolve the training distribution itself to automatically construct a progressively rigorous reasoning curriculum.

\subsection{Data Synthesis}
Synthetic data has become indispensable for scaling post-training beyond the limits of human-authored supervision. Evol-Instruct family (e.g., WizardLM \citep{xu2024wizardlm}, WizardCoder \citep{luo2023wizardcoder}, WizardMath \citep{luo2023wizardmath}) scale diversity and difficulty through iterative, heuristic rewriting. Recent frameworks integrate stronger structural and behavioral feedback: Prismatic Synthesis aligns diversity with model behavior rather than surface form. DataEnvGym \citep{khan2025dataenvgymdatagenerationagents}, ACES~\citep{pourcel2024aces}, EvalTree \citep{zeng2025evaltree}, and REASONING GYM \citep{stojanovski2025reasoninggymreasoningenvironments} introduce weakness-aware or procedurally controlled generation. However, these approaches remain bottlenecked by rigid taxonomies, manually specified generators, or unstructured textual heuristics, lacking a unified mechanism to jointly modulate compositional logic and parametric complexity. By coupling combinatorial skill crossover with targeted complexity mutation, \evotd autonomously constructs a dynamic, capability-adaptive reasoning curriculum.

\subsection{Self-Improving LLM}
A parallel line of research advances self-improving LLMs by closing the loop between autonomous task generation and policy optimization. Frameworks such as Absolute Zero \citep{zhao2025absolute} and R-Zero \citep{huang2025r} demonstrate that models can bootstrap reasoning capabilities from zero external data via verifiable rewards and challenger–solver co-evolution. Similarly, SPIRAL \citep{liu2025spiral} casts self-improvement as multi-turn, zero-sum self-play, while Agent0 \citep{xia2025agent0} further augments curriculum generation with tool-integrated execution. While these paradigms successfully reduce reliance on human supervision, their innovations center primarily on interaction protocols (competition, self-play, or tool augmentation) leaving the underlying representation of the generated task space largely opaque. In contrast, \evotd imposes a rigorous structural prior on self-improvement. By evolving tasks across a transparent skill-complexity manifold, we construct a curriculum that is both frontier-seeking and interpretable.

\section{Conclusion}
\label{section:conclusion} 
We presented \evotd, a framework that defines task synthesis as a directed evolutionary search over a structured skill-complexity manifold. By formalizing \texttt{Attribute Mutation} and \texttt{Skill Crossover} operators, \evotd moves beyond heuristic prompting to systematically traverse the task space, effectively decoupling algorithmic logic from instance complexity. Empirical evaluations across code generation and frontier mathematical reasoning benchmarks show that our method establishes a formidable performance for reasoning models while generalizing across different architectures, regimes, and scales. Overall, our findings confirm that structural evolution provides a far more reliable training signal for complex reasoning than the unconstrained heuristic generation common in standard prompting.

\bibliographystyle{plainnat}
\bibliography{ref}

@article{guo2025deepseek,
  author       = {Daya Guo and
                  Dejian Yang and
                  Haowei Zhang and et al.},
  title        = {DeepSeek-R1 incentivizes reasoning in LLMs through reinforcement learning},
  journal      = {Nat.},
  volume       = {645},
  number       = {8081},
  pages        = {633--638},
  year         = {2025},
  url          = {https://doi.org/10.1038/s41586-025-09422-z},
  doi          = {10.1038/S41586-025-09422-Z},
  bibsource    = {dblp computer science bibliography, https://dblp.org}
}

@article{gemini2025gemini,
  author       = {Gemma Team},
  title        = {Gemma 3 Technical Report},
  journal      = {CoRR},
  volume       = {abs/2503.19786},
  year         = {2025},
  url          = {https://doi.org/10.48550/arXiv.2503.19786},
  doi          = {10.48550/ARXIV.2503.19786},
  eprinttype   = {arXiv},
  eprint       = {2503.19786},
  bibsource    = {dblp computer science bibliography, https://dblp.org}
}

@article{phan2025humanity,
  author       = {Long Phan and
                  Alice Gatti and
                  Ziwen Han and et al.},
  title        = {Humanity's Last Exam},
  journal      = {CoRR},
  volume       = {abs/2501.14249},
  year         = {2025},
  url          = {https://doi.org/10.48550/arXiv.2501.14249},
  doi          = {10.48550/ARXIV.2501.14249},
  eprinttype   = {arXiv},
  eprint       = {2501.14249},
  bibsource    = {dblp computer science bibliography, https://dblp.org}
}

@misc{aime2025,
  author = {{AIME}},
  title = {American Invitational Mathematics Examination (AIME) Problems and Solutions},
  year = {2025},
  howpublished = {\url{https://artofproblemsolving.com/wiki/index.php/AIME_Problems_and_Solutions}},
  note = {Accessed: 2025-05}
}

@article{lambert2024tulu,
  author       = {Nathan Lambert and
                  Jacob Morrison and
                  Valentina Pyatkin and et al.},
  title        = {T{\"{U}}LU 3: Pushing Frontiers in Open Language Model Post-Training},
  journal      = {CoRR},
  volume       = {abs/2411.15124},
  year         = {2024},
  url          = {https://doi.org/10.48550/arXiv.2411.15124},
  doi          = {10.48550/ARXIV.2411.15124},
  eprinttype   = {arXiv},
  eprint       = {2411.15124},
  bibsource    = {dblp computer science bibliography, https://dblp.org}
}

@article{cobbe2021trainingverifierssolvemath,
  author       = {Karl Cobbe and
                  Vineet Kosaraju and
                  Mohammad Bavarian and et al.},
  title        = {Training Verifiers to Solve Math Word Problems},
  journal      = {CoRR},
  volume       = {abs/2110.14168},
  year         = {2021},
  url          = {https://arxiv.org/abs/2110.14168},
  eprinttype   = {arXiv},
  eprint       = {2110.14168},
  bibsource    = {dblp computer science bibliography, https://dblp.org}
}

@article{moshkov2025aimo2winningsolutionbuilding,
  author       = {Ivan Moshkov and
                  Darragh Hanley and
                  Ivan Sorokin and et al.},
  title        = {{AIMO-2} Winning Solution: Building State-of-the-Art Mathematical
                  Reasoning Models with OpenMathReasoning dataset},
  journal      = {CoRR},
  volume       = {abs/2504.16891},
  year         = {2025},
  url          = {https://doi.org/10.48550/arXiv.2504.16891},
  doi          = {10.48550/ARXIV.2504.16891},
  eprinttype   = {arXiv},
  eprint       = {2504.16891},
  bibsource    = {dblp computer science bibliography, https://dblp.org}
}

@article{zhang2025surveyreinforcementlearninglarge,
  author       = {Kaiyan Zhang and
                  Yuxin Zuo and
                  Bingxiang He and et al.},
  title        = {A Survey of Reinforcement Learning for Large Reasoning Models},
  journal      = {CoRR},
  volume       = {abs/2509.08827},
  year         = {2025},
  url          = {https://doi.org/10.48550/arXiv.2509.08827},
  doi          = {10.48550/ARXIV.2509.08827},
  eprinttype   = {arXiv},
  eprint       = {2509.08827},
  bibsource    = {dblp computer science bibliography, https://dblp.org}
}

@article{jung2025prismaticsynthesisgradientbaseddata,
  author       = {Jaehun Jung and
                  Seungju Han and
                  Ximing Lu and et al.},
  title        = {Prismatic Synthesis: Gradient-based Data Diversification Boosts Generalization
                  in {LLM} Reasoning},
  journal      = {CoRR},
  volume       = {abs/2505.20161},
  year         = {2025},
  url          = {https://doi.org/10.48550/arXiv.2505.20161},
  doi          = {10.48550/ARXIV.2505.20161},
  eprinttype   = {arXiv},
  eprint       = {2505.20161},
  bibsource    = {dblp computer science bibliography, https://dblp.org}
}

@article{zhao2025absolute,
  author       = {Andrew Zhao and
                  Yiran Wu and
                  Yang Yue and et al.},
  title        = {Absolute Zero: Reinforced Self-play Reasoning with Zero Data},
  journal      = {CoRR},
  volume       = {abs/2505.03335},
  year         = {2025},
  url          = {https://doi.org/10.48550/arXiv.2505.03335},
  doi          = {10.48550/ARXIV.2505.03335},
  eprinttype   = {arXiv},
  eprint       = {2505.03335},
  bibsource    = {dblp computer science bibliography, https://dblp.org}
}

@inproceedings{xu2024wizardlm,
  author       = {Can Xu and
                  Qingfeng Sun and
                  Kai Zheng and et al.},
  title        = {WizardLM: Empowering Large Pre-Trained Language Models to Follow Complex
                  Instructions},
  booktitle    = {The Twelfth International Conference on Learning Representations,
                  {ICLR} 2024, Vienna, Austria, May 7-11, 2024},
  publisher    = {OpenReview.net},
  year         = {2024},
  url          = {https://openreview.net/forum?id=CfXh93NDgH},
  bibsource    = {dblp computer science bibliography, https://dblp.org}
}

@inproceedings{luo2023wizardcoder,
  author       = {Ziyang Luo and
                  Can Xu and
                  Pu Zhao and et al.},
  title        = {WizardCoder: Empowering Code Large Language Models with Evol-Instruct},
  booktitle    = {The Twelfth International Conference on Learning Representations,
                  {ICLR} 2024, Vienna, Austria, May 7-11, 2024},
  publisher    = {OpenReview.net},
  year         = {2024},
  url          = {https://openreview.net/forum?id=UnUwSIgK5W},
  bibsource    = {dblp computer science bibliography, https://dblp.org}
}

@inproceedings{luo2023wizardmath,
  author       = {Haipeng Luo and
                  Qingfeng Sun and
                  Can Xu and et al.},
  title        = {WizardMath: Empowering Mathematical Reasoning for Large Language Models
                  via Reinforced Evol-Instruct},
  booktitle    = {The Thirteenth International Conference on Learning Representations,
                  {ICLR} 2025, Singapore, April 24-28, 2025},
  publisher    = {OpenReview.net},
  year         = {2025},
  url          = {https://openreview.net/forum?id=mMPMHWOdOy},
  bibsource    = {dblp computer science bibliography, https://dblp.org}
}

@article{shumailov2024curserecursiontraininggenerated,
  author       = {Ilia Shumailov and
                  Zakhar Shumaylov and
                  Yiren Zhao and et al.},
  title        = {The Curse of Recursion: Training on Generated Data Makes Models Forget},
  journal      = {CoRR},
  volume       = {abs/2305.17493},
  year         = {2023},
  url          = {https://doi.org/10.48550/arXiv.2305.17493},
  doi          = {10.48550/ARXIV.2305.17493},
  eprinttype   = {arXiv},
  eprint       = {2305.17493},
  bibsource    = {dblp computer science bibliography, https://dblp.org}
}

@article{kim2025sedi,
  author       = {Jungwoo Kim and
                  Minsang Kim and
                  Sungjin Lee},
  title        = {SeDi-Instruct: Enhancing Alignment of Language Models through Self-Directed
                  Instruction Generation},
  journal      = {CoRR},
  volume       = {abs/2502.04774},
  year         = {2025},
  url          = {https://doi.org/10.48550/arXiv.2502.04774},
  doi          = {10.48550/ARXIV.2502.04774},
  eprinttype   = {arXiv},
  eprint       = {2502.04774},
  bibsource    = {dblp computer science bibliography, https://dblp.org}
}

@article{xia2025agent0,
  author       = {Peng Xia and
                  Kaide Zeng and
                  Jiaqi Liu and et al.},
  title        = {Agent0: Unleashing Self-Evolving Agents from Zero Data via Tool-Integrated
                  Reasoning},
  journal      = {CoRR},
  volume       = {abs/2511.16043},
  year         = {2025},
  url          = {https://doi.org/10.48550/arXiv.2511.16043},
  doi          = {10.48550/ARXIV.2511.16043},
  eprinttype   = {arXiv},
  eprint       = {2511.16043},
  bibsource    = {dblp computer science bibliography, https://dblp.org}
}

@article{huang2025r,
  author       = {Chengsong Huang and
                  Wenhao Yu and
                  Xiaoyang Wang and et al.},
  title        = {R-Zero: Self-Evolving Reasoning {LLM} from Zero Data},
  journal      = {CoRR},
  volume       = {abs/2508.05004},
  year         = {2025},
  url          = {https://doi.org/10.48550/arXiv.2508.05004},
  doi          = {10.48550/ARXIV.2508.05004},
  eprinttype   = {arXiv},
  eprint       = {2508.05004},
  bibsource    = {dblp computer science bibliography, https://dblp.org}
}

@inproceedings{wang2023self,
  author       = {Yizhong Wang and
                  Yeganeh Kordi and
                  Swaroop Mishra and et al.},
  editor       = {Anna Rogers and
                  Jordan L. Boyd{-}Graber and
                  Naoaki Okazaki},
  title        = {Self-Instruct: Aligning Language Models with Self-Generated Instructions},
  booktitle    = {Proceedings of the 61st Annual Meeting of the Association for Computational
                  Linguistics (Volume 1: Long Papers), {ACL} 2023, Toronto, Canada,
                  July 9-14, 2023},
  pages        = {13484--13508},
  publisher    = {Association for Computational Linguistics},
  year         = {2023},
  url          = {https://doi.org/10.18653/v1/2023.acl-long.754},
  doi          = {10.18653/V1/2023.ACL-LONG.754},
  bibsource    = {dblp computer science bibliography, https://dblp.org}
}

@article{shah2025aiassistedgenerationdifficultmath,
  author       = {Vedant Shah and
                  Dingli Yu and
                  Kaifeng Lyu and et al.},
  title        = {AI-Assisted Generation of Difficult Math Questions},
  journal      = {CoRR},
  volume       = {abs/2407.21009},
  year         = {2024},
  url          = {https://doi.org/10.48550/arXiv.2407.21009},
  doi          = {10.48550/ARXIV.2407.21009},
  eprinttype   = {arXiv},
  eprint       = {2407.21009},
  bibsource    = {dblp computer science bibliography, https://dblp.org}
}

@article{zeng2025evaltree,
  author       = {Zhiyuan Zeng and
                  Yizhong Wang and
                  Hannaneh Hajishirzi and
                  Pang Wei Koh},
  title        = {EvalTree: Profiling Language Model Weaknesses via Hierarchical Capability
                  Trees},
  journal      = {CoRR},
  volume       = {abs/2503.08893},
  year         = {2025},
  url          = {https://doi.org/10.48550/arXiv.2503.08893},
  doi          = {10.48550/ARXIV.2503.08893},
  eprinttype   = {arXiv},
  eprint       = {2503.08893},
  bibsource    = {dblp computer science bibliography, https://dblp.org}
}

@inproceedings{khan2025dataenvgymdatagenerationagents,
  author       = {Zaid Khan and
                  Elias Stengel{-}Eskin and
                  Jaemin Cho and
                  Mohit Bansal},
  title        = {DataEnvGym: Data Generation Agents in Teacher Environments with Student
                  Feedback},
  booktitle    = {The Thirteenth International Conference on Learning Representations,
                  {ICLR} 2025, Singapore, April 24-28, 2025},
  publisher    = {OpenReview.net},
  year         = {2025},
  url          = {https://openreview.net/forum?id=00SnKBGTsz},
  bibsource    = {dblp computer science bibliography, https://dblp.org}
}

@article{smith2012measuring,
  author       = {Kate Smith{-}Miles and
                  Leo Lopes},
  title        = {Measuring instance difficulty for combinatorial optimization problems},
  journal      = {Comput. Oper. Res.},
  volume       = {39},
  number       = {5},
  pages        = {875--889},
  year         = {2012},
  url          = {https://doi.org/10.1016/j.cor.2011.07.006},
  doi          = {10.1016/J.COR.2011.07.006},
  bibsource    = {dblp computer science bibliography, https://dblp.org}
}

@article{doctor2023toward,
  author       = {Katarina Doctor and
                  Christine Task and
                  Eric J. Kildebeck and et al.},
  title        = {Toward Defining a Domain Complexity Measure Across Domains},
  journal      = {CoRR},
  volume       = {abs/2303.04141},
  year         = {2023},
  url          = {https://doi.org/10.48550/arXiv.2303.04141},
  doi          = {10.48550/ARXIV.2303.04141},
  eprinttype   = {arXiv},
  eprint       = {2303.04141},
  bibsource    = {dblp computer science bibliography, https://dblp.org}
}

@article{holland1973genetic,
  author       = {John H. Holland},
  title        = {Genetic Algorithms and the Optimal Allocation of Trials},
  journal      = {{SIAM} J. Comput.},
  volume       = {2},
  number       = {2},
  pages        = {88--105},
  year         = {1973},
  url          = {https://doi.org/10.1137/0202009},
  doi          = {10.1137/0202009},
  bibsource    = {dblp computer science bibliography, https://dblp.org}
}

@book{vygotsky1978mind,
  title={Mind in society: The development of higher psychological processes},
  author={Vygotsky, Lev S},
  volume={86},
  year={1978},
  publisher={Harvard university press}
}

@inproceedings{didolkar2024metacognitive,
  author       = {Aniket Didolkar and
                  Anirudh Goyal and
                  Nan Rosemary Ke and et al.},
  editor       = {Amir Globersons and
                  Lester Mackey and
                  Danielle Belgrave and
                  Angela Fan and
                  Ulrich Paquet and
                  Jakub M. Tomczak and
                  Cheng Zhang},
  title        = {Metacognitive Capabilities of LLMs: An Exploration in Mathematical
                  Problem Solving},
  booktitle    = {Advances in Neural Information Processing Systems 38: Annual Conference
                  on Neural Information Processing Systems 2024, NeurIPS 2024, Vancouver,
                  BC, Canada, December 10 - 15, 2024},
  year         = {2024},
  url          = {http://papers.nips.cc/paper\_files/paper/2024/hash/2318d75a06437eaa257737a5cf3ab83c-Abstract-Conference.html},
  bibsource    = {dblp computer science bibliography, https://dblp.org}
}

@article{liu2025spiral,
  author       = {Bo Liu and
                  Leon Guertler and
                  Simon Yu and et al.},
  title        = {{SPIRAL:} Self-Play on Zero-Sum Games Incentivizes Reasoning via Multi-Agent
                  Multi-Turn Reinforcement Learning},
  journal      = {CoRR},
  volume       = {abs/2506.24119},
  year         = {2025},
  url          = {https://doi.org/10.48550/arXiv.2506.24119},
  doi          = {10.48550/ARXIV.2506.24119},
  eprinttype   = {arXiv},
  eprint       = {2506.24119},
  bibsource    = {dblp computer science bibliography, https://dblp.org}
}

@article{shao2024deepseekmath,
  author       = {Zhihong Shao and
                  Peiyi Wang and
                  Qihao Zhu and et al.},
  title        = {DeepSeekMath: Pushing the Limits of Mathematical Reasoning in Open
                  Language Models},
  journal      = {CoRR},
  volume       = {abs/2402.03300},
  year         = {2024},
  url          = {https://doi.org/10.48550/arXiv.2402.03300},
  doi          = {10.48550/ARXIV.2402.03300},
  eprinttype   = {arXiv},
  eprint       = {2402.03300},
  bibsource    = {dblp computer science bibliography, https://dblp.org}
}

@article{schulman2017proximal,
  author       = {John Schulman and
                  Filip Wolski and
                  Prafulla Dhariwal and
                  Alec Radford and
                  Oleg Klimov},
  title        = {Proximal Policy Optimization Algorithms},
  journal      = {CoRR},
  volume       = {abs/1707.06347},
  year         = {2017},
  url          = {http://arxiv.org/abs/1707.06347},
  eprinttype   = {arXiv},
  eprint       = {1707.06347},
  bibsource    = {dblp computer science bibliography, https://dblp.org}
}

@article{yu2025dapo,
  author       = {Qiying Yu and
                  Zheng Zhang and
                  Ruofei Zhu and et al.},
  title        = {{DAPO:} An Open-Source {LLM} Reinforcement Learning System at Scale},
  journal      = {CoRR},
  volume       = {abs/2503.14476},
  year         = {2025},
  url          = {https://doi.org/10.48550/arXiv.2503.14476},
  doi          = {10.48550/ARXIV.2503.14476},
  eprinttype   = {arXiv},
  eprint       = {2503.14476},
  bibsource    = {dblp computer science bibliography, https://dblp.org}
}

@inproceedings{luo2025tree,
  author       = {Ziyang Luo and
                  Kaixin Li and
                  Hongzhan Lin and et al.},
  editor       = {Wanxiang Che and
                  Joyce Nabende and
                  Ekaterina Shutova and
                  Mohammad Taher Pilehvar},
  title        = {Tree-of-Evolution: Tree-Structured Instruction Evolution for Code
                  Generation in Large Language Models},
  booktitle    = {Proceedings of the 63rd Annual Meeting of the Association for Computational
                  Linguistics (Volume 1: Long Papers), {ACL} 2025, Vienna, Austria,
                  July 27 - August 1, 2025},
  pages        = {297--316},
  publisher    = {Association for Computational Linguistics},
  year         = {2025},
  url          = {https://aclanthology.org/2025.acl-long.14/},
  bibsource    = {dblp computer science bibliography, https://dblp.org}
}

@article{orvalho2025large,
  author       = {Pedro Orvalho and
                  Marta Kwiatkowska},
  title        = {Are Large Language Models Robust in Understanding Code Against Semantics-Preserving
                  Mutations?},
  journal      = {CoRR},
  volume       = {abs/2505.10443},
  year         = {2025},
  url          = {https://doi.org/10.48550/arXiv.2505.10443},
  doi          = {10.48550/ARXIV.2505.10443},
  eprinttype   = {arXiv},
  eprint       = {2505.10443},
  bibsource    = {dblp computer science bibliography, https://dblp.org}
}

@article{zhang2025semantic,
  author       = {Chong Zhang and
                  Xiang Li and
                  Jia Wang and et al.},
  title        = {Semantic-Preserving Adversarial Attacks on LLMs: An Adaptive Greedy
                  Binary Search Approach},
  journal      = {CoRR},
  volume       = {abs/2506.18756},
  year         = {2025},
  url          = {https://doi.org/10.48550/arXiv.2506.18756},
  doi          = {10.48550/ARXIV.2506.18756},
  eprinttype   = {arXiv},
  eprint       = {2506.18756},
  bibsource    = {dblp computer science bibliography, https://dblp.org}
}

@article{novikov2025alphaevolve,
  author       = {Alexander Novikov and
                  Ng{\^{a}}n Vu and
                  Marvin Eisenberger and et al.},
  title        = {AlphaEvolve: {A} coding agent for scientific and algorithmic discovery},
  journal      = {CoRR},
  volume       = {abs/2506.13131},
  year         = {2025},
  url          = {https://doi.org/10.48550/arXiv.2506.13131},
  doi          = {10.48550/ARXIV.2506.13131},
  eprinttype   = {arXiv},
  eprint       = {2506.13131},
  bibsource    = {dblp computer science bibliography, https://dblp.org}
}

@article{abed2025increasing,
  author       = {Amal Abed and
                  Ivan Lukic and
                  J{\"{o}}rg K. H. Franke and
                  Frank Hutter},
  title        = {Increasing {LLM} Coding Capabilities through Diverse Synthetic Coding
                  Tasks},
  journal      = {CoRR},
  volume       = {abs/2510.23208},
  year         = {2025},
  url          = {https://doi.org/10.48550/arXiv.2510.23208},
  doi          = {10.48550/ARXIV.2510.23208},
  eprinttype   = {arXiv},
  eprint       = {2510.23208},
  bibsource    = {dblp computer science bibliography, https://dblp.org}
}

@inproceedings{van2025code,
  author       = {Niki van Stein and
                  Anna V. Kononova and
                  Lars Kotthoff and
                  Thomas B{\"{a}}ck},
  editor       = {Bogdan Filipic},
  title        = {Code Evolution Graphs: Understanding Large Language Model Driven Design
                  of Algorithms},
  booktitle    = {Proceedings of the Genetic and Evolutionary Computation Conference,
                  {GECCO} 2025, {NH} Malaga Hotel, Malaga, Spain, July 14-18, 2025},
  pages        = {943--951},
  publisher    = {{ACM}},
  year         = {2025},
  url          = {https://doi.org/10.1145/3712256.3726328},
  doi          = {10.1145/3712256.3726328},
  bibsource    = {dblp computer science bibliography, https://dblp.org}
}

@article{hemberg2024evolving,
  author       = {Erik Hemberg and
                  Stephen Moskal and
                  Una{-}May O'Reilly},
  title        = {Evolving code with a large language model},
  journal      = {Genet. Program. Evolvable Mach.},
  volume       = {25},
  number       = {2},
  pages        = {21},
  year         = {2024},
  url          = {https://doi.org/10.1007/s10710-024-09494-2},
  doi          = {10.1007/S10710-024-09494-2},
  bibsource    = {dblp computer science bibliography, https://dblp.org}
}

@article{piantadosi2016logical,
  title={The logical primitives of thought: Empirical foundations for compositional cognitive models.},
  author={Piantadosi, Steven T and Tenenbaum, Joshua B and Goodman, Noah D},
  journal={Psychological review},
  volume={123},
  number={4},
  pages={392},
  year={2016},
  publisher={American Psychological Association}
}

@article{lippl2025algorithmic,
  author       = {Samuel Lippl and
                  Thomas McGee and
                  Kimberly Lopez and et al.},
  title        = {Algorithmic Primitives and Compositional Geometry of Reasoning in
                  Language Models},
  journal      = {CoRR},
  volume       = {abs/2510.15987},
  year         = {2025},
  url          = {https://doi.org/10.48550/arXiv.2510.15987},
  doi          = {10.48550/ARXIV.2510.15987},
  eprinttype   = {arXiv},
  eprint       = {2510.15987},
  bibsource    = {dblp computer science bibliography, https://dblp.org}
}

@article{power2022grokking,
  author       = {Alethea Power and
                  Yuri Burda and
                  Harri Edwards and
                  Igor Babuschkin and
                  Vedant Misra},
  title        = {Grokking: Generalization Beyond Overfitting on Small Algorithmic Datasets},
  journal      = {CoRR},
  volume       = {abs/2201.02177},
  year         = {2022},
  url          = {https://arxiv.org/abs/2201.02177},
  eprinttype   = {arXiv},
  eprint       = {2201.02177},
  bibsource    = {dblp computer science bibliography, https://dblp.org}
}

@article{shi2024can,
  author       = {Quan Shi and
                  Michael Tang and
                  Karthik Narasimhan and
                  Shunyu Yao},
  title        = {Can Language Models Solve Olympiad Programming?},
  journal      = {CoRR},
  volume       = {abs/2404.10952},
  year         = {2024},
  url          = {https://doi.org/10.48550/arXiv.2404.10952},
  doi          = {10.48550/ARXIV.2404.10952},
  eprinttype   = {arXiv},
  eprint       = {2404.10952},
  bibsource    = {dblp computer science bibliography, https://dblp.org}
}

@article{qwen3technicalreport,
  author       = {Qwen Team},
  title        = {Qwen3 Technical Report},
  journal      = {CoRR},
  volume       = {abs/2505.09388},
  year         = {2025},
  url          = {https://doi.org/10.48550/arXiv.2505.09388},
  doi          = {10.48550/ARXIV.2505.09388},
  eprinttype   = {arXiv},
  eprint       = {2505.09388},
  bibsource    = {dblp computer science bibliography, https://dblp.org}
}

@inproceedings{sheng2025hybridflow,
  author       = {Guangming Sheng and
                  Chi Zhang and
                  Zilingfeng Ye and et al.},
  title        = {HybridFlow: {A} Flexible and Efficient {RLHF} Framework},
  booktitle    = {Proceedings of the Twentieth European Conference on Computer Systems,
                  EuroSys 2025, Rotterdam, The Netherlands, 30 March 2025 - 3 April
                  2025},
  pages        = {1279--1297},
  publisher    = {{ACM}},
  year         = {2025},
  url          = {https://doi.org/10.1145/3689031.3696075},
  doi          = {10.1145/3689031.3696075},
  bibsource    = {dblp computer science bibliography, https://dblp.org}
}

@article{singh2025openai,
  author       = {OpenAI},
  title        = {OpenAI {GPT-5} System Card},
  journal      = {CoRR},
  volume       = {abs/2601.03267},
  year         = {2026},
  url          = {https://doi.org/10.48550/arXiv.2601.03267},
  doi          = {10.48550/ARXIV.2601.03267},
  eprinttype   = {arXiv},
  eprint       = {2601.03267},
  bibsource    = {dblp computer science bibliography, https://dblp.org}
}

@article{austin2021program,
  author       = {Jacob Austin and
                  Augustus Odena and
                  Maxwell I. Nye and et al.},
  title        = {Program Synthesis with Large Language Models},
  journal      = {CoRR},
  volume       = {abs/2108.07732},
  year         = {2021},
  url          = {https://arxiv.org/abs/2108.07732},
  eprinttype   = {arXiv},
  eprint       = {2108.07732},
  bibsource    = {dblp computer science bibliography, https://dblp.org}
}

@inproceedings{jain2024livecodebenchholisticcontaminationfree,
  author       = {Naman Jain and
                  King Han and
                  Alex Gu and et al.},
  title        = {LiveCodeBench: Holistic and Contamination Free Evaluation of Large
                  Language Models for Code},
  booktitle    = {The Thirteenth International Conference on Learning Representations,
                  {ICLR} 2025, Singapore, April 24-28, 2025},
  publisher    = {OpenReview.net},
  year         = {2025},
  url          = {https://openreview.net/forum?id=chfJJYC3iL},
  bibsource    = {dblp computer science bibliography, https://dblp.org}
}

@inproceedings{he2024olympiadbench,
  author       = {Chaoqun He and
                  Renjie Luo and
                  Yuzhuo Bai and et al.},
  editor       = {Lun{-}Wei Ku and
                  Andre Martins and
                  Vivek Srikumar},
  title        = {OlympiadBench: {A} Challenging Benchmark for Promoting {AGI} with
                  Olympiad-Level Bilingual Multimodal Scientific Problems},
  booktitle    = {Proceedings of the 62nd Annual Meeting of the Association for Computational
                  Linguistics (Volume 1: Long Papers), {ACL} 2024, Bangkok, Thailand,
                  August 11-16, 2024},
  pages        = {3828--3850},
  publisher    = {Association for Computational Linguistics},
  year         = {2024},
  url          = {https://doi.org/10.18653/v1/2024.acl-long.211},
  doi          = {10.18653/V1/2024.ACL-LONG.211},
  bibsource    = {dblp computer science bibliography, https://dblp.org}
}

@inproceedings{lin2024mitigating,
  author       = {Yong Lin and
                  Hangyu Lin and
                  Wei Xiong and et al.},
  editor       = {Yaser Al{-}Onaizan and
                  Mohit Bansal and
                  Yun{-}Nung Chen},
  title        = {Mitigating the Alignment Tax of {RLHF}},
  booktitle    = {Proceedings of the 2024 Conference on Empirical Methods in Natural
                  Language Processing, {EMNLP} 2024, Miami, FL, USA, November 12-16,
                  2024},
  pages        = {580--606},
  publisher    = {Association for Computational Linguistics},
  year         = {2024},
  url          = {https://doi.org/10.18653/v1/2024.emnlp-main.35},
  doi          = {10.18653/V1/2024.EMNLP-MAIN.35},
  bibsource    = {dblp computer science bibliography, https://dblp.org}
}

@article{surina2025algorithm,
  author       = {Anja Surina and
                  Amin Mansouri and
                  Lars Quaedvlieg and et al.},
  title        = {Algorithm Discovery With LLMs: Evolutionary Search Meets Reinforcement
                  Learning},
  journal      = {CoRR},
  volume       = {abs/2504.05108},
  year         = {2025},
  url          = {https://doi.org/10.48550/arXiv.2504.05108},
  doi          = {10.48550/ARXIV.2504.05108},
  eprinttype   = {arXiv},
  eprint       = {2504.05108},
  bibsource    = {dblp computer science bibliography, https://dblp.org}
}

@article{romera2024mathematical,
  author       = {Bernardino Romera{-}Paredes and
                  Mohammadamin Barekatain and
                  Alexander Novikov and et al.},
  title        = {Mathematical discoveries from program search with large language models},
  journal      = {Nat.},
  volume       = {625},
  number       = {7995},
  pages        = {468--475},
  year         = {2024},
  url          = {https://doi.org/10.1038/s41586-023-06924-6},
  doi          = {10.1038/S41586-023-06924-6},
  bibsource    = {dblp computer science bibliography, https://dblp.org}
}

@inproceedings{liu2024large,
  author       = {Shengcai Liu and
                  Caishun Chen and
                  Xinghua Qu and
                  Ke Tang and
                  Yew{-}Soon Ong},
  title        = {Large Language Models as Evolutionary Optimizers},
  booktitle    = {{IEEE} Congress on Evolutionary Computation, {CEC} 2024, Yokohama,
                  Japan, June 30 - July 5, 2024},
  pages        = {1--8},
  publisher    = {{IEEE}},
  year         = {2024},
  url          = {https://doi.org/10.1109/CEC60901.2024.10611913},
  doi          = {10.1109/CEC60901.2024.10611913},
  bibsource    = {dblp computer science bibliography, https://dblp.org}
}

@article{brahmachary2025large,
  author       = {Shuvayan Brahmachary and
                  Subodh M. Joshi and
                  Aniruddha Panda and et al.},
  title        = {Large language model-based evolutionary optimizer: Reasoning with
                  elitism},
  journal      = {Neurocomputing},
  volume       = {622},
  pages        = {129272},
  year         = {2025},
  url          = {https://doi.org/10.1016/j.neucom.2024.129272},
  doi          = {10.1016/J.NEUCOM.2024.129272},
  bibsource    = {dblp computer science bibliography, https://dblp.org}
}

@inproceedings{wang2024efficient,
  author       = {Haorui Wang and
                  Marta Skreta and
                  Cher Tian Ser and et al.},
  title        = {Efficient Evolutionary Search Over Chemical Space with Large Language
                  Models},
  booktitle    = {The Thirteenth International Conference on Learning Representations,
                  {ICLR} 2025, Singapore, April 24-28, 2025},
  publisher    = {OpenReview.net},
  year         = {2025},
  url          = {https://openreview.net/forum?id=awWiNvQwf3},
  bibsource    = {dblp computer science bibliography, https://dblp.org}
}

@article{gottweis2025towards,
  author       = {Juraj Gottweis and
                  Wei{-}Hung Weng and
                  Alexander N. Daryin and et al.},
  title        = {Towards an {AI} co-scientist},
  journal      = {CoRR},
  volume       = {abs/2502.18864},
  year         = {2025},
  url          = {https://doi.org/10.48550/arXiv.2502.18864},
  doi          = {10.48550/ARXIV.2502.18864},
  eprinttype   = {arXiv},
  eprint       = {2502.18864},
  bibsource    = {dblp computer science bibliography, https://dblp.org}
}

@article{stojanovski2025reasoninggymreasoningenvironments,
  author       = {Zafir Stojanovski and
                  Oliver Stanley and
                  Joe Sharratt and et al.},
  title        = {{REASONING} {GYM:} Reasoning Environments for Reinforcement Learning
                  with Verifiable Rewards},
  journal      = {CoRR},
  volume       = {abs/2505.24760},
  year         = {2025},
  url          = {https://doi.org/10.48550/arXiv.2505.24760},
  doi          = {10.48550/ARXIV.2505.24760},
  eprinttype   = {arXiv},
  eprint       = {2505.24760},
  bibsource    = {dblp computer science bibliography, https://dblp.org}
}

@article{du2025supergpqa,
  author       = {M{-}A{-}P Team},
  title        = {SuperGPQA: Scaling {LLM} Evaluation across 285 Graduate Disciplines},
  journal      = {CoRR},
  volume       = {abs/2502.14739},
  year         = {2025},
  url          = {https://doi.org/10.48550/arXiv.2502.14739},
  doi          = {10.48550/ARXIV.2502.14739},
  eprinttype   = {arXiv},
  eprint       = {2502.14739},
  bibsource    = {dblp computer science bibliography, https://dblp.org}
}

@inproceedings{wang2024mmlu,
  author       = {Yubo Wang and
                  Xueguang Ma and
                  Ge Zhang and et al.},
  editor       = {Amir Globersons and
                  Lester Mackey and
                  Danielle Belgrave and
                  Angela Fan and
                  Ulrich Paquet and
                  Jakub M. Tomczak and
                  Cheng Zhang},
  title        = {MMLU-Pro: {A} More Robust and Challenging Multi-Task Language Understanding
                  Benchmark},
  booktitle    = {Advances in Neural Information Processing Systems 38: Annual Conference
                  on Neural Information Processing Systems 2024, NeurIPS 2024, Vancouver,
                  BC, Canada, December 10 - 15, 2024},
  year         = {2024},
  url          = {http://papers.nips.cc/paper\_files/paper/2024/hash/ad236edc564f3e3156e1b2feafb99a24-Abstract-Datasets\_and\_Benchmarks\_Track.html},
  bibsource    = {dblp computer science bibliography, https://dblp.org}
}

@inproceedings{pourcel2024aces,
  author       = {Julien Pourcel and
                  C{\'{e}}dric Colas and
                  Gaia Molinaro and et al.},
  editor       = {Amir Globersons and
                  Lester Mackey and
                  Danielle Belgrave and
                  Angela Fan and
                  Ulrich Paquet and
                  Jakub M. Tomczak and
                  Cheng Zhang},
  title        = {{ACES:} Generating a Diversity of Challenging Programming Puzzles
                  with Autotelic Generative Models},
  booktitle    = {Advances in Neural Information Processing Systems 38: Annual Conference
                  on Neural Information Processing Systems 2024, NeurIPS 2024, Vancouver,
                  BC, Canada, December 10 - 15, 2024},
  year         = {2024},
  url          = {http://papers.nips.cc/paper\_files/paper/2024/hash/7d0c6ff18f16797b92e77d7cc95b3c53-Abstract-Conference.html},
  bibsource    = {dblp computer science bibliography, https://dblp.org}
}

@article{li2025staying,
  author       = {Ziheng Li and
                  Zexu Sun and
                  Jinman Zhao and et al.},
  title        = {Staying in the Sweet Spot: Responsive Reasoning Evolution via Capability-Adaptive
                  Hint Scaffolding},
  journal      = {CoRR},
  volume       = {abs/2509.06923},
  year         = {2025},
  url          = {https://doi.org/10.48550/arXiv.2509.06923},
  doi          = {10.48550/ARXIV.2509.06923},
  eprinttype   = {arXiv},
  eprint       = {2509.06923},
  bibsource    = {dblp computer science bibliography, https://dblp.org}
}

@article{zhou2025expo,
  author       = {Ruiyang Zhou and
                  Shuozhe Li and
                  Amy Zhang and
                  Liu Leqi},
  title        = {ExPO: Unlocking Hard Reasoning with Self-Explanation-Guided Reinforcement
                  Learning},
  journal      = {CoRR},
  volume       = {abs/2507.02834},
  year         = {2025},
  url          = {https://doi.org/10.48550/arXiv.2507.02834},
  doi          = {10.48550/ARXIV.2507.02834},
  eprinttype   = {arXiv},
  eprint       = {2507.02834},
  bibsource    = {dblp computer science bibliography, https://dblp.org}
}

\appendix
\renewcommand{\thesection}{\Alph{section}}
\newpage

\begin{center}
	{\Large \textbf{Appendix for \evotd}}
\end{center}

\startcontents[sections]
\printcontents[sections]{l}{1}{\setcounter{tocdepth}{2}}

\section{Preliminaries}
\label{section:preliminaries}

We leverage recent advances in RL techniques for LLM finetuning. We highlight two approaches that underpin our framework's design below.
\subsection{Reinforcement Learning with Verifiable Rewards}
RLVR is a training paradigm which fine-tunes a policy LLM
$\pi_\theta$ using the reward that is collected from a deterministic verifier \citep{lambert2024tulu}.
The deterministic verifier $v : \mathcal{X} \to \{0,1\}$ evaluates each model generation $x_i$ against task-specific correctness criteria:
\vspace{-2pt}
\begin{equation}
r_i = v(x_i) = \begin{cases}
1, & \text{if } x_i \text{ passes verification,} \\
0, & \text{otherwise.}
\end{cases}
\vspace{-2pt}
\end{equation}

The policy is optimized to maximize the expected verification reward over the data distribution.
This binary feedback mechanism proves effective for structured reasoning domains where outcomes can be objectively validated through execution or formal checks.
We adopt this verifier-based reward paradigm as the foundation for training our model.

\subsection{Policy Optimization}
\paragraph{Group Relative Policy Optimization (GRPO)} GRPO \citep{shao2024deepseekmath} is a widely adopted policy gradient algorithm in RLVR that trains a policy $\pi_\theta$ without a separate, learned value function. It forms
advantages by comparing multiple samples from the same prompt.
Given a prompt $x$, sample a group $\{y_i\}_{i=1}^{G}$ from $\pi_{\theta_{\text{old}}}(\cdot\mid x)$ and compute
scalar rewards $\{r_i\}_{i=1}^{G}$.
GRPO normalizes rewards within the group to obtain a response-level advantage:
\vspace{-2pt}
\begin{equation}
\hat{A}_i \;=\; \frac{r_i - \mathrm{mean}(r_{1:G})}{\mathrm{std}(r_{1:G})+\epsilon_{\text{norm}}}.
\label{eq:grpo_adv}
\end{equation}

Its policy updates follow a PPO \citep{schulman2017proximal} style clipped surrogate objective with an explicit KL penalty to control policy
drift:
\begin{equation}
\begin{aligned}
\mathcal{L}_{\text{GRPO}}(\theta)
= &-\frac{1}{G}\sum_{i=1}^{G}
\min\!\Big(
\rho_i(\theta),\ 
\mathrm{clip}(\rho_i(\theta),1-\epsilon,1+\epsilon)
\Big) \\
&\cdot \hat{A}_i+ \beta\,\mathrm{KL}\!\left(
\pi_\theta(\cdot\mid x)\,\|\,\pi_{\theta_{\text{old}}}(\cdot\mid x)
\right)
\end{aligned}
\end{equation}
where $\rho_i(\theta)=\frac{\pi_\theta(y_i\mid x)}{\pi_{\theta_{\text{old}}}(y_i\mid x)}$.
Maximizing $-\mathcal{L}_{\text{GRPO}}$ increases the likelihood of samples with positive group-relative advantages while keeping updates conservative.

\paragraph{Dynamic sAmpling Policy Optimization (DAPO)}
DAPO~\citep{yu2025dapo} refines the GRPO framework via algorithmic innovations. 
It uses an asymmetric clip-higher mechanism, removes the KL penalty, adopts token level policy gradient loss, uses dynamic sampling, and applies overlong reward shaping. The token-level advantages $\{\hat{A}_t^i\}$ are computed as in Equation \ref{eq:grpo_adv}.
The objective averages clipped token-level updates across sampled outputs:
\begin{equation}
\mathcal{L}_{\text{DAPO}}(\theta)= \frac{1}{\sum_{i=1}^G |o^i|} \sum_{i=1}^{G} \sum_{t=1}^{|o^i|}
\min\!\Big(
r_t^i(\theta), \mathrm{clip}(r_t^i(\theta), 1-\epsilon_{\text{low}}, 1+\epsilon_{\text{high}})
\Big)\hat{A}_t^i
\end{equation}
where $r_t^i(\theta) = \frac{\pi_\theta(o_t^i \mid q, o_{<t}^i)}{\pi_{\theta_{\text{old}}}(o_t^i \mid q, o_{<t}^i)}$ is the token-level probability ratio, subject to the constraint that $0 < |\{o^i \mid \text{is\_equivalent}(a, o^i)\}| < G$.

\section{Dataset and Benchmark Details}
\label{app:dataset_details}

We provide detailed information about the benchmarks used in our experiments, including domain, variant type, and task count, complementing the high-level summary in Table~\ref{tab:appendix-dataset-summary}. The selected benchmarks span rigorous code generation and competition-style symbolic reasoning.
LiveCodeBench evaluates Python code generation under contamination-controlled conditions, while MBPP+ is a refactored variant of MBPP that increases difficulty through paraphrased prompts and stricter evaluation criteria.
AIME'24, AIME'25, and OlympiadBench assess symbolic and algorithmic reasoning in competitive settings with rigorous correctness requirements and limited problem counts.
To further probe broad multi-disciplinary knowledge and reasoning beyond code and math, we additionally evaluate on MMLU-Pro, which extends MMLU with reasoning-focused questions across 14 disciplines under a 10-option format, and SuperGPQA, which scales evaluation to 285 graduate-level disciplines spanning STEM, humanities, and professional fields.
\begin{table*}[!htbp]
\vspace{-5pt}
\centering
\small
\caption{Overview of benchmarks used for evaluation.}
\begin{tabular}{lcc}
\toprule
\textbf{Dataset} & \textbf{Domain} & \textbf{\# Tasks} \\
\midrule
LiveCodeBench$^{\text{v6}}$~\citep{jain2024livecodebenchholisticcontaminationfree} & Code Generation (Python) & 1055 \\
MBPP+~\citep{austin2021program} & Intro Programming (Python) & 500 \\
AIME'24~\citep{aime2025} & Symbolic Math (Mixed) & 30 \\
AIME'25~\citep{aime2025} & Symbolic Math (Mixed) & 30 \\
OlympiadBench~\citep{he2024olympiadbench} & Symbolic Math/Informatics & 675 \\
MMLU-Pro~\citep{wang2024mmlu} & Multi-discipline Knowledge \& Reasoning & 12032 \\
SuperGPQA~\citep{du2025supergpqa} & Graduate-level Multi-discipline & 26529 \\
\bottomrule
\end{tabular}
\label{tab:appendix-dataset-summary}
\end{table*}

\section{Additional Experiment Results}
\label{app:add_result}

\subsection{\evotd Performance on General Domain Benchmarks}
\label{app:bench_general_domain}
We further examine whether \evotd's improvements transfer beyond code and math to broader reasoning benchmarks. \evotd attains the highest average accuracy on both, lifting the Qwen3-4B Thinking backbone from 64.0\% to 67.1\% on MMLU-Pro and from 35.5\% to 37.4\% on SuperGPQA, exceeding Evol-Instruct (66.3\% and 36.0\%) in each case. The per-domain pattern sharpens the interpretation: on MMLU-Pro, \evotd's largest gains concentrate in STEM-adjacent domains in which abstract reasoning is the binding constraint --- Chemistry (+5.6), Physics (+5.4), Math (+4.1), Engineering (+3.7), and Computer Science (+3.1) over the base. The same pattern recurs on SuperGPQA, where \evotd's clearest gains arise on Science (+3.3) and Engineering (+2.3). These results establish that \evotd's reasoning gains generalize cleanly beyond its in-domain training: the curriculum lifts overall Pass@1 on both general-domain benchmarks, with its largest improvements concentrated precisely where rigorous reasoning is the binding constraint.
\begin{table*}[t!]
\centering
\normalsize
\caption{Pass@1 Results on Thinking models across general domain reasoning benchmarks.}

\begin{subtable}[t]{0.49\textwidth}
\centering

\resizebox{\textwidth}{!}{%
\begin{tabular}{l | c c >{\columncolor{rowblue}}c }
  \toprule[1.2pt]
  \textbf{Domain /\ Model} & Qwen3-4B & +Evol-Instruct & +\evotd \\
  \midrule[1.2pt]
  Engineering & 35.3 & \underline{36.1} & \bfseries 37.6 \\
  Medicine & \underline{36.0} & \bfseries 36.3 & 35.9 \\
  Science & 37.3 & \underline{37.9} & \bfseries 40.6 \\
  Philosophy & 41.9 & \underline{42.2} & \bfseries 42.7 \\
  Military Science & 35.6 & \underline{35.2} & \bfseries 35.8 \\
  Economics & \underline{43.8} & 43.7 & \bfseries 44.2 \\
  Management & \underline{37.0} & 36.1 & \bfseries 37.1 \\
  Sociology & \underline{32.4} & 32.3 & 33.2 \\
  Literature\&Arts & \bfseries 25.4 & 25.1 & \underline{25.3} \\
  History & \underline{22.4} & \bfseries 22.7 & 22.3 \\
  Agronomy & \underline{33.6} & \bfseries 34.4 & 33.5 \\
  Law & \bfseries 36.2 & 35.7 & \underline{35.8} \\
  Education & \bfseries 35.7 & \underline{35.4} & 35.2 \\
  \midrule[1.2pt]
  \textbf{AVG} & 35.5 & \underline{36.0} & \bfseries 37.4 \\
  \bottomrule[1.2pt]
\end{tabular}
}
\caption{SuperGPQA}
\end{subtable}
\hfill
\begin{subtable}[t]{0.49\textwidth}
\centering
\resizebox{\textwidth}{!}{%
\begin{tabular}{l | c c >{\columncolor{rowblue}}c }
  \toprule[1.2pt]
  \textbf{Domain /\ Model} & Qwen3-4B & +Evol-Instruct & +\evotd \\
  \midrule[1.2pt]
  Computer Science & 71.1 & \underline{74.0} & \bfseries 74.2 \\
  Math & 84.8 & \underline{87.7} & \bfseries 88.9 \\
  Chemistry & 69.7 & \underline{72.4} & \bfseries 75.3 \\
  Engineering & 44.0 & \underline{45.2} & \bfseries 47.7 \\
  Law & 30.0 & \bfseries 32.8 & \underline{32.6} \\
  Biology & 81.6 & \underline{83.8} & \bfseries 84.1 \\
  Health & 62.7 & \bfseries 64.8 & \underline{64} \\
  Physics & 69.9 & \underline{72.6} & \bfseries 75.3 \\
  Business & 70.5 & \underline{73.0} & \bfseries 74.3 \\
  Philosophy & 54.7 & \bfseries 55.9 & \underline{55.4} \\
  Economics & 74.7 & \bfseries 58.5 & \underline{58.2} \\
  Other & 56.5 & \bfseries 58.5 & \underline{58.2} \\
  Psychology & 68.1 & \bfseries 69.0 & \underline{68.3} \\
  History & 49.1 & \bfseries 51.8 & \underline{50.8} \\
  \midrule[1.2pt]
  \textbf{AVG} & 64.0 & \underline{66.3} & \bfseries 67.1 \\
  \bottomrule[1.2pt]
\end{tabular}
}
\caption{MMLU-Pro}
\end{subtable}
\label{tab:general_results}

\end{table*}

\subsection{Proposer Ablation}
\label{app:proposer_ablation}
To evaluate \evotd's sensitivity to the choice of proposer, we ablate our default configuration and evaluate the framework across a diverse suite of models, including both closed-source (o4-mini) and open-weight architectures (deepseek-r1-0528, gpt-oss-120B). We run the entire framework from scratch using each proposer, ranging from skill and complexity attribute extraction to task proposal. We then train Qwen3-4B Thinking under same amount (2,000) of synthesized tasks from each proposer. As shown in Table~\ref{tab:proposer_ablation}, performance gains remain consistent across proposers, with both Pass@1 and Pass@8 varying minimally across all evaluated benchmarks.

We attribute this proposer-agnostic stability to our multi-objective fitness check. Rather than relying on the proposer to generate flawless tasks on the first attempt, \evotd regularizes the generated candidates through strict executability, skill alignment, and ZDP learnability checks. This ensures that only high-fidelity, structurally sound tasks enter the evolutionary population. Consequently, the framework neutralizes quality differences across LLM proposers, yielding a robust and scalable data synthesis pipeline that does not rely on the strongest closed-source models.

\begin{table*}[htb!]
\caption{Ablation study on proposer models across code and math benchmarks.}
\centering
\small
\resizebox{\textwidth}{!}{%
\begin{tabular}{l |
                c c |
                c c c |
                c c c}
  \toprule[1.2pt]
    \textbf{Proposer Model} & LCB\textsuperscript{v6} & MBPP\textsuperscript{+} 
    & AME24 & AME25 & Olympiad 
    & \textbf{CAvg} & \textbf{MAvg} & \textbf{AVG} \\
    \midrule[1pt]
    \rowhighlight
    \multicolumn{9}{c}{\textit{Qwen3-4B Thinking Pass@1}} \\
    \midrule[1pt]
    
    o4-mini 
    & 46.8 & 54.7 & 36.7 & 25.8 & 47.3 & 50.8 & 36.6 & 42.3 \\
    deepseek-r1-0528 
    & 47.5 & 55.1 & 37.3 & 23.8 & 48.3 & 51.3 & 36.5 & 42.4 \\
    gpt-oss-120B 
    & 46.7 & 55.2 & 37.4 & 25.1 & 48.2 & 51.0 & 36.9 & 42.5 \\

    \midrule[1.2pt]
    \rowhighlight
    \multicolumn{9}{c}{\textit{Qwen3-4B Thinking Pass@8}} \\
    \midrule[0.8pt]

    o4-mini 
    & 58.9 & 71.2 & 60.2 & 47.3 & 59.7 & 65.1 & 55.7 & 59.5 \\
    deepseek-r1-0528 
    & 59.5 & 70.1 & 54.7 & 44.2 & 60.7 & 64.8 & 53.2 & 57.8 \\
    gpt-oss-120B 
    & 58.8 & 71.4 & 56.3 & 48.4 & 61.5 & 65.1 & 55.4 & 59.3 \\

  \bottomrule[1.2pt]
\end{tabular}
}
\label{tab:proposer_ablation}
\end{table*}

\section{Additional Analysis}
\label{app:add_analysis}

\subsection{\evotd's Sensitivity to Skill Bank}
\label{app:evotd-sense}
While we previously established the high fidelity and coverage of our extracted skill bank (Section \ref{app:skill-qual}), a critical question remains: how sensitive is \evotd to a degraded or incomplete initial ontology? To investigate this, we randomly remove 50\% of the extracted skills (reducing the skill bank from 56 to 28) prior to the task generation phase. We then execute a single iteration of the evolutionary pipeline and train the Qwen3-4B Thinking model. As detailed in Table \ref{tab:half_skill_ablation}, this removal induces a uniform degradation in performance, lowering average Pass@1 by 1.1. This confirms that the framework's effectiveness is correlated with the breadth and quality of the initial skill bank. However, this ablation also highlights the robustness and sample efficiency of our evolutionary search mechanism. Even when bottlenecked by a constrained skill space, the single-iteration \evotd achieves competitive performances with three iterations Evol-Instruct (average 41.2 Pass@1).

\begin{table*}[hbt!]
\caption{\evotd's sensitivity to initial skill bank for Qwen3-4B Thinking.}
\vspace{-0.3em}
\centering
\small
\resizebox{\textwidth}{!}{%
\begin{tabular}{l |
                c c |
                c c c |
                c c c}
  \toprule[1.2pt]
    \textbf{Method} & LCB\textsuperscript{v6} & MBPP\textsuperscript{+} 
    & AME24 & AME25 & Olympiad 
    & \textbf{CAvg} & \textbf{MAvg} & \textbf{AVG} \\
    \midrule[1.2pt]
    \rowhighlight
    \multicolumn{9}{c}{\textit{Qwen3-4B Thinking Pass@1}} \\
    \midrule[1.2pt]
    \evotd (One Iteration) \\
    \quad +Full Skills & 46.8 & 54.7 & 36.7 & 25.8 & 47.3 & 50.8 & 36.6 & 42.3 \\
    \quad +Half Skills & 45.9 & 54.3 & 34.4 & 24.3 & 46.9 & 50.1 & 35.2 & 41.2 \\
    Evol-Instruct (Three Iteration) & 47.4 & 53.7 & 35.4 & 22.4 & 46.9 & 50.6 & 34.9 & 41.2 \\
    \midrule[1.2pt]
    \rowhighlight
    \multicolumn{9}{c}{\textit{Qwen3-4B Thinking Pass@8}} \\
    \midrule[1.2pt]
    \evotd (One Iteration) \\
    \quad +Full Skills & 58.9 & 70.9 & 60.2 & 47.3 & 59.7 & 64.9 & 55.7 & 59.4 \\
    \quad +Half Skills & 57.4 & 72.0 & 55.2 & 44.6 & 58.8 & 64.7 & 52.9 & 57.6 \\
    Evol-Instruct (Three Iteration) & 58.0 & 54.8 & 55.0 & 44.2 & 59.1 & 56.4 & 52.8 & 54.2 \\
  \bottomrule[1.2pt]
\end{tabular}
}
\label{tab:half_skill_ablation}
\end{table*}

\subsection{Skill Coverage and Quality}
\label{app:skill-qual}
As algorithmic skills are an important abstraction in our framework, it is critical to validate the quality and coverage of our extracted skill bank. Because there exists no universal gold standard for algorithmic skill ontologies, we validate our extraction by aligning it against the established, human-curated taxonomies from LeetCode\footnotemark[1] and Codeforces\footnotemark[2] (See full mappings in Table \ref{tab:leetcode_map} and \ref{tab:codeforces_map}). 
\footnotetext[1]{\href{https://leetcode.com}{https://leetcode.com}}
\footnotetext[2]{\href{https://codeforces.com}{https://codeforces.com}} As these official platforms frequently conflate canonical algorithmic concepts with broad or format-specific labels (e.g., ``implementation'', ``interactive''), we report a more informative metric of reverse alignment: the proportion of our skills that successfully map back to recognized platform tags. \textbf{91.1\%} of our extracted skills map directly to LeetCode, and \textbf{100.0\%} map to Codeforces, confirming that the LLM extraction captures authentic reasoning primitives rather than hallucinating arbitrary concepts. Furthermore, our automated extraction demonstrates a \textbf{1.87}$\boldsymbol{\times}$ and \textbf{1.47}$\boldsymbol{\times}$ increase in fine-grainedness over the LeetCode and Codeforces taxonomies, respectively. This fine-grained skill ``resolution'' (e.g. decomposing a coarse ``Dynamic Programming'' tag into precise skills like \texttt{tree\_dp} or \texttt{bitmask\_dp}) is a structural advantage. It yields the exact building blocks necessary for our \texttt{Skill Crossover} operator to compose novel reasoning tasks without relying on overly broad or ambiguous skills.

\begin{table*}[htb!]
\centering
\scriptsize
\setlength{\tabcolsep}{4pt}
\caption{Mapping from LeetCode tags to our skill bank.}
\renewcommand{\arraystretch}{0.5}
\begin{tabularx}{\textwidth}{>{\raggedright\arraybackslash}p{0.21\textwidth} >{\raggedright\arraybackslash}X}
\toprule
\textbf{LeetCode tag} & \textbf{Our matching skills} \\
\midrule
Array & prefix\_sum, difference\_array, range\_sum\_query \\
String & edit\_distance, palindrome\_expansion \\
Dynamic Programming & dp\_1d, knapsack\_dp, grid\_dp, tree\_dp, dp\_on\_dag, bitmask\_dp, edit\_distance, floyd\_warshall \\
Math & gcd, integer\_division, modular\_arithmetic, sieve\_of\_eratosthenes, prime\_factorization, divisor\_enumeration, convex\_hull, cross\_product, polygon\_area, rectangle\_union\_area, point\_in\_triangle, distance\_computation \\
Two Pointers & two\_pointers \\
Backtracking & brute\_force\_enumeration \\
Hash Table & hash\_table \\
Linked List & --- \\
Matrix & grid\_dp \\
Recursion & --- \\
Sorting & sorting, sweep\_line \\
Binary Search & binary\_search, parametric\_search \\
Stack & --- \\
Tree & tree\_dp \\
Binary Tree & --- \\
Bit Manipulation & prefix\_xor, range\_xor\_query, bitmask\_dp \\
Simulation & simulation \\
Depth-First Search & dfs, topological\_sort \\
Greedy & interval\_scheduling, graph\_coloring\_greedy, coupon\_selection\_greedy, fuel\_station\_greedy \\
Binary Search Tree & --- \\
Sliding Window & sliding\_window \\
Divide and Conquer & meet\_in\_the\_middle \\
Monotonic Stack & --- \\
Trie & --- \\
Heap (Priority Queue) & priority\_queue, dijkstra, prim \\
Merge Sort & --- \\
String Matching & kmp\_failure\_function, rolling\_hash \\
Combinatorics & binomial\_coefficients, inclusion\_exclusion \\
Memoization & --- \\
Breadth-First Search & bfs, zero\_one\_bfs \\
\bottomrule
\end{tabularx}
\label{tab:leetcode_map}
\end{table*}
\begin{table*}[hbt!]
\centering
\scriptsize
\setlength{\tabcolsep}{4pt}
\caption{Mapping from official Codeforces tags to our skill bank.}
\renewcommand{\arraystretch}{0.5}
\begin{tabularx}{\textwidth}{>{\raggedright\arraybackslash}p{0.21\textwidth} >{\raggedright\arraybackslash}X}
\toprule
\textbf{Codeforces tag} & \textbf{Our matching skills} \\
\midrule
greedy & interval\_scheduling, graph\_coloring\_greedy, coupon\_selection\_greedy, fuel\_station\_greedy \\
math & gcd, integer\_division, modular\_arithmetic, sieve\_of\_eratosthenes, prime\_factorization, divisor\_enumeration, binomial\_coefficients \\
implementation & simulation \\
dp & dp\_1d, knapsack\_dp, grid\_dp, tree\_dp, dp\_on\_dag, bitmask\_dp, edit\_distance, floyd\_warshall \\
constructive algorithms & --- \\
data structures & prefix\_sum, difference\_array, range\_sum\_query, fenwick\_tree, segment\_tree, priority\_queue, multiset \\
brute force & brute\_force\_enumeration \\
binary search & binary\_search, parametric\_search \\
sortings & sorting, sweep\_line, mst\_kruskal \\
graphs & bfs, topological\_sort, prim, mst\_kruskal \\
dfs and similar & dfs \\
trees & tree\_dp \\
number theory & gcd, modular\_arithmetic, sieve\_of\_eratosthenes, prime\_factorization, divisor\_enumeration \\
combinatorics & binomial\_coefficients, inclusion\_exclusion \\
strings & kmp\_failure\_function, rolling\_hash, edit\_distance, palindrome\_expansion \\
bitmasks & bitmask\_dp, prefix\_xor, range\_xor\_query \\
two pointers & two\_pointers, sliding\_window \\
*special & --- \\
geometry & convex\_hull, cross\_product, polygon\_area, rectangle\_union\_area, point\_in\_triangle, distance\_computation, sweep\_line \\
dsu & union\_find, mst\_kruskal \\
divide and conquer & meet\_in\_the\_middle \\
interactive & --- \\
shortest paths & dijkstra, zero\_one\_bfs, floyd\_warshall \\
games & --- \\
probabilities & --- \\
hashing & hash\_table, rolling\_hash \\
flows & --- \\
matrices & grid\_dp \\
fft & --- \\
graph matchings & --- \\
string suffix structures & --- \\
ternary search & --- \\
meet-in-the-middle & meet\_in\_the\_middle \\
expression parsing & --- \\
2-sat & --- \\
chinese remainder theorem & --- \\
schedules & interval\_scheduling \\
communication & --- \\
\bottomrule
\end{tabularx}
\label{tab:codeforces_map}
\end{table*}

\subsection{Crossover Task Fidelity}
\label{app:cross-align}
To verify that our \texttt{Skill Crossover} operator synthesizes genuinely synergistic tasks rather than superficial concatenations, we evaluate compositional fidelity using an LLM-as-a-Judge (\texttt{gpt-5-mini}). The judge rigorously audits structural interdependence (See prompt in Figure \ref{fig:cross_task_verify}), explicitly rejecting tasks if they resemble isolated sub-problems, utilize skills decoratively, or include targeted skills that can be dropped without altering solvability. Under these strict criteria, \textbf{93\%} (313 of 337) of the synthesized tasks are verified to possess deeply interconnected, non-trivial skill combinations. This confirms that our regularized crossover design effectively prevents combinatorial degeneration, faithfully yielding authentic, multi-skill reasoning tasks.

\begin{tcolorbox}[
  colback=gray!5!white,
  colframe=gray!75!black,
  title=\bfseries Compositional Synergy Evaluation Prompt,
  width=\textwidth,
  boxrule=0.8pt,
  arc=4pt,
  outer arc=4pt,
  boxsep=2pt,
  left=2pt,
  right=2pt,
  top=2pt,
  bottom=2pt,
  breakable,
  fontupper=\scriptsize
]

\setlength{\parskip}{0.2em}
\setlength{\parindent}{0pt}

\newcommand{\PartHeader}[1]{%
  \smallskip
  {\footnotesize\bfseries\scshape #1}%
  \par\smallskip
}

\setlist[itemize]{%
  leftmargin=1.4em,
  itemsep=0.2em,
  topsep=0pt,
  parsep=0pt,
  partopsep=0pt
}
\setlist[enumerate,1]{%
  leftmargin=1.9em,
  labelwidth=1.2em,
  labelsep=0.4em,
  align=left,
  itemsep=0.2em,
  topsep=0pt,
  parsep=0pt,
  partopsep=0pt
}
\PartHeader{System Role}

\begin{adjustwidth}{0.6em}{0pt}
You are an expert algorithm analyst. You answer every question with a single word: either Yes or No.
\end{adjustwidth}

\PartHeader{Task Objective}

\begin{adjustwidth}{0.6em}{0pt}
Your task is to determine whether the synthesized task meaningfully integrates all of the specified algorithmic skills.
\end{adjustwidth}

\PartHeader{Input Variables}

\begin{adjustwidth}{0.6em}{0pt}
You will be given:
\begin{enumerate}
    \item A synthesized task: \texttt{\{task\}}
    \item A list of target algorithmic skills: \texttt{\{skills\}}
    \item A description for each skill: \texttt{\{skill\_descriptions\}}
\end{enumerate}
\end{adjustwidth}

\PartHeader{Evaluation Rules}

\begin{adjustwidth}{0.6em}{0pt}
\begin{itemize}
    \item Answer ``Yes'' only if all listed skills are substantively involved in solving the task and interact in a unified way.
    \item Answer ``No'' if the task looks like a glued combination of skills rather than a coherent multi-skill problem.
    \item Answer ``No'' if any listed skill appears superficial, weakly involved, decorative, or droppable without substantially changing the task.
    \item Ignore superficial wording or explicit mentions of skills in the task description.
    \item Focus only on whether the core solving process genuinely requires the specified skills to work together.
\end{itemize}
\end{adjustwidth}

\PartHeader{Output Format}

\begin{adjustwidth}{0.6em}{0pt}
Does this task meaningfully integrate all of the specified skills?

Respond with only one word:\\
\texttt{Yes} \quad or \quad \texttt{No}
\end{adjustwidth}

\end{tcolorbox}
\vspace{-2pt}
\captionof{figure}{Compositional Synergy Evaluation Prompt}
\label{fig:cross_task_verify}
\subsection{Target Skill and Actual Solution Alignment}
\label{app:skill-solution-align}
One critical aspect of targeted data synthesis involves aligning the solver's behavior with the intended task logic, ensuring it employs the targeted algorithmic skills rather than exploiting alternative heuristics. Within \evotd, this solver-skill alignment is structurally guaranteed for Deduction ($T_{\text{ded}}$) and Abduction ($T_{\text{abd}}$) tasks, as the model reasons directly over provided code snippets where the targeted logic is already explicitly embedded. Induction tasks ($T_{\text{ind}}$) require \textit{de novo} code generation and are thus theoretically susceptible to alternative solution paths. However, their problem statement and I/O pairs are strictly derived from verified reference implementations, establishing a strong prior toward the intended logic. To empirically quantify this consistency, we employed an LLM-as-a-Judge (\texttt{gpt-5-mini}) to audit correct solver rollouts for Induction tasks (See prompt in Figure \ref{fig:solver_skill_align}). \textbf{87.4\%} (1,371 of 1,568) of successful solutions authentically implementing the targeted skills, confirming the robust target skill and solution alignment.

\begin{tcolorbox}[
  colback=gray!5!white,
  colframe=gray!75!black,
  title=\bfseries Solver-Skill Alignment Evaluation Prompt,
  width=\textwidth,
  boxrule=0.8pt,
  arc=4pt,
  outer arc=4pt,
  boxsep=2pt,
  left=2pt,
  right=2pt,
  top=2pt,
  bottom=2pt,
  breakable,
  fontupper=\scriptsize
]

\setlength{\parskip}{0.2em}
\setlength{\parindent}{0pt}

\newcommand{\PartHeader}[1]{%
  \smallskip
  {\footnotesize\bfseries\scshape #1}%
  \par\smallskip
}

\setlist[itemize]{%
  leftmargin=1.4em,
  itemsep=0.2em,
  topsep=0pt,
  parsep=0pt,
  partopsep=0pt
}
\setlist[enumerate,1]{%
  leftmargin=1.9em,
  labelwidth=1.2em,
  labelsep=0.4em,
  align=left,
  itemsep=0.2em,
  topsep=0pt,
  parsep=0pt,
  partopsep=0pt
}

\PartHeader{System Role}

\begin{adjustwidth}{0.6em}{0pt}
You are an expert algorithm analyst. You answer every question with a single word: either Yes or No.
\end{adjustwidth}

\PartHeader{Task Objective}

\begin{adjustwidth}{0.6em}{0pt}
Your task is to determine whether the code meaningfully implements or applies all of the specified skills.
\end{adjustwidth}

\PartHeader{Input Variables}

\begin{adjustwidth}{0.6em}{0pt}
You will be given:
\begin{enumerate}
    \item A Python code snippet: \texttt{\{code\}}
    \item A list of algorithmic skills: \texttt{\{skills\}}
    \item A description for each skill: \texttt{\{skill\_descriptions\}}
\end{enumerate}
\end{adjustwidth}

\PartHeader{Evaluation Rules}

\begin{adjustwidth}{0.6em}{0pt}
\begin{itemize}
    \item Answer ``Yes'' only if every listed skill is genuinely used in the code's core algorithmic logic.
    \item Answer ``No'' if even one listed skill is not meaningfully applied.
    \item Answer ``No'' if the code uses a different approach, even if the final output is correct.
    \item Ignore helper boilerplate such as input/output handling, type annotations, wrappers, or formatting details.
    \item Focus only on whether the essential reasoning or computation in the code depends on all specified skills.
\end{itemize}
\end{adjustwidth}

\PartHeader{Output Format}

\begin{adjustwidth}{0.6em}{0pt}
Respond with only one word:\\
\texttt{Yes} \quad or \quad \texttt{No}
\end{adjustwidth}

\end{tcolorbox}
\vspace{-2pt}
\captionof{figure}{Solver-Skill Alignment Evaluation Prompt}
\label{fig:solver_skill_align}

\subsection{Evolutionary Operators Ablation}
\label{app:evo-op-ablation}
Table \ref{tab:full_ablation} presents the fine-grained performance breakdown across individual benchmarks for the ablation study. This detailed view reveals distinct functional roles for each operator:
\begin{itemize}[topsep=0pt, itemsep=0.2pt, leftmargin=1.55em]
\item \textbf{Attribute Mutation for Hard Code \& Math:} Removing \texttt{Attribute Mutation} causes the most significant drops in LiveCodeBench (-1.0\%) and AIME 24 (-5.0\%). In contrast, MBPP+ (a simpler coding benchmark less reliant on complex reasoning) remains largely unaffected. This is expected: tasks requiring multi-step complex reasoning benefit most from fine-grained difficulty scaling, which is precisely what \texttt{Attribute Mutation} provides.
\item \textbf{Skill Crossover for Reasoning Transfer:} While the model without \texttt{Skill Crossover} maintains strong performance on standard coding tasks (e.g., MBPP+), it suffers on abstract reasoning benchmarks. This suggests that while simple coding skills can be learned via isolation, high-level mathematical reasoning (AIME/Olympiad) requires the diverse combinatorial exposure.
\end{itemize}
\begin{table*}[htb!]
\caption{Detailed performance breakdown on effectiveness of Attribute Mutation and Skill Crossover}
\centering
\small
\resizebox{\textwidth}{!}{%
\begin{tabular}{l |
                c c |
                c c c |
                c c c}
  \toprule[1.2pt]
    \textbf{Model} & LCB\textsuperscript{v6} & MBPP\textsuperscript{+}
    & AME24 & AME25 & Olympiad
    & \textbf{CAvg} & \textbf{MAvg} & \textbf{AVG} \\
    \midrule[1.2pt]
    \rowhighlight
    \multicolumn{9}{c}{Qwen3-4B Thinking \textit{Pass@1}} \\
    \midrule[1.2pt]

    \rowcolor{rowblue}
    \evotd
    & \bfseries 49.8 & 55.8 & \bfseries 42.2 & \bfseries 29.0 & \bfseries 49.4 & \underline{52.8} & \bfseries 40.2 & \bfseries 45.2 \\
    \quad w/o Attribute Mutation
    & 48.8 & \underline{56.5} & 37.2 & \underline{28.6} & 48.6 & 52.7 & 38.1 & 43.9 \\
    \quad w/o Skill Crossover
    & 49.1 & \bfseries 57.0 & \underline{39.1} & 27.4 & \underline{49.2} & \bfseries 53.1 & \underline{38.6} & \underline{44.4} \\

    \midrule[1.2pt]
    \rowhighlight
    \multicolumn{9}{c}{Qwen3-4B Thinking \textit{Pass@8}} \\
    \midrule[1.2pt]

    \rowcolor{rowblue}
    \evotd
    & \bfseries 61.5 & 71.2 & \bfseries 63.2 & \underline{49.9} & \underline{62.1} & \underline{66.4} & \bfseries 58.4 & \bfseries 61.6 \\
    \quad w/o Attribute Mutation
    & \underline{60.9} & \underline{71.7} & \underline{60.9} & 48.7 & 60.1 & 66.3 & 56.6 & 60.5 \\
    \quad w/o Skill Crossover
    & 60.5 & \bfseries 73.8 & 59.2 & \bfseries 50.0 & \bfseries 62.2 & \bfseries 67.2 & \underline{57.1} & \underline{61.1} \\

  \bottomrule[1.2pt]
\end{tabular}
}
\label{tab:full_ablation}
\end{table*}

\section{Computing Infrastructure}
\label{app:computing}
We conduct our experiments on a server equipped with 4 NVIDIA H200 (141GB VRAM) GPUs. We utilize the NVIDIA CUDA toolkit version 13.0. All experiments are implemented using Python 3.10.19, the PyTorch framework version 2.9.0, and the Transformer library version 4.57.3.

\section{Cost Analysis}
\label{app:latency_cost}

\paragraph{Latency} To quantify the operational overhead of our framework, we analyze the end-to-end wall-clock compute breakdown for both the Qwen3-4B Base and Qwen3-4B Thinking models (Table \ref{tab:latency}). This profiling demonstrates that our evolutionary data synthesis process (encompassing task proposal, skill alignment, and the learnability check) is highly efficient, accounting for only 13.5\% of the total runtime for the Thinking model and 16.1\% for the Base model.

\paragraph{API Cost} The API cost for o4-mini, gpt-oss-120b, and deepseek-r1-0528 per iteration synthesis is ~\$100, ~\$120, and ~\$110, respectively.

\paragraph{Proposal Efficiency} To ensure a fair comparison, we standardize the training set size for both \evotd and all baselines at approximately 6,000 samples across three iterations. In each iteration, \evotd generates roughly 3,142 candidates to yield 2,000 valid samples after multi-objective verification (\texttt{discard\_rate}=36.3\%). In contrast, Evol-Instruct requires approximately 5,463 candidates to secure the same volume of valid data (\texttt{discard\_rate}=63.4\%). This demonstrates that \evotd is a significantly more sample-efficient method for synthesizing diverse curricula.
\begin{table}[htpb]
\centering
\caption{Wall-clock compute breakdown for \evotd.}
\label{tab:latency}
\begin{tabular}{l c c c c}
\toprule
\textbf{Component} & \textbf{Qwen3-4B Thinking} & \textbf{Fraction} & \textbf{Qwen3-4B Base} & \textbf{Fraction} \\
\midrule
Task Proposal & 1.6 hr & 2.8\% & 1.6 hr & 6.1\% \\
Skill Alignment & 0.7 hr & 1.2\% & 0.7 hr & 2.7\% \\
Learnability & 5.5 hr & 9.5\% & 1.9 hr & 7.3\% \\
RLVR Training & 49.9 hr & 86.5\% & 21.9 hr & 83.9\% \\
\bottomrule
\end{tabular}
\end{table}

\section{Implementation Details}
\label{app:exp_details}
\paragraph{Hyperparameters} Table \ref{tab:base_train_hp} and \ref{tab:think_train_hp} present the training hyperparameters we use in Base model and Thinking model RL training, respectively. Table \ref{tab:base_eval_hp} and \ref{tab:think_eval_hp} show the evaluation hyperparameter configuration for Base and Thinking models.

\begin{table}[htbp]
\centering
\label{tab:training_hp}

\begin{minipage}[t]{0.49\linewidth}

\centering
\captionof{table}{Qwen3-4B Thinking Training Hyperparameters}
\resizebox{\linewidth}{!}{%
\begin{tabular}{|l|c|}
\hline
\textbf{Parameter}                  & \textbf{Value}          \\ \hline
\multicolumn{2}{|c|}{\textbf{Model Configuration}}  \\ \hline
Max Prompt Length                   & 6144                    \\ \hline
Max Response Length                 & 8096                    \\ \hline
Seed Batch Factor                   & 42                      \\ \hline
\multicolumn{2}{|c|}{\textbf{Training Settings}}    \\ \hline
Train Batch Size                    & 64                      \\ \hline
Generation Batch Size               & 88                      \\ \hline
Learning Rate                       & 1e-6                    \\ \hline
Optimizer                           & AdamW                   \\ \hline
Weight Decay                        & 0.01                    \\ \hline
Grad Clip                           & 1.0                     \\ \hline
Total Steps                         & 200                     \\ \hline
\multicolumn{2}{|c|}{\textbf{RL Settings}}          \\ \hline
Adv. Estimator                      & DAPO                    \\ \hline
KL Loss                             & 0.0                     \\ \hline
KL Reward                           & 0.0                     \\ \hline
Entropy Coefficient                 & 0.0                     \\ \hline
PPO Epochs                          & 16                      \\ \hline
$N$ Rollouts                        & 16                      \\ \hline
Rollout Temperature                 & 1.0                     \\ \hline
Rollout Top-P                       & 0.99                    \\ \hline
$N$ Samples to Estimate Task Accuracy & 10                    \\ \hline
Overlong Penalty Factor             & 1.0                     \\ \hline
Overlong Penalty Buffer             & 1024                     \\ \hline
Dynamic Filtering                   & True                    \\ \hline
Max Number Generation Batch         & 10                      \\ \hline
\end{tabular}%
\label{tab:think_train_hp}
}
\end{minipage}\hfill
\begin{minipage}[t]{0.49\linewidth}
\centering
\captionof{table}{Qwen3-4/8B Base, LLaMA-3.2-3B/3.1-8B Instruct Training Hyperparameters}
\resizebox{\linewidth}{!}{%
\begin{tabular}{|l|c|}
\hline
\textbf{Parameter}                  & \textbf{Value}          \\ \hline
\multicolumn{2}{|c|}{\textbf{Model Configuration}}  \\ \hline
Max Prompt Length                   & 6144                    \\ \hline
Max Response Length                 & 8096                    \\ \hline
Seed Batch Factor                   & 42                      \\ \hline
\multicolumn{2}{|c|}{\textbf{Training Settings}}    \\ \hline
Train Batch Size                    & 64                      \\ \hline
Generation Batch Size               & 88                      \\ \hline
Learning Rate                       & 1e-6                    \\ \hline
Optimizer                           & AdamW                   \\ \hline
Weight Decay                        & 0.01                    \\ \hline
Grad Clip                           & 1.0                     \\ \hline
Total Steps                         & 200                     \\ \hline
\multicolumn{2}{|c|}{\textbf{RL Settings}}          \\ \hline
Adv. Estimator                      & DAPO                    \\ \hline
KL Loss                             & 0.0                     \\ \hline
KL Reward                           & 0.0                     \\ \hline
Entropy Coefficient                 & 0.0                     \\ \hline
PPO Epochs                          & 16                      \\ \hline
$N$ Rollouts                        & 16                      \\ \hline
Rollout Temperature                 & 1.0                     \\ \hline
Rollout Top-P                       & 0.99                    \\ \hline
$N$ Samples to Estimate Task Accuracy & 10                    \\ \hline
Overlong Penalty Factor             & 0.0                     \\ \hline
Overlong Penalty Buffer             & 0                    \\ \hline
Dynamic Filtering                   & True                    \\ \hline
Max Number Generation Batch         & 10                      \\ \hline
\end{tabular}%
\label{tab:base_train_hp}
}
\end{minipage}
\end{table}

\begin{table}[htbp]
\centering
\label{tab:eval_hp}
\begin{minipage}[t]{0.49\linewidth}
\centering
\captionof{table}{Qwen3-4B Thinking Evaluation Hyperparameters}
\normalsize
\begin{tabular}{|l|c|}
\hline
\textbf{Parameter} & \textbf{Value} \\ \hline
Max Response Length & 8096 \\ \hline
Temperature & 0.6 \\ \hline
Top-P & 0.95 \\ \hline
Number of Samples & 8 \\ \hline
\end{tabular}
\label{tab:think_eval_hp}
\end{minipage}\hfill
\begin{minipage}[t]{0.49\linewidth}
\centering
\captionof{table}{Qwen3-4/8B Base, LLaMA-3.2-3B/3.1-8B Instruct Evaluation Hyperparameters}
\normalsize
\begin{tabular}{|l|c|}
\hline
\textbf{Parameter} & \textbf{Value} \\ \hline
Max Response Length & 8096 \\ \hline
Temperature & 0.6 \\ \hline
Top-P & 0.95 \\ \hline
Number of Samples & 8 \\ \hline
\end{tabular}
\label{tab:base_eval_hp}
\end{minipage}
\end{table}

\section{Qualitative Examples}
\label{app:qual_examples}
We present qualitative examples for our \texttt{Attribute Mutation} and \texttt{Skill Crossover} in Figure \ref{fig:mutate_qual_example} and Figure \ref{fig:skill_cross_example}, respectively.

\paragraph{\texttt{Attribute Mutation}} This mutation introduces a secondary sliding window phase that processes the output of the first window (the sums) rather than the raw input. Specifically, it applies a monotonic queue (deque) to efficiently find the maximum value within a second window of size m over the intermediate sums, increasing the algorithmic complexity from a simple aggregation to a multi-stage dual-window pipeline.
\definecolor{codegreen}{rgb}{0,0.6,0}
\definecolor{codegray}{rgb}{0.5,0.5,0.5}
\definecolor{codepurple}{rgb}{0.58,0,0.82}
\definecolor{backcolour}{rgb}{0.95,0.95,0.92}
\definecolor{mutationbg}{rgb}{1.0, 0.9, 0.9} 
\definecolor{mutationframe}{rgb}{0.8, 0.4, 0.4} 

\lstdefinestyle{mystyle}{
    language=Python,
    backgroundcolor=\color{backcolour},   
    commentstyle=\color{codegreen},
    keywordstyle=\color{magenta}\bfseries,
    numberstyle=\tiny\color{codegray},
    stringstyle=\color{codepurple},
    basicstyle=\ttfamily\footnotesize,
    breakatwhitespace=false,         
    breaklines=true,                 
    captionpos=b,                    
    keepspaces=true,                 
    numbers=left,                    
    numbersep=5pt,                  
    showspaces=false,                
    showstringspaces=false,
    showtabs=false,                  
    tabsize=4
}

\lstset{style=mystyle}
\captionsetup{skip=4pt}

\begin{figure}[htb]
\centering
\begin{tcolorbox}[
    title={Mutation: Window Size},
    colback=backcolour,
    colframe=black!70,
    fonttitle=\bfseries,
    width=0.95\linewidth,   
    boxsep=1pt,
    left=0.5pt,
    right=0.5pt,
    top=1pt,
    bottom=1pt,
    ]
    \begin{lstlisting}[frame=none, belowskip=-10pt]
from collections import deque

def f(arr: list[int], k: int, m: int):
    n = len(arr)
    if k <= 0 or k > n:
        return -1
    sums = []
    window_sum = sum(arr[:k])
    sums.append(window_sum)
    for i in range(k, n):
        window_sum += arr[i] - arr[i - k]
        sums.append(window_sum)
    \end{lstlisting}

    \begin{tcolorbox}[
        colback=mutationbg,
        colframe=mutationframe,
        boxrule=0.8pt,
        arc=0mm,
        boxsep=0pt,
        left=0pt, right=0pt, top=0pt, bottom=0pt
    ]
    \begin{lstlisting}[frame=none, firstnumber=13, aboveskip=0pt, belowskip=0pt]
    # === MUTATION START: Max-Window Phase ===
    m1 = len(sums)
    if m <= 0 or m > m1:
        return -1
    maxes = []
    dq = deque()
    for i in range(m1):
        while dq and sums[dq[-1]] <= sums[i]:
            dq.pop()
        dq.append(i)
        if dq[0] <= i - m:
            dq.popleft()
        if i >= m - 1:
            maxes.append(sums[dq[0]])
    # === MUTATION END ===
    \end{lstlisting}
    \end{tcolorbox}

    \begin{lstlisting}[frame=none, firstnumber=28, aboveskip=-3pt, belowskip=-2pt]
    L = min(len(sums), len(maxes))
    diffs = [maxes[i] - sums[i] for i in range(L)]
    min_diff = diffs[0]
    idx = 0
    for i in range(1, L):
        if diffs[i] < min_diff:
            min_diff = diffs[i]
            idx = i
    return idx
    \end{lstlisting}

\end{tcolorbox}
\caption{Illustration of Attribute Mutation, showing how complexity is increased through an additional max-window operation without altering the core algorithmic logic.}
\label{fig:mutate_qual_example}
\end{figure}

\paragraph{\texttt{Skill Crossover}}
The code integrates \textbf{parametric search} to optimize the number of updates, using a \textbf{binary search} framework to verify feasibility. Inside each verification step, it employs a \textbf{difference array} combined with prefix sums to efficiently reconstruct the polygon's y-coordinates in linear time, before finally applying the shoelace formula to calculate the \textbf{polygon area}. This layering allows for determining the minimal prefix of updates required to meet the area target without simulating the entire process quadratically.

\definecolor{codegray}{rgb}{0.5,0.5,0.5}
\definecolor{backcolour}{rgb}{0.97,0.97,0.97}

\definecolor{skillAreaBG}{rgb}{0.9, 0.95, 1.0}
\definecolor{skillAreaFrame}{rgb}{0.3, 0.5, 0.8}

\definecolor{skillDiffBG}{rgb}{0.9, 1.0, 0.9}
\definecolor{skillDiffFrame}{rgb}{0.3, 0.7, 0.3}

\definecolor{skillPrefixBG}{rgb}{1.0, 0.95, 0.85}
\definecolor{skillPrefixFrame}{rgb}{0.9, 0.6, 0.2}

\definecolor{skillSearchBG}{rgb}{0.95, 0.9, 1.0}
\definecolor{skillSearchFrame}{rgb}{0.6, 0.3, 0.8}

\lstdefinestyle{mystyle}{
    language=Python,
    backgroundcolor=\color{backcolour},   
    basicstyle=\ttfamily\footnotesize,
    keywordstyle=\color{magenta}\bfseries,
    commentstyle=\color{codegray},
    breaklines=true,                 
    numbers=left,                    
    numbersep=5pt,                  
    tabsize=4
}

\lstset{style=mystyle}

\begin{figure}[htb!]
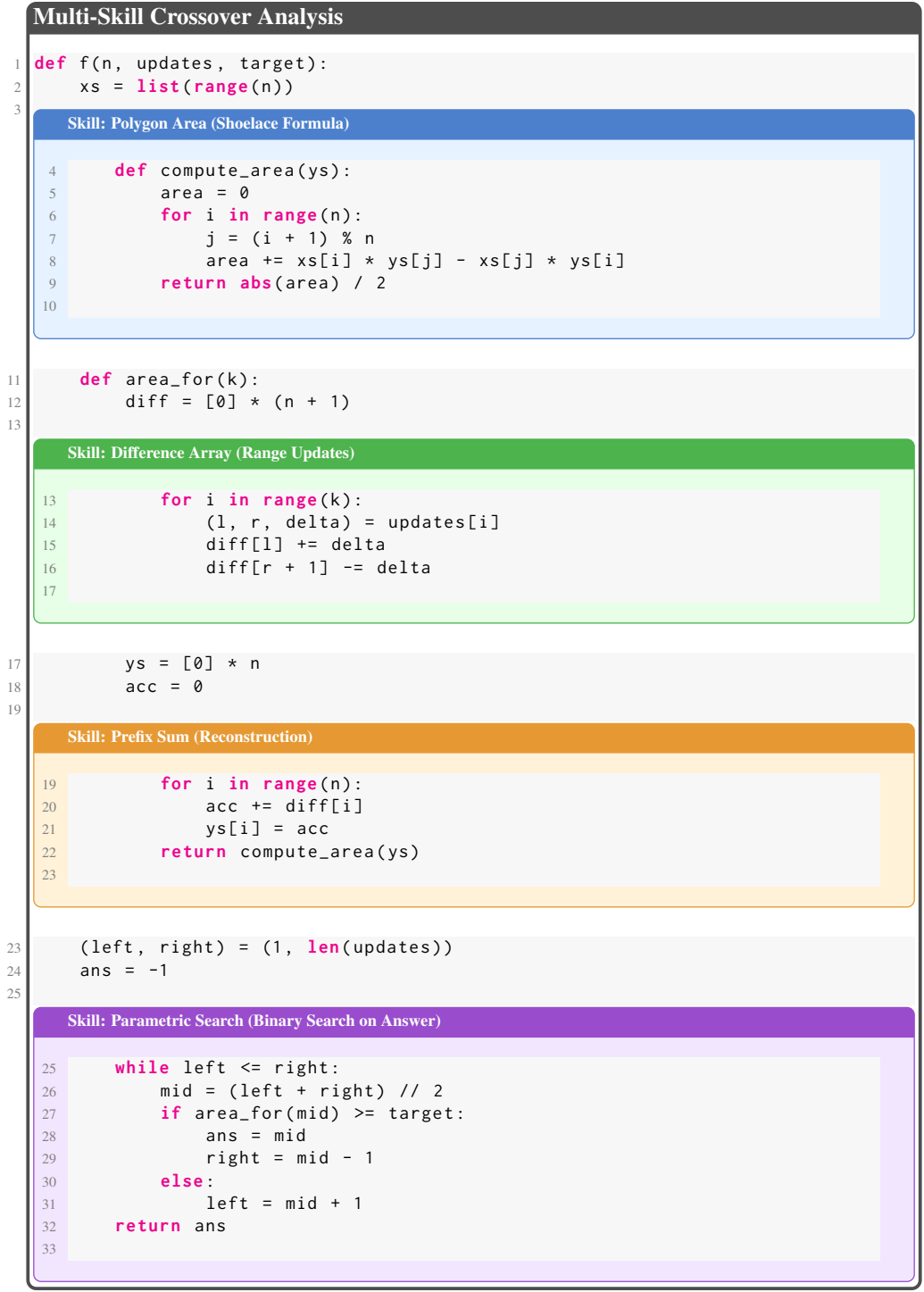

    \centering
\begin{tcolorbox}[
    title={Multi-Skill Crossover Analysis},
    colback=white,
    colframe=black!70,
    fonttitle=\bfseries,
    width=0.95\linewidth,   
    boxsep=1pt,
    left=0.5pt,
    right=0.5pt,
    top=1pt,
    bottom=1pt,
]
    \begin{lstlisting}[frame=none, belowskip=-5pt]
def f(n, updates, target):
    xs = list(range(n))
    \end{lstlisting}

    \begin{tcolorbox}[colback=skillAreaBG, colframe=skillAreaFrame, title=Skill: Polygon Area (Shoelace Formula), fonttitle=\bfseries\scriptsize, boxrule=0.5pt]
    \begin{lstlisting}[frame=none, firstnumber=4, aboveskip=0pt, belowskip=0pt]
    def compute_area(ys):
        area = 0
        for i in range(n):
            j = (i + 1) % n
            area += xs[i] * ys[j] - xs[j] * ys[i]
        return abs(area) / 2
    \end{lstlisting}
    \end{tcolorbox}

    \begin{lstlisting}[frame=none, firstnumber=11, belowskip=-5pt]
    def area_for(k):
        diff = [0] * (n + 1)
    \end{lstlisting}

    \begin{tcolorbox}[colback=skillDiffBG, colframe=skillDiffFrame, title=Skill: Difference Array (Range Updates), fonttitle=\bfseries\scriptsize, boxrule=0.5pt]
    \begin{lstlisting}[frame=none, firstnumber=13, aboveskip=0pt, belowskip=0pt]
        for i in range(k):
            (l, r, delta) = updates[i]
            diff[l] += delta
            diff[r + 1] -= delta
    \end{lstlisting}
    \end{tcolorbox}
    
    \begin{lstlisting}[frame=none, firstnumber=17, belowskip=-5pt]
        ys = [0] * n
        acc = 0
    \end{lstlisting}

    \begin{tcolorbox}[colback=skillPrefixBG, colframe=skillPrefixFrame, title=Skill: Prefix Sum (Reconstruction), fonttitle=\bfseries\scriptsize, boxrule=0.5pt]
    \begin{lstlisting}[frame=none, firstnumber=19, aboveskip=0pt, belowskip=0pt]
        for i in range(n):
            acc += diff[i]
            ys[i] = acc
        return compute_area(ys)
    \end{lstlisting}
    \end{tcolorbox}

    \begin{lstlisting}[frame=none, firstnumber=23, belowskip=-5pt]
    (left, right) = (1, len(updates))
    ans = -1
    \end{lstlisting}

    \begin{tcolorbox}[colback=skillSearchBG, colframe=skillSearchFrame, title=Skill: Parametric Search (Binary Search on Answer), fonttitle=\bfseries\scriptsize, boxrule=0.5pt]
    \begin{lstlisting}[frame=none, firstnumber=25, aboveskip=0pt, belowskip=0pt]
    while left <= right:
        mid = (left + right) // 2
        if area_for(mid) >= target:
            ans = mid
            right = mid - 1
        else:
            left = mid + 1
    return ans
    \end{lstlisting}
    
    \end{tcolorbox}

\end{tcolorbox}
\caption{Example code integrating parametric search, difference array, prefix sum, and polygon area.}
\label{fig:skill_cross_example}
\end{figure}

\newpage
\section{Skills and Attributes}
\label{section:skills-and-attributes}
Figures \ref{fig:skill_list} and \ref{fig:attribute_name_list} present the skills and complexity attributes we collect from USACO \cite{shi2024can}. Each skill and attribute are associated with descriptions that help the proposer better understand the meaning.

\begin{tcolorbox}[
  colback=gray!5!white,
  colframe=gray!75!black,
  title={\bfseries Skill Bank},
  width=\textwidth,
  boxrule=0.8pt,
  arc=4pt,
  outer arc=4pt,
  boxsep=4pt,
  left=3pt,
  right=6pt,
  top=4pt,
  bottom=4pt,
  fontupper=\footnotesize,
  breakable
]

\setlength{\parskip}{0.2em}
\setlength{\parindent}{0pt}

\newcommand{\SubHeader}[1]{%
  \par\smallskip
  {\bfseries #1}%
  \par\smallskip
}

\setlist[itemize]{%
  leftmargin=1.4em,
  itemsep=0.2em,
  topsep=0pt,
  parsep=0pt,
  partopsep=0pt
}

\SubHeader{Sorting \& Searching}
\begin{itemize}
  \item \textbf{sorting}: Arrange elements in non-decreasing order using a comparison-based sort in $O(n log n)$ time.
  \item \textbf{binary search}: Locate a target or insertion index in a sorted array in $O(log n)$ time.
  \item \textbf{two pointers}: Use two indices to traverse or maintain a window over a sequence under constraints in $O(n)$ time.
  \item \textbf{sliding window}: Maintain a variable or fixed-size window on a sequence, updating aggregate statistics in $O(n)$ time.
  \item \textbf{parametric search}: Perform binary search over the answer space using a monotonic feasibility predicate.
\end{itemize}

\SubHeader{Prefix \& Difference Arrays}
\begin{itemize}
  \item \textbf{prefix sum}: Compute cumulative sums of an array to answer range-sum queries in $O(1)$ or $O(log n)$ time.
  \item \textbf{difference array}: Use boundary differences to support range-add updates in $O(1)$ and reconstruct values via prefix sum.
  \item \textbf{range sum query}: Query the sum of any subarray using precomputed prefix sums or Fenwick tree in $O(1)$ or $O(log n)$.
  \item \textbf{prefix xor}: Compute cumulative XORs of an array to enable subarray XOR queries in $O(1)$.
  \item \textbf{range xor query}: Answer the XOR of any contiguous subarray by combining two prefix XOR values in $O(1)$.
\end{itemize}

\SubHeader{Data Structures}
\begin{itemize}
  \item \textbf{union find}: Maintain disjoint sets with union-by-rank and path compression, supporting near-constant-time union and find.
  \item \textbf{fenwick tree}: Support point updates and prefix-sum queries on an array in $O(log n)$ time.
  \item \textbf{segment tree}: Support point and range queries or updates on an array in $O(log n)$ time, with optional lazy propagation.
  \item \textbf{hash table}: Map keys to values with average $O(1)$ insert, delete, and lookup operations.
  \item \textbf{priority queue}: Retrieve and remove the minimum or maximum element in $O(log n)$ time.
  \item \textbf{multiset}: Maintain a multiset allowing duplicates with efficient insertions, deletions, and order queries.
\end{itemize}

\SubHeader{Graph \& Trees}
\begin{itemize}
  \item \textbf{bfs}: Traverse or compute distances in unweighted graphs level by level using a queue in $O(V+E)$.
  \item \textbf{dfs}: Recursively or iteratively traverse a graph depth-first to explore all reachable nodes in $O(V+E)$.
  \item \textbf{dijkstra}: Compute shortest paths in graphs with non-negative weights using a priority queue in $O((V+E)$ log V).
  \item \textbf{zero one bfs}: Compute shortest paths in graphs with 0/1 edge weights using a deque in $O(V+E)$.
  \item \textbf{floyd warshall}: Compute all-pairs shortest paths in a weighted graph in $O(n^3)$ time.
  \item \textbf{topological sort}: Order nodes of a DAG so that all edges go from earlier to later in $O(V+E)$ time.
  \item \textbf{mst kruskal}: Construct a minimum spanning tree by sorting edges and using union-find in $O(E log E)$ time.
  \item \textbf{prim}: Build a minimum spanning tree by growing from a start node with a priority queue in $O(E log V)$.
\end{itemize}

\SubHeader{Dynamic Programming}
\begin{itemize}
  \item \textbf{dp 1d}: Use a one-dimensional array to iteratively compute optimal values with local transitions in $O(n)$ time.
  \item \textbf{knapsack dp}: Solve the 0/1 knapsack problem by updating a capacity-indexed DP array in $O(nC)$ time.
  \item \textbf{grid dp}: Perform DP on a 2D grid with state transitions from neighboring cells in $O(n·m)$ time.
  \item \textbf{tree dp}: Aggregate child results at each node in a tree via post-order traversal in $O(n)$ time.
  \item \textbf{dp on dag}: Compute DP on a DAG by processing nodes in topological order and relaxing outgoing edges.
  \item \textbf{bitmask dp}: Perform DP over subsets represented by bitmasks, iterating through masks for combinatorial counts.
\end{itemize}

\SubHeader{Math \& Number Theory}
\begin{itemize}
  \item \textbf{gcd}: Compute the greatest common divisor of two integers via the Euclidean algorithm in $O(log min(a,b)$).
  \item \textbf{integer division}: Perform floor and ceiling division on integers in $O(1)$ time.
  \item \textbf{modular arithmetic}: Perform addition, subtraction, multiplication, and exponentiation under a modulus.
  \item \textbf{sieve of eratosthenes}: Precompute primality of numbers up to N in $O(N log log N)$ time.
  \item \textbf{prime factorization}: Factor an integer into primes by trial division up to its square root in $O(\sqrt{n})$ time.
  \item \textbf{divisor enumeration}: Enumerate all divisors of an integer by scanning up to its square root in $O(\sqrt{n})$ time.
  \item \textbf{binomial coefficients}: Precompute factorials and inverse factorials mod a prime to answer n choose k queries in $O(1)$.
\end{itemize}

\SubHeader{Geometry \& Computational Geometry}
\begin{itemize}
  \item \textbf{convex hull}: Compute the convex hull of a point set in $O(n log n)$ time using Graham scan or monotonic chain.
  \item \textbf{cross product}: Compute the 2D cross product to test orientation or compute signed area contributions.
  \item \textbf{polygon area}: Compute the area of a simple polygon via the shoelace formula in $O(n)$ time.
  \item \textbf{rectangle union area}: Compute union area of axis-aligned rectangles with a sweep-line and segment coverage in $O(n log n)$.
  \item \textbf{point in triangle}: Test if a point lies inside a triangle via area or barycentric coordinate checks in $O(1)$.
  \item \textbf{distance computation}: Compute Euclidean or Manhattan distance between points in $O(1)$ time.
\end{itemize}

\SubHeader{String Algorithms}
\begin{itemize}
  \item \textbf{kmp failure function}: Compute the prefix-function for a pattern in $O(m)$ to support efficient substring search.
  \item \textbf{rolling hash}: Compute hash fingerprints of strings or substrings for $O(1)$ comparison with precomputed powers.
  \item \textbf{edit distance}: Compute the minimum cost of insertions, deletions, and substitutions between two strings via DP in $O(nm)$.
  \item \textbf{palindrome expansion}: Enumerate palindromic substrings by expanding around centers in $O(n^2)$ time (or use Manacher’s for $O(n)$).
\end{itemize}

\SubHeader{Greedy Algorithms}
\begin{itemize}
  \item \textbf{interval scheduling}: Select the maximum number of non-overlapping intervals by sorting by end time and greedy selection in $O(n log n)$.
  \item \textbf{graph coloring greedy}: Assign colors to graph vertices in a fixed order by choosing the smallest available color per vertex.
  \item \textbf{coupon selection greedy}: Under a budget constraint, iteratively pick the locally optimal purchase option to maximize items acquired.
  \item \textbf{fuel station greedy}: Minimize fuel cost along a route by computing next cheaper station using a monotonic stack in $O(n)$.
\end{itemize}

\SubHeader{Miscellaneous Techniques}
\begin{itemize}
  \item \textbf{meet in the middle}: Split a problem in half, compute results on each side, and combine in $O(2^{n/2})$ time.
  \item \textbf{brute force enumeration}: Systematically iterate over all candidates in a finite search space to test each solution.
  \item \textbf{sweep line}: Process events in sorted order while maintaining active segments or counts in a data structure.
  \item \textbf{inclusion exclusion}: Count elements satisfying at least one of several conditions by alternately adding and subtracting intersection counts.
  \item \textbf{simulation}: Model and execute step-by-step state changes according to rules to obtain final outcomes.
\end{itemize}

\end{tcolorbox}

\captionof{figure}{Skill Categories and Descriptions}
\label{fig:skill_list}

\begin{tcolorbox}[
  colback=gray!5!white,
  colframe=gray!75!black,
  title=\bfseries Complexity Attribute List,
  width=\textwidth,
  boxrule=0.8pt,
  arc=4pt,
  outer arc=4pt,
  boxsep=4pt,
  left=3pt,
  right=6pt,
  top=4pt,
  bottom=4pt,
  fontupper=\footnotesize,
  breakable
]

\setlength{\parskip}{0.25em}
\setlength{\parindent}{0pt}

\textbf{input\_size\_n}: Primary scale N of the input data (e.g., number of elements, string length, graph nodes); directly influences time and space complexity.\par
\textbf{test\_case\_count}: Number of independent test cases T; total runtime and memory scale linearly with T.\par
\textbf{query\_count}: Number of queries Q or subproblem invocations; overall work typically multiplies by Q.\par
\textbf{update\_count}: Number of update or modification operations U; more updates increase total processing time accordingly.\par
\textbf{alphabet\_size}: Number of distinct symbols or characters $\sigma$; affects automaton construction, transition branching, and memory for alphabet‐indexed structures.\par
\textbf{value\_range}: Range or domain size of numeric input values; large ranges may require coordinate compression or affect data‐structure performance.\par
\textbf{bit\_length}: Bit‐length of integer values; larger magnitudes increase the cost of arithmetic operations like multiplication, division, and modulo.\par
\textbf{graph\_vertices}: Number of vertices N in a graph; determines sizes of adjacency lists/matrices and base complexity of graph algorithms.\par
\textbf{graph\_edges}: Number of edges M in a graph; influences traversal and update costs by scaling adjacency operations.\par
\textbf{directed\_graph}: Flag indicating whether a graph is directed; directed edges alter adjacency considerations and may change algorithmic choices.\par
\textbf{dynamic\_graph}: Flag for dynamic graph updates (insertions/removals); dynamic scenarios require specialized data structures and often incur extra log- or amortized costs.\par
\textbf{pattern\_length}: Length m of a pattern or substring; affects DP dimensions, automaton states, and preprocessing time, typically linearly in m.\par
\textbf{window\_size}: Size k of a sliding window or local segment; controls how many elements are processed per window operation.\par
\textbf{segment\_count}: Number of segments or intervals N (e.g., line segments, time intervals); larger counts raise event‐based sweep or segment‐tree operation costs.\par
\textbf{dp\_state\_dimensions}: Number of independent parameters tracked in a DP state; each additional dimension multiplies the DP table and transition work.\par
\textbf{dp\_transition\_branching}: Average number of transitions or options per DP state; higher branching factors increase per‐state transition complexity.\par
\textbf{search\_space\_size}: Total size of the combinatorial search space; directly bounds exhaustive or backtracking algorithm runtimes.\par
\textbf{branching\_factor}: Branching factor in recursive or enumeration algorithms; high factors lead to exponential state‐space growth.\par
\textbf{recursion\_depth}: Maximum depth of recursion or number of iterative steps; deeper recursion increases total computation and call‐stack usage.\par
\textbf{monotonicity}: Boolean indicating whether a predicate or metric behaves monotonically, enabling search optimizations like two-pointers or binary search.\par
\textbf{grid\_dimensions}: Dimensions (rows, columns) of a grid or matrix; larger grids raise state counts and traversal costs, often quadratically in size.\par
\textbf{coordinate\_dimension}: Spatial dimensionality D of geometric inputs; higher dimensions expand neighbor counts and combinatorial complexity.\par
\textbf{constraint\_count}: Number of constraints C (e.g., inequalities, logic rules); more constraints typically increase checking and preprocessing time.\par
\textbf{constraint\_type\_count}: Number of distinct constraint forms or operators; more types enlarge case‐handling logic and may raise constant factors.\par
\textbf{operation\_type\_count}: Number of allowed operation types (e.g., insertion, deletion, substitution); more types expand DP branching and search transitions.\par
\textbf{operation\_cost\_variability}: Degree of variability in operation costs; nonuniform costs require weighted algorithms or more complex DP logic.\par
\textbf{objectives\_count}: Number of simultaneous objectives to optimize; multi‐objective problems often require vector comparisons and higher‐dimensional reasoning.\par
\end{tcolorbox}

\captionof{figure}{Complexity Attribute List}
\label{fig:attribute_name_list}

\section{Limitations}
\label{app:limits}
\paragraph{Dependence on the initial skill bank.}
The effectiveness of \textsc{EvoTD} is partly contingent on the quality and breadth of the initial skill bank. We mitigate this through a data-driven extraction pipeline that leverages LLM metacognition (\S\ref{subsection:abstraction}) and verify robustness to substantial taxonomy reduction in Appendix~\ref{app:evotd-sense}, where halving the skill bank still yields competitive performance. Nevertheless, gains may attenuate in domains where authoritative skill primitives are sparse or poorly structured.

\paragraph{Reliance on programmatically verifiable rewards.}
Our multi-objective fitness check is naturally instantiated in domains that admit deterministic verification---algorithmic code and symbolic mathematics in this work. Extension to open-ended reasoning settings without programmatic oracles (e.g., long-form argumentation, multimodal grounding) would require substituting the executor with alternative validators such as learned reward models, LLM-as-judge protocols, or finer-grained rubrics rewards. We view this as a natural avenue for future investigation rather than an obstacle to the underlying methodology.

\paragraph{Static skill and complexity-attribute taxonomies.}
Both the skill bank $\mathcal{S}$ and the complexity attribute set $\mathcal{C}$ are extracted once from the seed corpus and held fixed throughout the evolutionary process. Although our quality and coverage analysis (Appendix~\ref{app:skill-qual}) confirms the extracted bank is already comprehensive, finer-grained primitives or emergent skill compositions that arise as the solver matures are not automatically incorporated. Online taxonomy expansion, where new skills are discovered or refined during training, is a promising direction we leave to subsequent work.

\newpage
\section{\evotd Algorithm}
\begin{algorithm}[H]

\normalsize
\caption{\textbf{EvoTD Algorithm}}
\label{alg:evotd}
\begin{algorithmic}[1]
\REQUIRE Proposer $\pi_{\text{prop}}$; Solver $\pi_{\text{sol}}$;
Verifier $\mathcal{V}$; Seed $\mathcal{D}_{\text{seed}}$
\ENSURE Training Tasks $\mathcal{D}_{\text{final}}$

\STATE $\mathcal{S} \gets \mathrm{ExtractSkills}(\pi_{\text{prop}}, \mathcal{D}_{\text{seed}})$, $\mathcal{D}^{\text{ind}} \gets \emptyset$, $\mathcal{C} \gets \mathrm{ExtractAttributes}(\pi_{\text{prop}}, \mathcal{D}_{\text{seed}})$, $\mathcal{D}_{\text{final}} \gets \emptyset$

\FOR{each $\textnormal{type} \in \{\textnormal{Ded}, \textnormal{Abd}\}$}
    \STATE $\mathcal{D}^{\text{type}}_{\text{skill}} \gets \emptyset$,
           $\mathcal{D}^{\text{type}}_{\text{mutate}} \gets \emptyset$,
           $\mathcal{D}^{\text{type}}_{\text{crossover}} \gets \emptyset$

    \STATE $\triangleright$ \textit{\textcolor{tblue}{Seeding: Skill-conditioned Task Synthesis}}
    \WHILE{$\mathcal{S} \neq \emptyset$}
        \STATE $t^{\text{type}}_s \gets \mathrm{SkillSeeding}(\pi_{\text{prop}}, s, \textnormal{type})$
        \IF{$\mathcal{V}(t^{\text{type}}_s, \pi_{\text{sol}}, s)$}
            \STATE $\mathcal{D}^{\text{type}}_{\text{skill}}.\mathrm{append}(t^{\text{type}}_s)$, $\mathcal{S}.\mathrm{remove}(s)$
        \ENDIF
    \ENDWHILE

    \STATE $\triangleright$ \textit{\textcolor{tblue}{Mutation: Mutate over Applicable Attributes}}
    \WHILE{$\mathcal{S} \neq \emptyset$}
        \STATE $t^{\text{type}}_s \gets \mathcal{D}^{\text{type}}_{\text{skill}}[s]$, $\textnormal{Mutated} \gets \textnormal{False}$
        \WHILE{$n \neq 0$}
            \STATE $t^{\text{type}'}_s \gets \mathcal{M}_{\text{attr}}(t^{\text{type}}_s, \pi_{\text{prop}})$
            \IF{$\mathcal{V}(t^{\text{type}'}_s, \pi_{\text{sol}}, s)$}
                \STATE $\mathcal{D}^{\text{type}}_{\text{mutate}}.\mathrm{append}(t^{\text{type}'}_s)$, $\textnormal{Mutated} \gets \textnormal{True}$
            \ENDIF
            \IF{$\textnormal{Mutated} = \textnormal{True}$}
                \STATE $\mathcal{S}.\mathrm{remove}(s)$
            \ENDIF
        \ENDWHILE
    \ENDWHILE

    \STATE $\triangleright$ \textit{\textcolor{tblue}{Crossover: Explore Novel Skill Combinations}}
    \WHILE{$\mathcal{S} \neq \emptyset$}
        \STATE $t^{\text{type,new}}_s \gets \mathcal{X}_{\text{skill}}\!\left(
            \Lambda\!\left(\mathcal{D}^{\text{type}}_{\text{skill}} \cup \mathcal{D}^{\text{type}}_{\text{mutate}}\right),
            \pi_{\text{prop}}
        \right)$, $\textnormal{Crossovered} \gets \textnormal{False}$
        \IF{$\mathcal{V}(t^{\text{type,new}}_s, \pi_{\text{sol}}, s)$}
            \STATE $\mathcal{D}^{\text{type}}_{\text{crossover}}.\mathrm{append}(t^{\text{new}}_s)$, $\textnormal{Crossovered} \gets \textnormal{True}$
        \ENDIF
        \IF{$\textnormal{Crossovered} = \textnormal{True}$}
            \STATE $\mathcal{S}.\mathrm{remove}(s)$
        \ENDIF
    \ENDWHILE

    \STATE $\mathcal{D}_{\text{final}} \gets
    \mathcal{D}^{\text{type}}_{\text{skill}} \cup
    \mathcal{D}^{\text{type}}_{\text{mutate}} \cup
    \mathcal{D}^{\text{type}}_{\text{crossover}}$
\ENDFOR

\STATE $\triangleright$ \textit{\textcolor{tblue}{Synthesize Ind. Task from Ded. and Abd. Task}}
\FOR{$t \in \mathcal{D}_{\text{final}}$}
    \STATE $t^{\text{ind}} \gets \mathrm{Synthesis}(\pi_{\text{prop}}, t)$
    \IF{$\mathcal{V}(t^{\text{ind}}, \pi_{\text{sol}})$}
        \STATE $\mathcal{D}^{\text{ind}}.\mathrm{append}(t^{\text{ind}})$
    \ENDIF
\ENDFOR

\STATE \Return $\mathcal{D}_{\text{final}} \cup \mathcal{D}^{\text{ind}}$
\end{algorithmic}
\end{algorithm}

\section{Prompts}

We organize proposer synthesis prompts by their functional role in \evotd.

\paragraph{Skill and Attribute.}
Figures~\ref{fig:skill_attribute_prompt}, \ref{fig:cluster_skill_prompt}, \ref{fig:cluster_attribute_prompt}, and \ref{fig:skill_reflection_prompt} correspond to prompts used for extracting atomic skills and attributes, clustering skills and attributes, and reflecting on skill usage.

\paragraph{Abduction Tasks.}
We list prompts for skill-based abduction task synthesizing (Figures~\ref{fig:code_input_prompt}), mutating abduction tasks (Figures~\ref{fig:code_input_mutation_prompt}), crossovering abduction tasks (Figures~\ref{fig:code_input_crossover_prompt}). Figure~\ref{fig:code_input_predictor_prompt} shows the prompt for the solver to predict the synthesized abduction tasks.

\paragraph{Deduction Tasks.}
Similar to Abduction tasks, we present prompts for skill-based deduction task synthesizing (Figures~\ref{fig:code_output_prompt}), mutating deduction tasks (Figures~\ref{fig:code_output_mutation_prompt}), crossovering deduction tasks (Figures~\ref{fig:code_output_crossover_prompt}). Figure~\ref{fig:code_output_predictor_prompt} shows the prompt for the solver to predict the synthesized deduction tasks.

\paragraph{Induction Tasks.}
Figures~\ref{fig:code_function_prompt} present induction task prompts. Inspired by \citet{li2025staying, zhou2025expo}, we ask the proposer to generate leetcode style hints to difficult induction tasks (solver \textit{pass@10}=0) (Figures~\ref{fig:code_function_hint_prompt}) Figures~\ref{fig:code_function_predictor_prompt} shows the prompt for the solver to predict the synthesized induction tasks.

\begin{tcolorbox}[
  colback=gray!5!white,
  colframe=gray!75!black,
  title=\bfseries Skill Attribute Prompt,
  width=\textwidth,
  boxrule=0.8pt,
  arc=4pt,
  outer arc=4pt,
  boxsep=4pt,
  left=3pt,
  right=6pt,
  top=4pt,
  bottom=4pt,
  breakable,
  fontupper=\scriptsize
]

\setlength{\parskip}{0.2em}
\setlength{\parindent}{0pt}

\newcommand{\PartHeader}[1]{%
  \smallskip
  {\footnotesize\bfseries\scshape #1}%
  \par\smallskip
}
\newcommand{\SubHeader}[1]{%
  \par\smallskip
  \hspace*{0.15em}{\footnotesize\bfseries #1}%
  \par\smallskip
}

\setlist[itemize]{%
  leftmargin=1.4em,
  itemsep=0.2em,
  topsep=0pt,
  parsep=0pt,
  partopsep=0pt
}

\setlist[enumerate,1]{%
  leftmargin=1.9em,   
  labelwidth=1.2em,
  labelsep=0.4em,
  align=left,
  itemsep=0.2em,
  topsep=0pt,
  parsep=0pt,
  partopsep=0pt
}

\setlist[enumerate,2]{%
  leftmargin=2.6em,
  labelwidth=1.2em,
  labelsep=0.4em,
  align=left,
  itemsep=0.15em,
  topsep=0pt,
  parsep=0pt,
  partopsep=0pt
}

\begin{adjustwidth}{0.6em}{0pt}

You are an expert Computer Science professor and a seasoned competitive programming coach.\\
Your task is to analyze the provided programming problem and its reference solutions to identify:

\begin{enumerate}
    \item The \textbf{ATOMIC} algorithmic skills needed to solve the problem
    \item The \textbf{complexity attributes} -- all characteristics that affect the problem's complexity/difficulty or could be varied to create mutations
\end{enumerate}

Your analysis must be precise, accurate, and adhere to the standardized terminology of the field.

\end{adjustwidth}

\PartHeader{Part 1: Skill Identification}

\begin{adjustwidth}{0.6em}{0pt}

\begin{enumerate}
    \item \textbf{Holistic Analysis}: You must consider both the problem statement and the provided solutions.
    \begin{itemize}
        \item The problem statement provides context and hints at the required complexity.
        \item The solution code provides the definitive implementation, revealing the exact algorithms skills used.
        \item If multiple solutions exist, include the union of distinct skills they use. Do not infer skills not evidenced by the code.
    \end{itemize}

    \item \textbf{Skills}
    \begin{itemize}
        \item Identify the single most important and necessary concept that the problem is testing. Then, break down the concept into secondary or sub-level techniques/skills of the concept required to solve the problem.
        \item Report only algorithmic programming-level techniques/data-structure operations (e.g., binary\_search, two\_pointers, monotonic\_stack, dijkstra, fenwick\_tree, prefix\_sum, etc.).
        \item Avoid overly generic terms (e.g., ``logic,'' ``math'') or trivial implementation details (e.g., ``variables,'' ``loops'').
        \item Skills must be ATOMIC: a single, independent technique—not a composite or pipeline.
        \begin{itemize}
            \item Split composites into primitives (e.g., segment\_tree\_with\_lazy\_propagation $\rightarrow$ segment\_tree, lazy\_propagation; binary\_search\_on\_answer $\rightarrow$ parametric\_search; etc.).
        \end{itemize}
        \item Names must be \textbf{lower\_snake\_case}, descriptive, and canonical (no synonyms, no duplicates).
        \item Provide a brief, precise description of the skill.
    \end{itemize}
\end{enumerate}

\end{adjustwidth}

\PartHeader{Part 2: Complexity Attributes}

\begin{adjustwidth}{0.6em}{0pt}

Analyze the solution and identify \textbf{every attribute that contributes to the problem's complexity or could be modified to create variations}. Think like a problem setter who needs to create variants of the problem with varying complexities.

\end{adjustwidth}

\SubHeader{Discovery Process:}

\begin{adjustwidth}{0.6em}{0pt}

For each major component of the solution (input/output, loops, data structures, operations, conditions, etc.), ask:
\begin{itemize}
    \item What choices were made here?
    \item What constraints or bounds exist?
    \item What could be different while maintaining the core algorithm (skill)?
\end{itemize}

\end{adjustwidth}

\SubHeader{Categories to Explore (non-exhaustive):}

\begin{adjustwidth}{0.6em}{0pt}

\textbf{Quantitative Attributes} - Numeric values that affect complexity:
\begin{itemize}
    \item Input size constraints (array length, string length, matrix dimensions)
    \item Value ranges (minimum/maximum values for integers, characters used)
    \item Iteration bounds (loop limits, recursion depth)
    \item Precision requirements (decimal places, modulo values)
    \item Count limits (number of operations allowed, query limits)
\end{itemize}

\textbf{Structural Attributes} - How the solution is organized:
\begin{itemize}
    \item Data structure choices (array vs linked list, set vs map, etc.)
    \item Traversal patterns (left-to-right, inside-out, breadth-first vs depth-first, etc.)
    \item Processing order (sorted vs unsorted, online vs offline, etc.)
    \item Storage strategy (in-place vs auxiliary space, etc.)
\end{itemize}

\textbf{Operational Attributes} - What operations are performed:
\begin{itemize}
    \item Core operations (addition vs multiplication, min vs max, etc.)
    \item Comparison types (strict vs non-strict inequality, etc.)
    \item Combination methods (sum, product, XOR, concatenation)
    \item Update strategies (increment, replace, accumulate, etc.)
\end{itemize}

\textbf{Logical Attributes} - Decision-making patterns:
\begin{itemize}
    \item Branching conditions (what triggers different paths, etc.)
    \item Early termination conditions (when to stop)
    \item Selection criteria (how elements are chosen, etc.)
    \item Validation rules (what makes a solution valid)
\end{itemize}

\textbf{Complexity Factors} - What makes this instance hard:
\begin{itemize}
    \item Number of nested structures
    \item Dependencies between computations
    \item State tracking requirements
    \item Edge case handling needs
\end{itemize}

\end{adjustwidth}

\SubHeader{Extraction Guidelines:}

\begin{adjustwidth}{0.6em}{0pt}

For each complexity attribute you identify, provide:
\begin{enumerate}
    \item \textbf{Attribute name}: A descriptive \textbf{lower\_snake\_case} identifier
    \item \textbf{Description}: A clear definition of what this attribute represents. How varying this attribute affects problem difficulty.
\end{enumerate}

\end{adjustwidth}

\PartHeader{Part 3: Output Format}

\begin{adjustwidth}{0.6em}{0pt}

Your final response \textbf{must be valid JSON} exactly matching the schema below. Please output \textbf{ONLY} the JSON object. Do not include any extra things.
\begin{verbatim}
{"skills": {
    <name of the skill>: <description of the skill>,
    ...
  },
  "attributes": {
    <name of the attribute>: <description of the attribute>,
    ...
  }}
\end{verbatim}

\textbf{Input}

Problem:\\
\{problem\}

Reference Code Solution:\\
\{code\_solution\}

\end{adjustwidth}

\end{tcolorbox}
\vspace{-4pt}
\captionof{figure}{Skill Attribute Prompt}
\label{fig:skill_attribute_prompt}

\begin{tcolorbox}[
  colback=gray!5!white,
  colframe=gray!75!black,
  title=\bfseries Cluster Skill Prompt,
  width=\textwidth,
  boxrule=0.8pt,
  arc=4pt,
  outer arc=4pt,
  boxsep=4pt,
  left=3pt,
  right=6pt,
  top=4pt,
  bottom=4pt,
  breakable,
  fontupper=\scriptsize
]

\setlength{\parskip}{0.2em}
\setlength{\parindent}{0pt}

\newcommand{\PartHeader}[1]{%
  \smallskip
  {\footnotesize\bfseries\scshape #1}%
  \par\smallskip
}
\newcommand{\SubHeader}[1]{%
  \par\smallskip
  \hspace*{0.15em}{\footnotesize\bfseries #1}%
  \par\smallskip
}

\setlist[itemize]{%
  leftmargin=1.4em,
  itemsep=0.2em,
  topsep=0pt,
  parsep=0pt,
  partopsep=0pt
}
\setlist[enumerate,1]{%
  leftmargin=1.9em,
  labelwidth=1.2em,
  labelsep=0.4em,
  align=left,
  itemsep=0.2em,
  topsep=0pt,
  parsep=0pt,
  partopsep=0pt
}

You are an expert computer science curriculum designer with extensive experience in creating structured learning paths for algorithmic problem-solving.

Your goal is to take a large, potentially redundant list of skills and perform a two-step refinement process:
\begin{adjustwidth}{0.6em}{0pt}
\begin{enumerate}
    \item \textbf{Deduplicate}: Merge overlapping or similar skills into a set of canonical, non-overlapping skills with clear, consolidated descriptions.
    \item \textbf{Categorize}: Group the resulting canonical skills into broader categories based on primary algorithmic concepts or data-structure families.
\end{enumerate}
\end{adjustwidth}

\PartHeader{Input}

\begin{adjustwidth}{0.6em}{0pt}
You will be given a list of skills, where each skill has a \texttt{skill} name and a \texttt{description}.\\
\texttt{\{skill\_list\}}
\end{adjustwidth}

\PartHeader{Your Task}

\SubHeader{Step 1: Deduplicate into Canonical Skills (Mental Step)}

\begin{adjustwidth}{0.6em}{0pt}
\begin{itemize}
    \item First, mentally merge all synonymous, overlapping, or related skills from the input list into a single, canonical skill.
    \item If a skill from the input list is already distinct and does not overlap with any others, treat it as a canonical skill on its own; no deduplication is needed for it.
    \item For each canonical skill, define a standard \texttt{skill} name (in \texttt{lower\_snake\_case}) and write a new concise \texttt{description} that synthesizes the core concept. For example, \texttt{array\_sorting} and \texttt{custom\_sorting\_logic} should be merged into a single canonical skill called \texttt{sorting}.
\end{itemize}
\end{adjustwidth}

\SubHeader{Step 2: Group Canonical Skills into Categories}

\begin{adjustwidth}{0.6em}{0pt}
\begin{itemize}
    \item Now, take the set of canonical skills you just defined.
    \item Group these skills into high-level categories (e.g., \texttt{graph\_algorithms}, \texttt{dynamic\_programming}, \texttt{greedy\_algorithms}).
    \item The final output should be a list of these categories, each containing the relevant canonical skills you created.
\end{itemize}
\end{adjustwidth}

\PartHeader{Output}

\begin{adjustwidth}{0.6em}{0pt}
Your final response \textbf{must be a list of valid JSON objects} exactly matching the schema below.\\
Please output \textbf{ONLY} the JSON. Do not include any extra things.

List of JSON SCHEMA TO FOLLOW:
\end{adjustwidth}

\begin{verbatim}
  {
    "category": <name of the category>,
    "members": [
      {
        "skill": <name of the skill>,
        "description": <description of the skill>
      },
      ...
    ]
  },
  ...
\end{verbatim}

\end{tcolorbox}
\vspace{-2pt}
\captionof{figure}{Cluster Skill Prompt}
\label{fig:cluster_skill_prompt}

\begin{tcolorbox}[
  colback=gray!5!white,
  colframe=gray!75!black,
  title=\bfseries Cluster Attribute Prompt,
  width=\textwidth,
  boxrule=0.8pt,
  arc=4pt,
  outer arc=4pt,
  boxsep=4pt,
  left=3pt,
  right=6pt,
  top=4pt,
  bottom=4pt,
  breakable,
  fontupper=\scriptsize
]

\setlength{\parskip}{0.2em}
\setlength{\parindent}{0pt}

\newcommand{\PartHeader}[1]{%
  \smallskip
  {\footnotesize\bfseries\scshape #1}%
  \par\smallskip
}
\newcommand{\SubHeader}[1]{%
  \par\smallskip
  \hspace*{0.15em}{\footnotesize\bfseries #1}%
  \par\smallskip
}

\setlist[itemize]{%
  leftmargin=1.4em,
  itemsep=0.2em,
  topsep=0pt,
  parsep=0pt,
  partopsep=0pt
}
\setlist[enumerate,1]{%
  leftmargin=1.9em,
  labelwidth=1.2em,
  labelsep=0.4em,
  align=left,
  itemsep=0.2em,
  topsep=0pt,
  parsep=0pt,
  partopsep=0pt
}

You are an expert computer science curriculum designer with extensive experience in creating structured learning paths for algorithmic problem-solving.

Your goal is to take a large, potentially redundant list of attributes that affect the complexity of a coding problem and group overlapping or similar attributes into a set of canonical, non-overlapping attributes with clear, consolidated descriptions.

\PartHeader{Input}

\begin{adjustwidth}{0.6em}{0pt}
You will be given a list of complexity attributes, where each attribute has an attribute name and a description of its definition and impact on problem complexity.\\
\texttt{\{attribute\_list\}}
\end{adjustwidth}

\PartHeader{Clustering Requirements}

\SubHeader{Step 1: Group Similar Concepts}

\begin{adjustwidth}{0.6em}{0pt}
\begin{itemize}
    \item Identify attributes that represent the same underlying complexity dimension.
    \item Look for semantic equivalence, not just naming similarities.
    \item Consider whether attributes would be varied together when creating problem mutations.
\end{itemize}
\end{adjustwidth}

\SubHeader{Step 2: Resolve Overlaps}

\begin{adjustwidth}{0.6em}{0pt}
\begin{itemize}
    \item When attributes partially overlap, determine whether they should merge or remain distinct
    \item Preserve distinctions only if they represent independently variable complexity dimensions
    \item Eliminate redundancy while maintaining coverage
\end{itemize}
\end{adjustwidth}

\SubHeader{Step 3: Create Canonical Attributes}

\begin{adjustwidth}{0.6em}{0pt}
For each unified concept:
\begin{itemize}
    \item Choose the most general and widely applicable name in \textbf{lower\_snake\_case}.
    \item Write a description that:
    \begin{itemize}
        \item Defines what the attribute controls
        \item Explains its impact on complexity when varied
    \end{itemize}
\end{itemize}
\end{adjustwidth}

\SubHeader{Quality Criteria}

\begin{adjustwidth}{0.6em}{0pt}
\begin{itemize}
    \item Each canonical attribute should be independently mutable
    \item No two attributes should be redundant or co-dependent
    \item Names should be intuitive and follow standard computer science terminology
\end{itemize}
\end{adjustwidth}

\PartHeader{Output Format}

\begin{adjustwidth}{0.6em}{0pt}
Your final response must be a list of valid JSON objects exactly matching the schema below.
Please output ONLY the list of JSON objects. Do not include any extra things.

List of JSON SCHEMA TO FOLLOW:
\end{adjustwidth}

\begin{verbatim}
  {
    <name_of_attribute>: <description_of_attribute>
  },
  ...
\end{verbatim}

\end{tcolorbox}
\vspace{-2pt}
\captionof{figure}{Cluster Attribute Prompt}
\label{fig:cluster_attribute_prompt}

\begin{tcolorbox}[
  colback=gray!5!white,
  colframe=gray!75!black,
  title=\bfseries Skill Reflection Prompt,
  width=\textwidth,
  boxrule=0.8pt,
  arc=4pt,
  outer arc=4pt,
  boxsep=4pt,
  left=3pt,
  right=6pt,
  top=4pt,
  bottom=4pt,
  breakable,
  fontupper=\scriptsize
]

\setlength{\parskip}{0.2em}
\setlength{\parindent}{0pt}

\newcommand{\PartHeader}[1]{%
  \smallskip
  {\footnotesize\bfseries\scshape #1}%
  \par\smallskip
}
\newcommand{\SubHeader}[1]{%
  \par\smallskip
  \hspace*{0.15em}{\footnotesize\bfseries #1}%
  \par\smallskip
}

\setlist[itemize]{%
  leftmargin=1.4em,
  itemsep=0.2em,
  topsep=0pt,
  parsep=0pt,
  partopsep=0pt
}
\setlist[enumerate,1]{%
  leftmargin=1.9em,
  labelwidth=1.2em,
  labelsep=0.4em,
  align=left,
  itemsep=0.2em,
  topsep=0pt,
  parsep=0pt,
  partopsep=0pt
}

You are an expert Computer Science professor and a seasoned competitive programming coach.

\PartHeader{Task}

\begin{adjustwidth}{0.6em}{0pt}
Identify the necessary algorithmic skills used in the provided Python code snippet.

You are given a list of algorithmic skills and a Python code snippet. You need to identify all the algorithmic skills used in the code snippet.
\end{adjustwidth}

\PartHeader{Requirements}

\begin{adjustwidth}{0.6em}{0pt}
\begin{itemize}
    \item Review the provided skill list and the code snippet carefully and thoroughly.
    \item Use only the skills listed in the provided skill set. Avoid creating new skills or altering the existing skill names.
\end{itemize}

Each code snippet may involve multiple skills, but there is exactly one \textbf{core testing skill} that is the most important and necessary. If there is no other skills except the core testing skill, then the \texttt{other\_skills} field should be an empty list.
\end{adjustwidth}

\PartHeader{Output Format}

\begin{adjustwidth}{0.6em}{0pt}
Your final response \textbf{must be valid JSON} exactly matching the schema below.\\
Please output \textbf{ONLY} the JSON object. Do not include any extra things.
\end{adjustwidth}
\vspace{-0.6em}
\begin{verbatim}
{
  "main_skill": <name_of_the_skill>,
  "other_skills": [
    <name_of_the_skill>,
    ...
  ]
}
\end{verbatim}
\vspace{-0.6em}
\PartHeader{Input}

\begin{adjustwidth}{0.6em}{0pt}
\textbf{Skill List:}\\
\texttt{\{skill\_list\}}

\medskip
\textbf{Code Snippet:}\\
\texttt{\{code\_snippet\}}
\end{adjustwidth}

\end{tcolorbox}
\vspace{-2pt}
\captionof{figure}{Skill Reflection Prompt}
\label{fig:skill_reflection_prompt}


\begin{tcolorbox}[
  colback=gray!5!white,
  colframe=gray!75!black,
  title=\bfseries Abduction Task Prompt,
  width=\textwidth,
  boxrule=0.8pt,
  arc=4pt,
  outer arc=4pt,
  boxsep=4pt,
  left=3pt,
  right=6pt,
  top=4pt,
  bottom=4pt,
  breakable,
  fontupper=\scriptsize
]

\setlength{\parskip}{0.2em}
\setlength{\parindent}{0pt}

\newcommand{\PartHeader}[1]{%
  \smallskip
  {\footnotesize\bfseries\scshape #1}%
  \par\smallskip
}
\newcommand{\SubHeader}[1]{%
  \par\smallskip
  \hspace*{0.15em}{\footnotesize\bfseries #1}%
  \par\smallskip
}

\setlist[itemize]{%
  leftmargin=1.4em,
  itemsep=0.2em,
  topsep=0pt,
  parsep=0pt,
  partopsep=0pt
}
\setlist[enumerate,1]{%
  leftmargin=1.9em,
  labelwidth=1.2em,
  labelsep=0.4em,
  align=left,
  itemsep=0.2em,
  topsep=0pt,
  parsep=0pt,
  partopsep=0pt
}

\PartHeader{Task}

\begin{adjustwidth}{0.6em}{0pt}
Create a new Python code snippet (custom classes allowed, which must be defined at the top of the snippet) with exactly one matching input.
\end{adjustwidth}

\PartHeader{Failure Information}

\begin{adjustwidth}{0.6em}{0pt}
\texttt{\{failure\_info\}}
\end{adjustwidth}

\PartHeader{Skills}

\begin{adjustwidth}{0.6em}{0pt}
\texttt{\{skill\_str\}}
\end{adjustwidth}

\PartHeader{Code Requirements}

\begin{adjustwidth}{0.6em}{0pt}
\begin{itemize}
  \item The provided skill(s) should be the main testing concept of the code snippet.
  \item Name the entry function \texttt{f} (e.g., \texttt{def f(...): ...}). You may define nested definitions inside \texttt{f}.
  \item Ensure the function returns a value.
  \item Include at least one input parameter.
  \item Make the function deterministic.
  \item Make the snippet require state tracking across multiple data transformations, ensuring the task requires long multi-step reasoning.
  \item Avoid unnecessary function nesting; use simple, readable structure when possible. Only use nested functions if natural for the algorithm.
  \item \textbf{Avoid the following:}
  \begin{itemize}
    \item Random functions or variables
    \item Date/time operations
    \item I/O operations (reading files, network requests)
    \item Printing or logging
    \item Any external state
  \end{itemize}
  \item Ensure execution completes within 10 seconds on a modern CPU.
  \item All imports and class definitions should be at the very top of the code snippet.
  \item The snippet should end with a return statement from the main function \texttt{f}; anything after will be removed.
\end{itemize}

\texttt{\{remove\_input\_from\_snippet\_prompt\}\{remove\_after\_return\_prompt\}}
\end{adjustwidth}

\PartHeader{Input Requirements}

\begin{adjustwidth}{0.6em}{0pt}
\begin{itemize}
  \item Provide exactly one test input for your function.
  \item Format multiple arguments with commas between them.
  \item Remember to add quotes around string arguments.
\end{itemize}
\end{adjustwidth}

\PartHeader{Formatting}

\begin{adjustwidth}{0.6em}{0pt}
Format your code with:
\end{adjustwidth}
\vspace{-0.6em}
\begin{verbatim}
```python
def f(...):
    # your code here
    return ...
```
\end{verbatim}
\vspace{-0.6em}
\begin{adjustwidth}{0.6em}{0pt}
Format your input with:
\end{adjustwidth}
\vspace{-0.6em}
\begin{verbatim}
```input
arg1, arg2, ...
```
\end{verbatim}
\vspace{-0.6em}
\PartHeader{Example Format}
\vspace{-0.6em}
\begin{verbatim}
```python
def f(name: str, info: dict):
    # code logic here
    return result
```

```input
'John', {'age': 20, 'city': 'New York'}
```
\end{verbatim}
\vspace{-0.6em}
\PartHeader{Evaluation Criteria}

\begin{adjustwidth}{0.6em}{0pt}
\begin{itemize}
  \item \textbf{Relevance (High Priority)}: The task should mainly examine the ability of the test subject to understand / reason / apply the skill(s).
  \item Executability: Your code should be executable given your input.
  \item Difficulty in reverse-engineering your \texttt{input} from (1) your \texttt{python} code and (2) the deterministic \texttt{output} obtained from your input.
  \item Creativity: The code must be sufficiently different from the provided reference snippets.
  \item Restricted usage of certain keywords and packages: You are not allowed to use the following words in any form, even in comments: \texttt{<|BANNED\_KEYWORDS|>}.
\end{itemize}
\end{adjustwidth}

\PartHeader{Final Instruction}

\begin{adjustwidth}{0.6em}{0pt}
First, carefully devise a clear plan: how to make the skill(s) the main testing concept, how the snippet will be challenging and creative, and how to resolve any prior failure information. Then, write the final code snippet and its input.
\end{adjustwidth}

\end{tcolorbox}
\vspace{-2pt}
\captionof{figure}{Abduction Task Prompt}
\label{fig:code_input_prompt}

\begin{tcolorbox}[
  colback=gray!5!white,
  colframe=gray!75!black,
  title=\bfseries Abduction Task Mutation Prompt,
  width=\textwidth,
  boxrule=0.8pt,
  arc=4pt,
  outer arc=4pt,
  boxsep=4pt,
  left=3pt,
  right=6pt,
  top=4pt,
  bottom=4pt,
  breakable,
  fontupper=\scriptsize
]

\setlength{\parskip}{0.2em}
\setlength{\parindent}{0pt}

\newcommand{\PartHeader}[1]{%
  \smallskip
  {\footnotesize\bfseries\scshape #1}%
  \par\smallskip
}
\newcommand{\SubHeader}[1]{%
  \par\smallskip
  \hspace*{0.15em}{\footnotesize\bfseries #1}%
  \par\smallskip
}

\setlist[itemize]{%
  leftmargin=1.4em,
  itemsep=0.2em,
  topsep=0pt,
  parsep=0pt,
  partopsep=0pt
}
\setlist[enumerate,1]{%
  leftmargin=1.9em,
  labelwidth=1.2em,
  labelsep=0.4em,
  align=left,
  itemsep=0.2em,
  topsep=0pt,
  parsep=0pt,
  partopsep=0pt
}

\PartHeader{Task}

Create multiple variants of an original code reasoning task by systematically applying complexity attributes to increase difficulty and reasoning requirements.

The original task is a code reasoning deduction task that requires recovering a hidden input from a Python code snippet and its output.

You will be provided with:
\begin{itemize}
  \item The original task (one code snippet and one output)
  \item The core algorithmic skill being tested
  \item A list of complexity attributes that may be applied
\end{itemize}

\PartHeader{Your Mission}

\SubHeader{Step 1: Analyze the Original Task}

\begin{adjustwidth}{0.6em}{0pt}
Examine the provided task and identify:
\begin{itemize}
  \item The current complexity level for each \textbf{applicable} attribute
  \item Opportunities to increase complexity while preserving solvability
\end{itemize}
\end{adjustwidth}

\SubHeader{Step 2: Generate Diverse Variants}

\begin{adjustwidth}{0.6em}{0pt}
Create \textbf{at least 10 distinct variants} using:
\begin{itemize}
  \item Single-attribute mutations
  \item Multi-attribute mutations
  \item Comprehensive mutations applying all relevant attributes
\end{itemize}
\end{adjustwidth}

\PartHeader{Requirements for Each Variant}

\SubHeader{Code Requirements}

\begin{adjustwidth}{0.6em}{0pt}
\begin{itemize}
  \item The provided skill(s) must remain the primary testing concept
  \item The entry function must be named \texttt{f}
  \item The function must return a value and accept at least one input parameter
  \item The code must be deterministic and free of randomness, I/O, logging, or external state
  \item Execution must complete within 10 seconds on a modern CPU
  \item All imports and class definitions must appear at the top of the snippet
  \item The snippet must end with a return statement from \texttt{f}
\end{itemize}
\end{adjustwidth}

\SubHeader{Input Requirements}

\begin{adjustwidth}{0.6em}{0pt}
\begin{itemize}
  \item Provide exactly one hidden test input
  \item Format multiple arguments with commas
  \item Quote string arguments
\end{itemize}
\end{adjustwidth}

\SubHeader{Formatting}

\begin{adjustwidth}{0.6em}{0pt}
Code format:
\begin{verbatim}
```python
def f(...):
    # code here
    return ...
```
\end{verbatim}
\vspace{-0.6em}

Input format:
\vspace{-0.6em}
\begin{verbatim}
```
arg1, arg2, ...
```
\end{verbatim}
\vspace{-0.6em}
\end{adjustwidth}

\SubHeader{Complexity Enhancements}

\begin{adjustwidth}{0.6em}{0pt}
For each variant:
\begin{itemize}
  \item Preserve the core algorithmic skill
  \item Increase difficulty meaningfully, not artificially
  \item Explain which complexity attributes were applied and why the variant is harder
\end{itemize}
\end{adjustwidth}

\PartHeader{Output Format}

\begin{adjustwidth}{0.6em}{0pt}
Your final response must be valid JSON exactly matching the schema below. Output \textbf{ONLY} the JSON.
\end{adjustwidth}
\vspace{-0.6em}
\begin{verbatim}
{
  "variant_1": {
    "complexity_attributes": [...],
    "description": "...",
    "code": "```python\n...\n```",
    "input": "```input\n...\n```"
  },
  ...
}
\end{verbatim}
\vspace{-0.6em}
\PartHeader{Original Task}

\begin{adjustwidth}{0.6em}{0pt}
Code:\\
\texttt{\{code\}}

Input:\\
\texttt{\{input\}}

Skill:\\
\texttt{\{skill\}}

Complexity Attributes:\\
\texttt{\{complexity\_attributes\}}
\end{adjustwidth}

\end{tcolorbox}
\vspace{-2pt}
\captionof{figure}{Abduction Task Mutation Prompt}
\label{fig:code_input_mutation_prompt}

\begin{tcolorbox}[
  colback=gray!5!white,
  colframe=gray!75!black,
  title=\bfseries Abduction Task Crossover Prompt,
  width=\textwidth,
  boxrule=0.8pt,
  arc=4pt,
  outer arc=4pt,
  boxsep=4pt,
  left=3pt,
  right=6pt,
  top=4pt,
  bottom=4pt,
  breakable,
  fontupper=\scriptsize
]

\setlength{\parskip}{0.2em}
\setlength{\parindent}{0pt}

\newcommand{\PartHeader}[1]{%
  \smallskip
  {\footnotesize\bfseries\scshape #1}%
  \par\smallskip
}
\newcommand{\SubHeader}[1]{%
  \par\smallskip
  \hspace*{0.15em}{\footnotesize\bfseries #1}%
  \par\smallskip
}

\setlist[itemize]{%
  leftmargin=1.4em,
  itemsep=0.2em,
  topsep=0pt,
  parsep=0pt,
  partopsep=0pt
}
\setlist[enumerate,1]{%
  leftmargin=1.9em,
  labelwidth=1.2em,
  labelsep=0.4em,
  align=left,
  itemsep=0.2em,
  topsep=0pt,
  parsep=0pt,
  partopsep=0pt
}

\PartHeader{Task}

Create a new Python code reasoning deduction task: a Python code snippet and one matching hidden input.\\
The task must use a novel combination of skills centered on the target core skill.\\
Custom classes are allowed and should be defined at the top of the snippet.

\PartHeader{Available Information}

\SubHeader{Skill Pool}
\begin{adjustwidth}{0.6em}{0pt}
A comprehensive list of algorithmic coding skills (with descriptions) you may choose from:\\
\texttt{\{skill\_pool\}}
\end{adjustwidth}

\SubHeader{Target Core Skill}
\begin{adjustwidth}{0.6em}{0pt}
The primary skill that must be the central focus of your code snippet:\\
\texttt{\{target\_skill\}}
\end{adjustwidth}

\SubHeader{Existing Skill Combinations}
\begin{adjustwidth}{0.6em}{0pt}
Previously created combinations of the target skill with other skills:\\
\texttt{\{existing\_combinations\}}
\end{adjustwidth}

\PartHeader{Your Task}

You must create a \textbf{novel crossover} by combining the target skill with other compatible skills from the skill pool. This requires:
\begin{adjustwidth}{0.6em}{0pt}
\begin{enumerate}
  \item Examine existing combinations to understand what has already been done.
  \item Identify skills from the pool that naturally work with the target skill but \textbf{have not been combined yet}.
  \item Create a code snippet where all chosen skills are \textbf{essential and interconnected} to hide the true input while still producing the provided output.
\end{enumerate}
\end{adjustwidth}

The key challenge is to ensure your skill combination is both \textbf{NEW} (not in \texttt{existing\_combinations}) and \textbf{MEANINGFUL} (skills genuinely complement each other, not forced).

\PartHeader{Code Requirements}

\begin{adjustwidth}{0.6em}{0pt}
\begin{itemize}
  \item The target skill \textbf{must} be the central concept; other skills should naturally support or enhance it.
  \item Name the entry function \texttt{f} (e.g., \texttt{def f(...): ...}); nested definitions inside \texttt{f} are allowed.
  \item Ensure the function returns a value.
  \item Include at least one input parameter.
  \item Make the function deterministic.
  \item Demonstrate \textbf{clear interdependence} between all skills in the combination (not isolated usage).
  \item Require state tracking across multiple data transformations to force multi-step reasoning.
  \item Avoid unnecessary function nesting; keep structure simple and readable unless nesting is natural for the algorithm.
  \item \textbf{Avoid:}
  \begin{itemize}
    \item Random functions or variables
    \item Date/time operations
    \item I/O operations (reading files, network requests)
    \item Printing or logging
    \item Any external state
  \end{itemize}
  \item Ensure execution completes within 10 seconds on a modern CPU.
  \item All imports and class definitions must be at the very top of the code snippet.
  \item The snippet must end with a \texttt{return} statement from the main function \texttt{f}. Anything after may be removed.\\
  \texttt{\{remove\_input\_from\_snippet\_prompt\}\{remove\_after\_return\_prompt\}}
\end{itemize}
\end{adjustwidth}

\PartHeader{Skill Combination Requirements}

\begin{adjustwidth}{0.6em}{0pt}
\begin{itemize}
  \item The new combination must include the target skill plus \textbf{at least one} other skill from the skill pool.
  \item The combination must be meaningfully different from all existing combinations.
  \item Skills should work together \textbf{synergistically}, not just appear sequentially.
  \item You must justify why your chosen skills complement each other and how they work together to obfuscate the hidden input.
\end{itemize}
\end{adjustwidth}

\PartHeader{Input Requirements}

\begin{adjustwidth}{0.6em}{0pt}
\begin{itemize}
  \item Provide exactly one hidden test input for your function.
  \item Format multiple arguments with commas between them.
  \item Remember to add quotes around string arguments.
\end{itemize}
\end{adjustwidth}

\PartHeader{Formatting}

\begin{adjustwidth}{0.6em}{0pt}
Format your code as:
\end{adjustwidth}
\vspace{-0.6em}
\begin{verbatim}
```python
def f(...):
    # your code here
    return ...
```
\end{verbatim}
\vspace{-0.6em}
\begin{adjustwidth}{0.6em}{0pt}
Format your input as:
\end{adjustwidth}

\begin{verbatim}
```input
arg1, arg2, ...
```
\end{verbatim}

\PartHeader{Output Format}

\begin{adjustwidth}{0.6em}{0pt}
Your final response \textbf{must be valid JSON} exactly matching the schema below.\\
Please output \textbf{ONLY} the JSON object. Do not include any extra text.
\end{adjustwidth}

\begin{verbatim}
{
  "skill_combination": [<target_skill>, <new_combined_skill_1>,
  <new_combined_skill_2>, ...],
  "crossover_justification": <brief explanation of why and how these skills 
  work well together>,
  "code": "```python\n<code snippet of the new skill combination>\n```",
  "input": "```input\n<input of the new skill combination>\n```"
}
\end{verbatim}

\PartHeader{Evaluation Criteria}

\begin{adjustwidth}{0.6em}{0pt}
\begin{itemize}
  \item \textbf{Novelty (Critical):} skill combination differs from all existing combinations.
  \item \textbf{Integration (High Priority):} all skills work together meaningfully.
  \item \textbf{Target Skill Focus:} target skill is the dominant concept being tested.
  \item \textbf{Executability:} code runs successfully with the provided input.
  \item \textbf{Difficulty:} reversing the hidden input requires understanding the interaction between skills.
  \item \textbf{Creativity:} scenario is distinct from existing snippets.
  \item \textbf{Complexity:} emphasize algorithmic reasoning/logic complexity (data structures, control flow, DP, recursion).
  \item \textbf{Restricted Keywords:} you cannot use the following words in any form: \texttt{<|BANNED\_KEYWORDS|>}.
\end{itemize}
\end{adjustwidth}

\PartHeader{Process}

\begin{adjustwidth}{0.6em}{0pt}
\begin{enumerate}
  \item Analyze \texttt{existing\_combinations} to identify gaps.
  \item Propose a new skill combination with a clear justification.
  \item Devise a plan for how the skills interact to conceal the input.
  \item Implement the code snippet ensuring every skill is essential and requirements are met.
  \item Provide the hidden input that exercises the full combination.
\end{enumerate}
\end{adjustwidth}

If there is previous failure information, address those issues explicitly in your new attempt.

\end{tcolorbox}
\vspace{-2pt}
\captionof{figure}{Abduction Task Crossover Prompt}
\label{fig:code_input_crossover_prompt}

\begin{tcolorbox}[
  colback=gray!5!white,
  colframe=gray!75!black,
  title=\bfseries Abduction Task Predictor Prompt,
  width=\textwidth,
  boxrule=0.8pt,
  arc=4pt,
  outer arc=4pt,
  boxsep=4pt,
  left=3pt,
  right=6pt,
  top=4pt,
  bottom=4pt,
  breakable,
  fontupper=\scriptsize
]

\setlength{\parskip}{0.2em}
\setlength{\parindent}{0pt}

\newcommand{\PartHeader}[1]{%
  \smallskip
  {\footnotesize\bfseries\scshape #1}%
  \par\smallskip
}
\newcommand{\SubHeader}[1]{%
  \par\smallskip
  \hspace*{0.15em}{\footnotesize\bfseries #1}%
  \par\smallskip
}

\setlist[itemize]{%
  leftmargin=1.4em,
  itemsep=0.2em,
  topsep=0pt,
  parsep=0pt,
  partopsep=0pt
}
\setlist[enumerate,1]{%
  leftmargin=1.9em,
  labelwidth=1.2em,
  labelsep=0.4em,
  align=left,
  itemsep=0.2em,
  topsep=0pt,
  parsep=0pt,
  partopsep=0pt
}

\PartHeader{Task}

\begin{adjustwidth}{0.6em}{0pt}
Provide One Possible Input of a Python Code Snippet Given the Code and Output
Given the following Code Snippet and the Output, think step by step then provide one possible input that produced the output. The input needs to be wrapped in \texttt{```input```} tags. Remember if an argument is a string, wrap it in quotes. If the function requires multiple arguments, separate them with commas.
\end{adjustwidth}

\PartHeader{Code Snippet}

\begin{adjustwidth}{0.6em}{0pt}
\begin{verbatim}
```python
{snippet}
```
\end{verbatim}
\end{adjustwidth}

\PartHeader{Output}

\begin{adjustwidth}{0.6em}{0pt}
\vspace{-0.6em}
\begin{verbatim}
```output
{output}
```
\end{verbatim}
\vspace{-0.6em}
\end{adjustwidth}

\PartHeader{Output Format}

\begin{adjustwidth}{0.6em}{0pt}
\vspace{-0.6em}
\begin{verbatim}
```input
arg1, arg2, ...
```
\end{verbatim}
\vspace{-0.6em}
\end{adjustwidth}

\SubHeader{Example Format}

\begin{adjustwidth}{0.6em}{0pt}
\begin{verbatim}
```input
'John', {{'age': 20, 'city': 'New York'}}
```
\end{verbatim}
\end{adjustwidth}

\end{tcolorbox}
\vspace{-2pt}
\captionof{figure}{Abduction Task Predictor Prompt}
\label{fig:code_input_predictor_prompt}

\begin{tcolorbox}[
  colback=gray!5!white,
  colframe=gray!75!black,
  title=\bfseries Deduction Task Prompt,
  width=\textwidth,
  boxrule=0.8pt,
  arc=4pt,
  outer arc=4pt,
  boxsep=4pt,
  left=3pt,
  right=6pt,
  top=4pt,
  bottom=4pt,
  breakable,
  fontupper=\scriptsize
]

\setlength{\parskip}{0.2em}
\setlength{\parindent}{0pt}

\newcommand{\PartHeader}[1]{%
  \smallskip
  {\footnotesize\bfseries\scshape #1}%
  \par\smallskip
}
\newcommand{\SubHeader}[1]{%
  \par\smallskip
  \hspace*{0.15em}{\footnotesize\bfseries #1}%
  \par\smallskip
}

\setlist[itemize]{%
  leftmargin=1.4em,
  itemsep=0.2em,
  topsep=0pt,
  parsep=0pt,
  partopsep=0pt
}
\setlist[enumerate,1]{%
  leftmargin=1.9em,
  labelwidth=1.2em,
  labelsep=0.4em,
  align=left,
  itemsep=0.2em,
  topsep=0pt,
  parsep=0pt,
  partopsep=0pt
}

\PartHeader{Task}

\begin{adjustwidth}{0.6em}{0pt}
Create a New Python Code Snippet (where custom classes are allowed, which should be defined at the top of the code snippet) with one Matching 

\smallskip
\texttt{\{failure\_info\}}

\smallskip
\textbf{Skills:}\\
\texttt{\{skill\_str\}}
\end{adjustwidth}

\PartHeader{Code Requirements}

\begin{adjustwidth}{0.6em}{0pt}
\begin{itemize}
  \item The provided skill(s) should be the main testing concept of the code snippet.
  \item Name the entry function \texttt{f} (e.g., \texttt{def f(...): ...}), you can have nested definitions inside \texttt{f}
  \item Ensure the function returns a value
  \item Include at least one input parameter
  \item Make the function deterministic
  \item Make the snippet require state tracking across multiple data transformations, ensuring the task requires long multi-step reasoning.
  \item Avoid unnecessary function nesting; use simple, readable code structure when possible. Only use nested functions if it is natural for the algorithm
  \item AVOID THE FOLLOWING::
    \begin{itemize}
      \item Random functions or variables
      \item Date/time operations
      \item I/O operations (files, network requests)
      \item Printing or logging
      \item Any external state
    \end{itemize}
  \item Ensure execution completes within \textbf{10 seconds} on a modern CPU
  \item All imports and class definitions must be at the very top of the code snippet
  \item The snippet should end with a return statement from the main function  \texttt{f}, anything after will be removed.
  \texttt{\{remove\_input\_from\_snippet\_prompt\}\{remove\_after\_return\_prompt\}}
\end{itemize}
\end{adjustwidth}

\PartHeader{Input Requirements}

\begin{adjustwidth}{0.6em}{0pt}
\begin{itemize}
  \item Provide exactly one test input for your function
  \item Format multiple arguments with commas between them
  \item Remember to add quotes around string arguments
\end{itemize}
\end{adjustwidth}

\PartHeader{Formatting}

\begin{adjustwidth}{0.6em}{0pt}
\begin{itemize}
  \item Format your code with:
\end{itemize}
\end{adjustwidth}
\vspace{-0.6em}
\begin{verbatim}
```python
def f(...):
    # your code here
    return ...
```
\end{verbatim}
\vspace{-0.6em}
\begin{adjustwidth}{0.6em}{0pt}
\begin{itemize}
  \item Format your input with:
\end{itemize}
\end{adjustwidth}
\vspace{-0.6em}
\begin{verbatim}
```input
arg1, arg2, ...
```
\end{verbatim}
\vspace{-0.6em}
\SubHeader{Example Format:}
\vspace{-0.6em}
\begin{verbatim}
```python
def f(name: str, info: dict):
    # code logic here
    return result
```

```input
'John', {{'age': 20, 'city': 'New York'}}
```
\end{verbatim}
\vspace{-0.6em}
\PartHeader{Evaluation Criteria}

\begin{adjustwidth}{0.6em}{0pt}
\begin{itemize}
  \item Relevance (High Priority), the task should mainly focus on examining the ability of the test subject to understand / reason / apply the skill(s)
  \item Executability, your code should be executable given your input
  \item Difficulty in predicting your \texttt{input} from 1) your \texttt{python} code and 2) the deterministic \texttt{output} that will be obtained from your \texttt{input}. Focus on either algorithmic reasoning or logic complexity. For example, you can define complex data structure classes and operate on them like trees, heaps, stacks, queues, graphs, etc, or use complex control flow, dynamic programming, recursions, divide and conquer, greedy, backtracking, etc
  \item Creativity, the code needs to be sufficiently different from the provided reference snippets
  \item Restricted usage of certain keywords and packages, you are not allowed to use the following words in any form, even in comments: \texttt{<|BANNED\_KEYWORDS|>}
\end{itemize}
\end{adjustwidth}

\begin{adjustwidth}{0.6em}{0pt}
First, carefully devise a clear plan: e.g. how to make the skill(s) the main testing concept of the task, identify how your snippet will be challenging and creative. If there is previous failure information, think about how to resolve the failure and create acceptable code snippet. Then, write the final code snippet and its inputs.
\end{adjustwidth}

\end{tcolorbox}
\vspace{-2pt}
\captionof{figure}{Deduction Task Prompt}
\label{fig:code_output_prompt}

\begin{tcolorbox}[
  colback=gray!5!white,
  colframe=gray!75!black,
  title=\bfseries Deduction Task Mutation Prompt,
  width=\textwidth,
  boxrule=0.8pt,
  arc=4pt,
  outer arc=4pt,
  boxsep=4pt,
  left=3pt,
  right=6pt,
  top=4pt,
  bottom=4pt,
  breakable,
  fontupper=\scriptsize
]

\setlength{\parskip}{0.2em}
\setlength{\parindent}{0pt}

\newcommand{\PartHeader}[1]{%
  \smallskip
  {\footnotesize\bfseries\scshape #1}%
  \par\smallskip
}
\newcommand{\SubHeader}[1]{%
  \par\smallskip
  \hspace*{0.15em}{\footnotesize\bfseries #1}%
  \par\smallskip
}

\setlist[itemize]{%
  leftmargin=1.4em,
  itemsep=0.2em,
  topsep=0pt,
  parsep=0pt,
  partopsep=0pt
}
\setlist[enumerate,1]{%
  leftmargin=1.9em,
  labelwidth=1.2em,
  labelsep=0.4em,
  align=left,
  itemsep=0.2em,
  topsep=0pt,
  parsep=0pt,
  partopsep=0pt
}

\PartHeader{Task}

\begin{adjustwidth}{0.6em}{0pt}
Create Multiple Variants of an Original Task by Systematically Applying Complexity Attributes to Increase Complexity and Reasoning Requirements.

\smallskip
The original task is a code-reasoning deduction task that demands deep algorithmic reasoning to deduce the output from the input and Python code snippet.

\smallskip
You will be provided with:
\begin{itemize}
  \item Original task: one Python code snippet and one input
  \item Skill: the algorithmic skill(s) that the original task mainly tests
  \item Complexity attributes: a list of complexity attributes you can consider to increase the complexity and reasoning requirements of the original task.
\end{itemize}

\smallskip
Be aware that not all attributes are applicable to the original task. You must first determine which attributes could be used.
\end{adjustwidth}

\PartHeader{Your Mission}

\SubHeader{Step 1: Analyze the Original Task}

\begin{adjustwidth}{0.6em}{0pt}
Examine the provided task and identify:
\begin{itemize}
  \item Current complexity level for each \textbf{applicable} attribute
  \item Potential for complexity enhancement in each dimension
\end{itemize}
\end{adjustwidth}

\SubHeader{Step 2: Generate Diverse Variants}

\begin{adjustwidth}{0.6em}{0pt}
Create \textbf{at least 10 different variants} via the following approaches:
\begin{itemize}
  \item Single-attribute mutations: Enhance only one complexity dimension
  \item Multi-attribute mutations: Modify multiple but not all attributes at the same time
  \item Comprehensive mutations: Adjust all applicable attributes together
\end{itemize}
\end{adjustwidth}

\PartHeader{Requirements for Each Variant}

\SubHeader{Code Requirements}

\begin{adjustwidth}{0.6em}{0pt}
The provided skill(s) should be the main testing concept of the code snippet.
\begin{itemize}
  \item Name the entry function \texttt{f} (e.g., \texttt{def f(...): ...}), you can have nested definitions inside \texttt{f}
  \item Ensure the function returns a value.
  \item Include at least one input parameter.
  \item Make the function deterministic.
  \item Make the snippet require state tracking across multiple data transformations, ensuring the task requires long multi step reasoning.
  \item Avoid unnecessary function nesting; use simple, readable code structure when possible. Only use nested functions if it is natural for the algorithm
  \item AVOID THE FOLLOWING:
    \begin{itemize}
      \item Random functions or variables
      \item Date/time operations
      \item I/O operations (reading files, network requests)
      \item Printing or logging
      \item Any external state
    \end{itemize}
  \item Ensure execution completes within 10 seconds on a modern CPU
  \item All imports and class definitions should be at the very top of the code snippet
  \item The snippet should end with a return statement from the main function `f`, anything after will be removed
  \item \texttt{\{remove\_input\_from\_snippet\_prompt\}\{remove\_after\_return\_prompt\}}
\end{itemize}
\end{adjustwidth}

\SubHeader{Input Requirements}

\begin{adjustwidth}{0.6em}{0pt}
\begin{itemize}
  \item Provide exactly one test input for your function
  \item Format multiple arguments with commas between them
  \item Remember to add quotes around string arguments.
\end{itemize}
\end{adjustwidth}

\SubHeader{Formatting}

\begin{adjustwidth}{0.6em}{0pt}
\begin{itemize}
  \item Format your code with:
\end{itemize}
\end{adjustwidth}
\vspace{-0.6em}
\begin{verbatim}
```python
def f(...):
    # your code here
    return ...
```
\end{verbatim}
\vspace{-0.6em}
\begin{adjustwidth}{0.6em}{0pt}
\begin{itemize}
  \item Format your input with:
\end{itemize}
\end{adjustwidth}
\vspace{-0.6em}
\begin{verbatim}
```input
arg1, arg2, ...
```
\end{verbatim}
\vspace{-0.6em}
\SubHeader{Example Format}
\vspace{-0.6em}
\begin{verbatim}
```python
def f(name: str, info: dict):
    # code logic here
    return result
```

```input
'John', {{{{'age': 20, 'city': 'New York'}}}}
```
\end{verbatim}
\vspace{-0.6em}
\SubHeader{Complexity Enhancements}

\begin{adjustwidth}{0.6em}{0pt}
\begin{itemize}
  \item Preserve the core algorithmic skill being tested.
  \item Consider the original task's difficulty (pass rate) when increasing complexity. The variants should not be too easy (pass rate = 100\%) or too hard (pass rate = 0\%).
  \item Each variant must be meaningfully different from others.
  \item Maintain solvability while increasing complexity.
  \item Output the complexity attributes that you considered for each variant and one concise description of how you increase the complexity by each attribute and why it is more complex than the original task.
\end{itemize}
\end{adjustwidth}

\PartHeader{Output Format}

\begin{adjustwidth}{0.6em}{0pt}
Your final response \textbf{must be valid JSON} exactly matching the schema below.\\
Please output \textbf{ONLY} the JSON object. Do not include any extra things.
JSON SCHEMA TO FOLLOW:
\end{adjustwidth}
\vspace{-0.6em}
\begin{verbatim}
{
  "variant_1": {
    "complexity_attributes": [
      <name of the complexity attribute being considered>,
      ...
    ],
    "description": <description of how variant_1 is more complex than the original task
    by the complexity attributes you considered>,
    "code": "```python\n<code snippet of variant_1>\n```",
    "input": "```input\n<input of variant_1>\n```"
  },
  "variant_2": ...,
  ...
}
\end{verbatim}
\vspace{-0.6em}
\PartHeader{Generation Strategy}

\begin{adjustwidth}{0.6em}{0pt}
\begin{enumerate}
  \item \textbf{Breadth First:} Generate variants that explore different complexity attributes before combining them
  \item \textbf{Difficulty Progression:} Include variants ranging from moderate increases to significant complexity jumps
  \item \textbf{Reasoning Diversity:} Create variants that fail for different reasons (timeout, overflow, logic errors, edge cases).
\end{enumerate}
\end{adjustwidth}

\PartHeader{Evaluation Criteria}

\begin{adjustwidth}{0.6em}{0pt}
\begin{itemize}
  \item Relevance to core algorithmic skills: Each variant should test the same core skill.
  \item Complexity increases should be meaningful, not artificial.
  \item Variants should explore different failure modes and challenge different aspects of problem-solving.
  \item Executability: Each variant must be executable given its input.
\end{itemize}
\end{adjustwidth}

\begin{adjustwidth}{0.6em}{0pt}
Remember: The goal is to create a rich set of practice problems that gradually build expertise by challenging different aspects of the skill through varied complexity enhancements.
\end{adjustwidth}

\PartHeader{Original Task}

\begin{adjustwidth}{0.6em}{0pt}
\textbf{Code:}\\
\texttt{\{code\}}

\smallskip
\textbf{Input:}\\
\texttt{\{input\}}

\smallskip
\textbf{Skill:}\\
\texttt{\{skill\}}

\smallskip
\textbf{Complexity Attributes:}\\
\texttt{\{complexity\_attributes\}}
\end{adjustwidth}

\end{tcolorbox}
\vspace{-2pt}
\captionof{figure}{Deduction Task Mutation Prompt}
\label{fig:code_output_mutation_prompt}

\begin{tcolorbox}[
  colback=gray!5!white,
  colframe=gray!75!black,
  title=\bfseries Deduction Task Crossover Prompt,
  width=\textwidth,
  boxrule=0.8pt,
  arc=4pt,
  outer arc=4pt,
  boxsep=4pt,
  left=3pt,
  right=6pt,
  top=4pt,
  bottom=4pt,
  breakable,
  fontupper=\scriptsize
]

\setlength{\parskip}{0.2em}
\setlength{\parindent}{0pt}

\newcommand{\PartHeader}[1]{%
  \smallskip
  {\footnotesize\bfseries\scshape #1}%
  \par\smallskip
}
\newcommand{\SubHeader}[1]{%
  \par\smallskip
  \hspace*{0.15em}{\footnotesize\bfseries #1}%
  \par\smallskip
}

\setlist[itemize]{%
  leftmargin=1.4em,
  itemsep=0.2em,
  topsep=0pt,
  parsep=0pt,
  partopsep=0pt
}
\setlist[enumerate,1]{%
  leftmargin=1.9em,
  labelwidth=1.2em,
  labelsep=0.4em,
  align=left,
  itemsep=0.2em,
  topsep=0pt,
  parsep=0pt,
  partopsep=0pt
}

\PartHeader{Task}

\begin{adjustwidth}{0.6em}{0pt}
Create a New Python Code Snippet (where custom classes are allowed, which should be defined at the top of the code snippet) with one Matching Input and with Novel Skill Combination
\end{adjustwidth}

\PartHeader{Available Information}

\begin{adjustwidth}{0.6em}{0pt}
\textbf{Skill Pool:}\\
\texttt{\{skill\_pool\}}\\
A comprehensive list of algorithmic coding skills along with their descriptions that you can choose from to create new combinations

\smallskip
\textbf{Target Core Skill:}\\
\texttt{\{target\_skill\}}\\
The primary skill that must be the central focus of your code snippet along with its description

\smallskip
\textbf{Existing Skill Combinations:}\\
\texttt{\{existing\_combinations\}}\\
Previously created combinations of the target skill with other skills
\end{adjustwidth}

\PartHeader{Your Task}

\begin{adjustwidth}{0.6em}{0pt}
Create a novel crossover by combining the target skill with other compatible skills from the pool. This requires:
\begin{enumerate}
  \item Examining existing combinations to understand what has already been done
  \item Identifying skills from the pool that would naturally work with the target skill but haven't been combined yet
  \item Creating a code snippet where all chosen skills are essential and interconnected to solve the problem
\end{enumerate}

\smallskip
The key challenge is to ensure your skill combination is both NEW (not in existing combinations) and MEANINGFUL (skills genuinely complement each other rather than being artificially forced together).
\end{adjustwidth}

\PartHeader{Code Requirements}

\begin{adjustwidth}{0.6em}{0pt}
\begin{itemize}
  \item The target skill MUST be the central concept, with other skills naturally supporting or enhancing it
  \item Name the entry function \texttt{f} (e.g., \texttt{def f(...): ...}), you can have nested definitions inside
  \item Ensure the function returns a value
  \item Include at least one input parameter
  \item Make the function deterministic
  \item The snippet should demonstrate clear interdependence between all skills in the combination (not just using skills in isolation)
  \item Require state tracking across multiple data transformations, ensuring multi-step reasoning
  \item Avoid unnecessary function nesting; use simple, readable code structure when possible
  \item AVOID THE FOLLOWING
    \begin{itemize}
      \item Random functions or variables
      \item Date/time operations
      \item I/O operations (reading files, network requests)
      \item Printing or logging
      \item Any external state
    \end{itemize}
  \item Ensure execution completes within 10 seconds on a modern CPU
  \item All imports and class definitions should be at the very top of the code snippet
  \item The snippet should end with a return statement from the main function \texttt{f}.
  \item \texttt{\{remove\_input\_from\_snippet\_prompt\}\{remove\_after\_return\_prompt\}}
\end{itemize}
\end{adjustwidth}

\PartHeader{Skill Combination Requirements}

\begin{adjustwidth}{0.6em}{0pt}
\begin{itemize}
  \item Your combination must include the target skill plus at least one other skill from the skill pool
  \item The combination must be meaningfully different from all existing combinations
  \item Skills should work together synergistically, not just appear sequentially
  \item Justify why your chosen skills complement each other naturally and how they work together in your code snippet
\end{itemize}
\end{adjustwidth}

\PartHeader{Input Requirements}

\begin{adjustwidth}{0.6em}{0pt}
\begin{itemize}
  \item Provide exactly one test input for your function
  \item Format multiple arguments with commas between them
  \item Remember to add quotes around string arguments
\end{itemize}
\end{adjustwidth}

\PartHeader{Formatting}

\begin{adjustwidth}{0.6em}{0pt}
\begin{itemize}
  \item Format your code with:
\end{itemize}
\end{adjustwidth}
\vspace{-0.6em}
\begin{verbatim}
```python
def f(...):
    # your code here
    return ...
```
\end{verbatim}
\vspace{-0.6em}
\begin{adjustwidth}{0.6em}{0pt}
\begin{itemize}
  \item Format your input with:
\end{itemize}
\end{adjustwidth}
\vspace{-0.6em}
\begin{verbatim}
```input
arg1, arg2, ...
```
\end{verbatim}
\vspace{-0.6em}
\SubHeader{Example Format:}
\vspace{-0.6em}
\begin{verbatim}
```python
def f(name: str, info: dict):
    # code logic here
    return result
```

```input
'John', {{'age': 20, 'city': 'New York'}}
```
\end{verbatim}
\vspace{-0.6em}
\PartHeader{Output Format}

\begin{adjustwidth}{0.6em}{0pt}
Your final response \textbf{must be valid JSON} exactly matching the schema below. Please output \textbf{ONLY} the JSON object. Do not include any extra things.
JSON SCHEMA TO FOLLOW:
\end{adjustwidth}
\vspace{-0.6em}
\begin{verbatim}
{
  "skill_combination": [<target_skill>, <new_combined_skill_1>, <new_combined_skill_2>,
  
  ...],
  "crossover_description": <brief explanation of why and how these skills work well
  together>,
  "code": "```python\n<code snippet of the new skill combination>\n```",
  "input": "```input\n<input of the new skill combination>\n```"
}
\end{verbatim}
\vspace{-0.6em}
\PartHeader{Evaluation Criteria}

\begin{adjustwidth}{0.6em}{0pt}
\begin{itemize}
  \item Novelty (Critical): The skill combination must be different from all existing combinations
  \item Integration (High Priority): All skills must work together meaningfully, not just appear independently
  \item Target Skill Focus: The target skill should be the dominant concept being tested
  \item Executability: Your code should run successfully with the provided input
  \item Difficulty: The task should require understanding the interaction between skills, not just individual skill mastery
  \item Creativity: The scenario should be distinct from existing code snippets
  \item Complexity: Focus on algorithmic reasoning or logic complexity (e.g., complex data structures, control flow, dynamic programming, recursion)
  \item Restricted Keywords: You cannot use the following words in any form: \texttt{<|BANNED\_KEYWORDS|>}.
\end{itemize}
\end{adjustwidth}

\PartHeader{Process}

\begin{adjustwidth}{0.6em}{0pt}
\begin{enumerate}
  \item First, analyze the existing combinations to understand patterns and identify gaps
  \item Second, propose your new skill combination with clear justification
  \item Third, devise a plan for how these skills will interact in your code
  \item Fourth, implement the code snippet ensuring all skills are essential to the solution and following the code requirements
  \item Finally, create an input that exercises the full combination
\end{enumerate}

\smallskip
If there is previous failure information, address those issues explicitly in your new attempt.
\end{adjustwidth}

\end{tcolorbox}
\vspace{-2pt}
\captionof{figure}{Deduction Task Crossover Prompt}
\label{fig:code_output_crossover_prompt}


\begin{tcolorbox}[
  colback=gray!5!white,
  colframe=gray!75!black,
  title=\bfseries Deduction Task Predictor Prompt,
  width=\textwidth,
  boxrule=0.8pt,
  arc=4pt,
  outer arc=4pt,
  boxsep=4pt,
  left=3pt,
  right=6pt,
  top=4pt,
  bottom=4pt,
  breakable,
  fontupper=\scriptsize
]

\setlength{\parskip}{0.2em}
\setlength{\parindent}{0pt}

\newcommand{\PartHeader}[1]{%
  \smallskip
  {\footnotesize\bfseries\scshape #1}%
  \par\smallskip
}
\newcommand{\SubHeader}[1]{%
  \par\smallskip
  \hspace*{0.15em}{\footnotesize\bfseries #1}%
  \par\smallskip
}

\setlist[itemize]{%
  leftmargin=1.4em,
  itemsep=0.2em,
  topsep=0pt,
  parsep=0pt,
  partopsep=0pt
}
\setlist[enumerate,1]{%
  leftmargin=1.9em,
  labelwidth=1.2em,
  labelsep=0.4em,
  align=left,
  itemsep=0.2em,
  topsep=0pt,
  parsep=0pt,
  partopsep=0pt
}

\PartHeader{Task}

\begin{adjustwidth}{0.6em}{0pt}

Deduce the Output of a Python Code Snippet Given the Code and Input
Given the following Code Snippet and the Input, think step by step then deduce the output that will be produced from plugging the Input into the Code Snippet. Put your output in \texttt{```output```} tags. Remember if the output is a string, wrap it in quotes. If the function returns multiple values, remember to use a tuple to wrap them.

\end{adjustwidth}

\PartHeader{Code Snippet}
\vspace{-0.6em}
\begin{verbatim}
```python
{snippet}
```
\end{verbatim}
\vspace{-0.6em}

\PartHeader{Input}
\vspace{-0.6em}
\begin{verbatim}
```input
{input_args}
```
\end{verbatim}
\vspace{-0.6em}
\SubHeader{Example Output:}
\vspace{-0.6em}
\begin{verbatim}
```output
{'age': 20, 'city': 'New York'}
```
\end{verbatim}
\vspace{-0.6em}
\end{tcolorbox}
\vspace{-2pt}
\captionof{figure}{Deduction Task Predictor Prompt}
\label{fig:code_output_predictor_prompt}

\begin{tcolorbox}[
  colback=gray!5!white,
  colframe=gray!75!black,
  title=\bfseries Induction Task Prompt,
  width=\textwidth,
  boxrule=0.8pt,
  arc=4pt,
  outer arc=4pt,
  boxsep=4pt,
  left=3pt,
  right=6pt,
  top=4pt,
  bottom=4pt,
  breakable,
  fontupper=\scriptsize
]

\setlength{\parskip}{0.2em}
\setlength{\parindent}{0pt}

\newcommand{\PartHeader}[1]{%
  \smallskip
  {\footnotesize\bfseries\scshape #1}%
  \par\smallskip
}
\newcommand{\SubHeader}[1]{%
  \par\smallskip
  \hspace*{0.15em}{\footnotesize\bfseries #1}%
  \par\smallskip
}

\setlist[itemize]{%
  leftmargin=1.4em,
  itemsep=0.2em,
  topsep=0pt,
  parsep=0pt,
  partopsep=0pt
}
\setlist[enumerate,1]{%
  leftmargin=1.9em,
  labelwidth=1.2em,
  labelsep=0.4em,
  align=left,
  itemsep=0.2em,
  topsep=0pt,
  parsep=0pt,
  partopsep=0pt
}

\PartHeader{Task}

\begin{adjustwidth}{0.6em}{0pt}
Generate a natural coding problem related to the code snippet, and \texttt{\{num\_inputs\}} inputs that can be plugged into the code snippet to produce a diverse set of outputs.

Using the code snippet provided below, design comprehensive \texttt{\{num\_inputs\}} inputs that can be plugged into the code snippet to produce a diverse set of outputs. A subset of your given input and its deterministically produced outputs will be given to a test subject to deduce the function, which is meant to be an I.Q. test. You should also create a natural coding problem for which the given code snippet would be a valid solution, and your generated inputs would be the test inputs for the problem. 
\end{adjustwidth}

\PartHeader{Input Requirements}

\begin{adjustwidth}{0.6em}{0pt}
\begin{itemize}
  \item Provide \texttt{\{num\_inputs\}} valid inputs for the code snippet that comprehensively cover the code's behavior.
  \item For each input, format multiple arguments with commas between them
  \item Remember to add quotes around string arguments
  \item Each input should be individually wrapped in \texttt{```input```} tags
  \item Ensure diversity: inputs should test different aspects and branches of the code
\end{itemize}
\end{adjustwidth}

\PartHeader{Problem Requirements}

\begin{adjustwidth}{0.6em}{0pt}
\begin{itemize}
  \item Create a natural coding problem that clearly describes what needs to be solved. Do not include examples or constraints
  \item Write in an engaging, scenario-based style when possible (e.g., "You are given an array of meeting times..." or "A company needs to process customer orders...")
  \item The provided code snippet must be a correct and complete solution to the problem you describe
  \item Ensure that solving the problem statement would naturally lead to implementing logic similar to the given code snippet
  \item The problem statement should be wrapped in \texttt{```problem```} tags
  \item You cannot include or leak the code snippet in the problem statement
\end{itemize}
\end{adjustwidth}

\PartHeader{Formatting}

\begin{adjustwidth}{0.6em}{0pt}
\begin{itemize}
  \item Format your input with:
\end{itemize}
\end{adjustwidth}
\vspace{-0.6em}
\begin{verbatim}
```input
arg1, arg2, ...
```
\end{verbatim}
\vspace{-0.6em}
\SubHeader{Example Format:}
\vspace{-0.6em}
\begin{verbatim}
```input
'John', {{'age': 20, 'city': 'New York'}}
```
```input
'Sammy', {{'age': 37, 'city': 'Los Angeles'}}
```
\end{verbatim}
\vspace{-0.6em}
\PartHeader{Evaluation Criteria}

\begin{adjustwidth}{0.6em}{0pt}
\begin{itemize}
  \item Executability, the code should be executable given your inputs
  \item Coverage, the inputs should cover the whole input space of the code snippet
  \item Creativity, the inputs need to be sufficiently different from each other
  \item The overall selection of inputs and message combined should be challenging for the test subject, but not impossible for them to solve
  \item Problem Quality, The problem statement should read like a real coding assessment question
\end{itemize}
\end{adjustwidth}

\PartHeader{Process}

\begin{adjustwidth}{0.6em}{0pt}
First, carefully devise a clear plan: e.g. understand the code snippet, then identify how your proposed inputs have high coverage, and why the inputs will be challenging and creative. Then, write the inputs and message. Remember to wrap your inputs in \texttt{```input```} tags, and your message in \texttt{```message```} tags.
\end{adjustwidth}

\PartHeader{Code Snippet}
\vspace{-0.6em}
\begin{verbatim}
```python
{code}
```
\end{verbatim}
\vspace{-0.6em}
\end{tcolorbox}
\vspace{-2pt}
\captionof{figure}{Induction Task Prompt}
\label{fig:code_function_prompt}

\begin{tcolorbox}[
  colback=gray!5!white,
  colframe=gray!75!black,
  title=\bfseries Induction Task Hint Prompt,
  width=\textwidth,
  boxrule=0.8pt,
  arc=4pt,
  outer arc=4pt,
  boxsep=4pt,
  left=3pt,
  right=6pt,
  top=4pt,
  bottom=4pt,
  breakable,
  fontupper=\scriptsize
]

\setlength{\parskip}{0.2em}
\setlength{\parindent}{0pt}

\newcommand{\PartHeader}[1]{%
  \smallskip
  {\footnotesize\bfseries\scshape #1}%
  \par\smallskip
}
\newcommand{\SubHeader}[1]{%
  \par\smallskip
  \hspace*{0.15em}{\footnotesize\bfseries #1}%
  \par\smallskip
}

\setlist[itemize]{%
  leftmargin=1.4em,
  itemsep=0.2em,
  topsep=0pt,
  parsep=0pt,
  partopsep=0pt
}
\setlist[enumerate,1]{%
  leftmargin=1.9em,
  labelwidth=1.2em,
  labelsep=0.4em,
  align=left,
  itemsep=0.2em,
  topsep=0pt,
  parsep=0pt,
  partopsep=0pt
}

\PartHeader{Task}

\begin{adjustwidth}{0.6em}{0pt}
Generate progressive hints for a coding problem.

\smallskip
The coding problem below was generated based on a code snippet and has proven too challenging for test subjects to solve. Your task is to create a series of progressive hints that guide the solver toward the solution WITHOUT giving away the complete answer
\end{adjustwidth}

\PartHeader{Original Problem}
\vspace{-0.6em}
\begin{verbatim}
```problem
{problem}
```
\end{verbatim}
\vspace{-0.6em}
\PartHeader{Code Snippet}
\vspace{-0.6em}
\begin{verbatim}
```python
{code}
```
\end{verbatim}
\vspace{-0.6em}
\PartHeader{Hints Requirements}

\begin{adjustwidth}{0.6em}{0pt}
\begin{itemize}
  \item Generate 3--4 progressive hints that gradually reveal the solution approach.
  \item Follow the hint style:
    \begin{itemize}
      \item Start with high-level intuition or pattern recognition
      \item Progress to algorithm/data-structure suggestions
      \item Then provide implementation details or key insights
      \item Final hint can outline the approach at a coarse level but should not reveal the actual code implementation
    \end{itemize}
  \item Each hint should build upon the previous one
  \item Hints should be concise, to the point, and guide thinking without revealing the exact solution
  \item Ensure the hints are grounded to the code snippet, which is the solution to the problem
  \item Wrap each hint in \texttt{```hint```} tags.
\end{itemize}
\end{adjustwidth}

\PartHeader{Formatting}

\begin{adjustwidth}{0.6em}{0pt}
\begin{itemize}
  \item Format your hints with:
\end{itemize}
\end{adjustwidth}
\vspace{-0.6em}
\begin{verbatim}
```hint
<hint_content>
```
\end{verbatim}
\vspace{-0.6em}
\SubHeader{Example Format:}

```hint
Can you think of this problem in terms of a decision tree, where at each step we have n decisions, where n is the size of the array?
```

```hint
We can use backtracking to recursively traverse these paths and make decisions to choose an element at each step.
```

\PartHeader{Evaluation Criteria}

\begin{adjustwidth}{0.6em}{0pt}
\begin{itemize}
  \item Progressive Difficulty: Each hint should reveal slightly more than the previous
  \item Guidance Quality: Hints should genuinely help stuck test subjects to make progress, not just give answers
  \item Clarity: Use simple language and clear explanations
\end{itemize}
\end{adjustwidth}

\begin{adjustwidth}{0.6em}{0pt}
First, carefully devise a clear plan: e.g. understand the code snippet, then identify what concepts or patterns might not be obvious, what edge cases could trip up solvers, what algorithmic insight is needed?. Then, create the hints. Remember to wrap your hints in \texttt{```hint```} tags.
\end{adjustwidth}

\end{tcolorbox}
\vspace{-2pt}
\captionof{figure}{Induction Task Hint Prompt}
\label{fig:code_function_hint_prompt}

\begin{tcolorbox}[
  colback=gray!5!white,
  colframe=gray!75!black,
  title=\bfseries Induction Task Predictor Prompt,
  width=\textwidth,
  boxrule=0.8pt,
  arc=4pt,
  outer arc=4pt,
  boxsep=4pt,
  left=3pt,
  right=6pt,
  top=4pt,
  bottom=4pt,
  breakable,
  fontupper=\scriptsize
]

\setlength{\parskip}{0.2em}
\setlength{\parindent}{0pt}

\newcommand{\PartHeader}[1]{%
  \smallskip
  {\footnotesize\bfseries\scshape #1}%
  \par\smallskip
}

\PartHeader{Task}

Deduce the Function that Produced the Outputs from the Inputs

Given a set of input/output pairs and a problem that describes the function, think through the problem step by step to deduce a general code snippet. This code should produce the hidden outputs from the hidden inputs, matching the original data-generating code that created the input/output pairs. Place your final answer inside python tags! It may be helpful to work through each input/output pair individually to test your function. If your function doesn't work as expected, revise it until it does. The final code snippet will be used to evaluate your response, which is wrapped in \texttt{```python```} tags.

\PartHeader{Code Requirements}

\begin{adjustwidth}{0.6em}{0pt}
\begin{itemize}
  \item Name the entry function \texttt{f} (e.g., \texttt{def f(...): ...}), you can have nested definitions inside \texttt{`f`}
  \item Ensure the function returns a value
  \item Include at least one input parameter
  \item Make the function deterministic
  \item AVOID THE FOLLOWING:
    \begin{itemize}
      \item Random functions or variables
      \item Date/time operations
      \item I/O operations (reading files, network requests)
      \item Printing or logging
      \item Any external state
    \end{itemize}
  \item Ensure execution completes within 10 seconds on a modern CPU
  \item All imports and class definitions should be at the very top of the code snippet
  \item The snippet should end with a return statement from the main function \texttt{`f()`}, anything after will be removed
\end{itemize}
\end{adjustwidth}
\PartHeader{Input and Output Pairs}
\vspace{-0.6em}
\begin{verbatim}
{input_output_pairs}
\end{verbatim}
\vspace{-0.6em}
\PartHeader{Problem}
\vspace{-0.6em}
\begin{verbatim}
{problem}
\end{verbatim}
\vspace{-0.6em}
\PartHeader{Example Output}
\vspace{-0.6em}
\begin{verbatim}
```python
def f(a):
    return a
```
\end{verbatim}
\vspace{-0.6em}
Name your entry function \texttt{f()}.

\end{tcolorbox}
\vspace{-2pt}
\captionof{figure}{Induction Task Predictor Prompt}
\label{fig:code_function_predictor_prompt}


\end{document}